\documentclass[journal]{IEEEtran}

\usepackage{cite}
\usepackage{amsmath,amssymb,amsfonts}
\usepackage{algcompatible}
\usepackage{algorithm,algorithmicx,algpseudocode}\usepackage{graphicx}
\usepackage{textcomp}
\usepackage{xcolor,url}
\usepackage{rotating}
\usepackage{color}
\usepackage{verbatim}		
\usepackage{subfigure,balance}
\newcommand{\tabincell}[2]{\begin{tabular}{@{}#1@{}}#2\end{tabular}}
\usepackage{booktabs}
\usepackage{tikz}
\usepackage{algorithm}  
\usepackage{multirow}
\usepackage{multicol}
\usepackage{bm}
\usepackage{epstopdf} 
\usepackage{rotating}

\usepackage{amsmath}  
\usepackage{multirow}
\usepackage{longtable}
\usepackage{tabularx}
\usepackage{supertabular}
\usepackage{adjustbox}
\usepackage{graphicx}  
\usepackage{float,balance}  

\usepackage{color,colortbl}
\definecolor{mygrey}{gray}{.9}

\def\NoNumber#1{{\def\alglinenumber##1{}\State #1}\addtocounter{ALG@line}{-1}}

\newcommand{\D}{\mathcal{D}}

\newcommand{\M}{\mathcal{M}}
\newcommand{\s}{\mathcal{S}}

\def\NoNumber#1{{\def\alglinenumber##1{}\State #1}\addtocounter{ALG@line}{-1}}

\begin{document}

\title{Mitigating Unfairness via \\ Evolutionary Multi-objective Ensemble Learning}

\author{Qingquan Zhang,~\IEEEmembership{Student Member,~IEEE},
Jialin Liu,~\IEEEmembership{Senior Member,~IEEE}, Zeqi Zhang, Junyi Wen, Bifei Mao,
Xin Yao,~\IEEEmembership{Fellow,~IEEE}

\thanks{This work has been accepted by IEEE Transactions on Evolutionary Computation.}

\thanks{Q. Zhang, J. Liu and X. Yao are with the Research Institute of Trustworthy Autonomous System, Southern University of Science and Technology, Shenzhen 518055, China. J. Liu and X. Yao are also with the Guangdong Provincial Key Laboratory of Brain-inspired Intelligent Computation, Department of Computer Science and Engineering, Southern University of Science and Technology, Shenzhen 518055, China. Corresponding author: Jialin Liu (liujl@sustech.edu.cn).}%
\thanks{Z. Zhang, J. Wen and B. Mao are with the Trustworthiness Theory Research Center, Huawei Technologies Co., Ltd., Shenzhen 518129, China.}%
\thanks{X. Yao is also with the School of Computer Science, University of Birmingham, UK.}

}

\maketitle

\begin{abstract}

In the literature of mitigating unfairness in machine learning, many fairness measures are designed to evaluate predictions of learning models and also utilised to guide the training of fair models. It has been theoretically and empirically shown that there exist conflicts and inconsistencies among accuracy and multiple fairness measures. Optimising one or several fairness measures may sacrifice or deteriorate other measures. Two key questions should be considered, how to simultaneously optimise accuracy and multiple fairness measures, and how to optimise all the considered fairness measures more effectively. In this paper, we view the mitigating unfairness problem as a multi-objective learning problem considering the conflicts among fairness measures. A multi-objective evolutionary learning framework is used to simultaneously optimise several metrics (including accuracy and multiple fairness measures) of machine learning models. Then, ensembles are constructed based on the learning models in order to automatically balance different metrics. Empirical results on eight well-known datasets demonstrate that compared with the state-of-the-art approaches for mitigating unfairness, our proposed algorithm can provide decision-makers with better tradeoffs among accuracy and multiple fairness metrics. Furthermore, the high-quality models generated by the framework can be used to construct an ensemble to automatically achieve a better tradeoff among all the considered fairness metrics than other ensemble methods. Our code is publicly available at https://github.com/qingquan63/FairEMOL
\end{abstract}

\begin{IEEEkeywords}
Multi-objective learning, Fairness in machine learning, Ensembles of learning machines, AI ethics, Fairness measures.
\end{IEEEkeywords}

\IEEEpeerreviewmaketitle

\section{Introduction}\label{sec:intro} 

\IEEEPARstart{A}{rtificial} intelligence ethics, including fairness, has been a very important topic~\cite{hutchinson201950,caton2020fairness,huang2022overview}. Fairness is viewed as a significant element in artificial intelligent ethics, which refers to \emph{``absence of any prejudice or favouritism toward an individual or a group based on their inherent or acquired characteristics''}~\cite{mehrabi2021survey} in the context of decision making. Due to the potential bias and discrimination of training data and algorithms~\cite{caton2020fairness}, unfair data-driven models and unfair decisions may be made. 

For over 50 years, myriad types of fairness quantification have emerged from many disciplines~\cite{hutchinson201950}, e.g., employment, education and finance, aiming to determine and evaluate (un)fairness. Since fairness in different contexts can be interpreted into different quantitative definitions to emphasise different perspectives, there is a lack of consensus among different measures and, no single measure has been accepted as a universal notion of fairness quantification~\cite{caton2020fairness,hutchinson201950,verma2018fairness}. Research has shown that different fairness measures often have conflicts~\cite{caton2020fairness,hutchinson201950,verma2018fairness}. In other words, if the performance of prediction results on a certain fairness measure is improved, the predictions may perform worse on at least one other fairness measure.

Many approaches have been proposed to mitigate the unfairness in machine learning (ML)~\cite{verma2018fairness,hutchinson201950,caton2020fairness}. Since data-driven models affected by unfair data will cause the unfair prediction results, mitigating bias could sacrifice the accuracy. Therefore, the main challenge of mitigating unfairness is how to make better tradeoffs between accuracy and fairness of learning models. Various mechanisms try to focus on one or more fairness metrics to mitigate bias. Two dilemmas exist for mitigating unfairness in the context of ML~\cite{caton2020fairness}: conflict among accuracy and fairness, and conflict among multiple fairness measures, which have been theoretically and empirically shown~\cite{caton2020fairness,friedler2019comparative}.

Existing work~\cite{caton2020fairness,berk2017convex,Goel_Yaghini_Faltings_2018} often performs a weighted average of metrics (including accuracy and one or more fairness metrics) to deal with the two dilemmas. For the first dilemma, due to the bias and unfair training data, improving fairness of models may degrade accuracy. Some approaches for mitigating unfairness consider a single fairness measure as a regularisation term~\cite{huang2019stable} or a constraint~\cite{zafar2017fairness} to get a tradeoff between the model accuracy and fairness. Regarding the latter dilemma, due to the incompatibility and complementarity among different fairness measures, such as individual fairness and group fairness~\cite{speicher2018unified}, the weighted sum approach was also used to combine two metrics into one~\cite{caton2020fairness}.

However, two challenges should be considered for the methods using the weighted sum approach. First, weights of different objectives (accuracy or fairness metrics) are difficult to determine. The slight difference among the weights may also lead to a large difference in the performance of the ML model. Second, the weighted sum approach can provide only one ML model with one specific tradeoff among conflicting metrics.

A group of diverse models with different tradeoffs among accuracy and multiple fairness are needed for decision-makers. Such a diverse set of fair models not only help decision makers to make a more informed choice but also facilitate the formation of ensemble of fair ML models.

A few studies~\cite{wu2021multifr, QingquanFair2021,padh2021addressing,liu2022accuracy} view the unfairness mitigation problem as a multi-objective optimisation problem. For example, the studies~\cite{wu2021multifr,padh2021addressing} proposed to convert gradients of multiple objectives (fairness measures) into one loss to train a model, where weights among objectives are adaptively determined. In the study~\cite{QingquanFair2021}, we proposed a framework based on multi-objective optimisation evolutionary learning and revealed that the proposed framework can simultaneously optimise accuracy and multiple fairness measures and obtain a group of diverse models. In this paper, we study this framework in depth. First, we evaluate this framework on a broader range of different datasets. Second, we investigate a more appropriate and representative set of metrics to use as objectives. Third, we develop ensembles of fair models to improve the accuracy and fairness. Our study will be organised around the following four research questions. ($\mathcal{Q}1$) Can multi-objective learning simultaneously optimise several fairness measures without sacrificing accuracy? ($\mathcal{Q}2$) Can we obtain a group of diverse models by applying multi-objective learning? ($\mathcal{Q}3$) Can multi-objective learning improve all fairness measures including those not used in model training? ($\mathcal{Q}4$) Can multi-objective learning generate an ensemble model combined from base models to balance accuracy and multiple fairness measures?

The novel contributions beyond~\cite{QingquanFair2021} are as follows:
\begin{itemize}
    \item[(i)] We apply a multi-objective evolutionary learning framework~\cite{QingquanFair2021} to train fairer ML models. Multi-objective learning is applied to consider model accuracy and multiple fairness measures simultaneously during training. We have implemented our framework in two scenarios. Empirical results and comprehensive analyses on eight well-known benchmark datasets reveal that our framework can simultaneously optimise the accuracy and multiple fairness objectives  (up to eight different fairness metrics).
    \item[(ii)] The obtained models can act as good candidates for human decision-makers' use with different preferences and can be fully utilised to create an ensemble with a good trade-off between accuracy and fairness.
    \item[(iii)] Our framework can improve fairness according to a broad range of fairness metrics, including those not used in our multi-objective learning algorithms. The robustness of our proposed multi-objective learning approach has been shown in the following sense: our learned model performed well not only on the accuracy and the eight fairness metrics used in the training, it also performed well according to other eight fairness metrics that were never used in training. In other words, our model is very robust against different fairness metrics used to assess it.
\end{itemize}

The remainder of this paper is organised as follows. 
Section \ref{sec:background} presents the background.
The framework and the designed algorithm based on this framework are presented in Section \ref{sec:framework}. Section \ref{sec:xp} shows and discusses the empirical study. Section \ref{sec:conclusion} concludes the paper.

\section{Background}\label{sec:background}
In this section, we first introduce the definitions of different fairness metrics and the relationship among the metrics. 
Then, existing approaches in mitigating unfairness, including multi-objective learning and ensemble methods, are presented.

\subsection{Measuring Fairness in Machine Learning}\label{sec:fair_indicator}

Many fairness measures have been defined to measure (un)fairness~\cite{caton2020fairness} from the perspectives of ethics in the context of fairness. While some measures are positively correlated~\cite{anahideh2021choice}, others are in conflict with each others~\cite{caton2020fairness}. So far, no one fairness measure is accepted as a universal notion of fairness quantification since different perspectives of fairness can be interpreted into different quantitative definitions~\cite{caton2020fairness,hutchinson201950,verma2018fairness,mehrabi2021survey,whittlestone2019role}. Generally speaking, existing fairness measures can be divided into two conflicting but complementary categories~\cite{speicher2018unified,caton2020fairness}: individual fairness and group fairness. Individual fairness means that similar individuals should be treated similarly, while the group fairness considers different groups relying on insensitive attributes. Typically, sensitive (also called protected) attributes are traits considered to be discriminative by law, such as gender, race, age, and so on. 

From the perspective of economic and social welfare, the work~\cite{speicher2018unified} proposed to quantify individual unfairness ($f_I$) and group unfairness ($f_G$) using inequality indices, namely generalised entropy indices. Specifically, they use each prediction result of an algorithm to calculate the corresponding benefit and measure the degree of inequality of all the benefits at the individual and group levels, respectively, formulated as~\cite{speicher2018unified}:
\begin{equation}\label{eq:individual}\small
    f_{I}=\frac{1}{n\alpha(\alpha-1)} \sum_{i=1}^{n} \left [ \left( \frac{b_i}{\mu}  \right)^{\alpha}-1  \right] ,
\end{equation}
\begin{equation}\label{eq:group}\small
    f_{G}=\frac{1}{n\alpha(\alpha-1)} \sum_{g=1}^{|G|} n_g\left[ \left(\frac{\mu_{g}}{\mu}\right)^{\alpha}-1 \right],
\end{equation}
where $|G|$ is the number of groups, $n_g$ refers to the size of group $g$ (e.g., male, female), $n$ is the number of observations (i.e., $n=\sum_{g} n_g$), and $\alpha$ is a positive constant. 

In Eqs. \ref{eq:individual} and \ref{eq:group}, $\mu$ is the mean value of all the $b_i$, whereas $\mu_g$ is the mean value of $b_i$ in group $g$. Speicher \emph{et al.}~\cite{speicher2018unified} quantified the notion of \emph{benefit vector} as $b_i=\theta x_i - y_i+1$, where $y_i$, $\theta$ and $x_i$ denote the true labels, parameters of models and input data, respectively. The aim of introducing $b_i$ is to map the algorithmic outcomes $\theta x_i$ to a scalar value based on its true label $y_i$, which can capture the desirability~\cite{speicher2018unified} of the predictive outcomes for input data $x_i$. In other words, $b_i$ indicates how much benefit the data $x_i$ receives according to the algorithmic outcomes. In \cite{speicher2018unified}, $\theta x_i - y_i + 1$ is applied as the benefit assignment rule. Then, $f_I$ (Eq. \ref{eq:individual}) and $f_G$ (Eq. \ref{eq:group}) can quantify inequality (unfairness) of the benefits $b_i$ based on all the considered data. $f_I$ aims to capture the inequality degree of each $x_i$, whereas $f_G$ captures the potential inequality among subgroups of each $x_i$. The study~\cite{speicher2018unified} has empirically and theoretically shown that $f_I$ and $f_G$ are conflicting but complementary in real-world problems. Many good properties of $f_I$ and $f_G$ in quantifying unfairness, such as anonymity, population invariance, transfer principle/Pigou-Dalton principle, zero-normalization, and subgroup decomposability were also investigated~\cite{speicher2018unified}. 

There are a variety of group fairness metrics, including parity-based metrics~\cite{caton2020fairness}, calibration-based metrics, score-based metrics, and confusion matrix-based metrics. Parity-based metrics are usually concerned about the predicted positive rates over each group, such as statistical parity. Both calibration-based and score-based metrics consider a predictive probability or score rather than predictive values. Confusion matrix-based metrics have attracted much attention recently. Table~\ref{tab:Fairness_metrics} summarises 16 metrics belonging to this category, where $G$, $y$, and $\hat{y}$ denote sensitive attributes, true labels, and predicted labels obtained by learning models, respectively. The work~\cite{anahideh2021choice} analysed the correlations among Fair1--Fair16 metrics based on the prediction results of learning models on four datasets. They conclude from the obtained correlations that Fair1--Fair8 metrics are the representative fairness metrics among Fair1--Fair16 and can represent Fair1--Fair16~\cite{anahideh2021choice}. More specifically, Fair4 can represent Fair10, Fair13, Fair14, and Fair15; Fair9, Fair11, and Fair12 can be represented by Fair2; Fair3 represents Fair16.

\subsection{Mitigating Unfairness in Machine Learning}

In the literature of mitigating unfairness, many methods were proposed to mitigate unfairness in the model training process~\cite{caton2020fairness,pessach2022review,mehrabi2021survey}. When the optimised fairness metrics are non-differentiable, many algorithms aim to make a proxy to these fairness metrics~\cite{zafar2017fairness, celis2019classification, Goh2016Satisfying,zhang2018mitigating}. For example, regarding equalised odds fairness metric, studies in~\cite{zafar2017fairness} used a proxy as a constraint to the objective function. The study~\cite{Goh2016Satisfying} applied the ramp loss to constrain non-convex optimisation to optimise disparate impact fairness metric. The work~\cite{zhang2018mitigating}, focusing on statistical parity, equalised odds or equality of opportunity, constructed an adversarial model to detect and mitigate unfairness of the predictor model through an adversarial learning strategy. There are also other types of algorithms, such as bandits~\cite{ensign2018decision} and  causal inference~\cite{kusner2017counterfactual}.

The fairness metrics can also be directly treated as objectives. Then, the problem of mitigating multiple unfairness metrics is considered as multi-objective optimisation problems~\cite{Geden_Andrews_2021, wu2021multifr,QingquanFair2021}. For example, Geden and Andrews~\cite{Geden_Andrews_2021} focused on three hiring problems and investigated the performance of different many-objective evolutionary optimisation methods for fair, interpretable and legally compliant hiring. Wu \emph{et al.}~\cite{wu2021multifr} used a weighted sum approach to combine several objectives into one. Only one model with a pre-defined set of weights was obtained. The study~\cite{QingquanFair2021} proposed a framework based on multi-objective evolutionary learning to balance accuracy and multiple fairness metrics and then verified that the obtained model set had a good performance in terms of diversity and convergence.

\begin{table}[htbp]
  \centering
  \caption{Summary of 16 fairness metrics~\cite{anahideh2021choice}, Fair1--Fair16 in terms of fairness notion and formulation.}
\begin{adjustbox} {max width=\linewidth}
    \begin{tabular}{lll}
    \toprule
    ID &Fairness Notion  & Formulation    \\
    \midrule
    Fair1 & Average Odd Difference & \tabincell{c}{$0.5*|P(\hat{y}=1|y=0,G=g_1)+P(\hat{y}=1|y=1,G=g_1) -$\\$P(\hat{y}=1|y=0,G=g_2)-P(\hat{y}=1|y=1,G=g_2)|$}    \\
    Fair2 & Error Difference & $|P(\hat{y}\neq y, G=g_1)-P(\hat{y}\neq y, G=g_2)|$   \\
    Fair3 & Discovery Ratio &   $\frac{P(y=0|\hat{y}=1, G=g_1)}{P(y=0|\hat{y}=1, G=g_2)}$\\
    Fair4 & Predictive Equality  & $|P(\hat{y}=1|y=0, G=g_1)-P(\hat{y}=1|y=0, G=g_2)|$   \\
    Fair5 & FOR difference & $|P(y=1|\hat{y}=0, G=g_1)-P(y=1|\hat{y}=0, G=g_2)|$   \\
    Fair6 & FOR Ratio & $\frac{P(y=1|\hat{y}=0, G=g_1)}{P(y=1|\hat{y}=0, G=g_2)}$      \\
    Fair7 & FNR Difference  & $|P(\hat{y}=0|y=1, G=g_1)-P(\hat{y}=0|y=1, G=g_2)|$    \\
    Fair8 & FNR Ratio & $\frac{P(\hat{y}=0|y=1, G=g_1)}{P(\hat{y}=0|y=1, G=g_2)}$    \\
    \midrule
    Fair9 & Error Ratio & $\frac{P(\hat{y}\neq y, G=g_1)}{P(\hat{y}\neq y, G=g_2)}$  \\
    Fair10 & Discovery Difference & $|P(y=0|\hat{y}=1, G=g_1)-P(y=0|\hat{y}=1, G=g_2)|$    \\
    Fair11 & FPR Ratio & $\frac{P(\hat{y}=1|y=0, G=g_1)}{P(\hat{y}=1|y=0, G=g_2)}$    \\
    Fair12 & Disparate Impact & $\frac{P(\hat{y}=1|, G=g_1)}{P(\hat{y}=1|, G=g_2)}$    \\
    Fair13 & Statistical Parity & $|P(\hat{y}=1|, G=g_1)-P(\hat{y}=1|, G=g_2)|$    \\
    Fair14 & Equal Opportunity & $|P(\hat{y}=1|y=1, G=g_1) - P(\hat{y}=1|y=1, G=g_2)|$    \\
    Fair15 & Equalised Odds & \tabincell{c}{$0.5*(|P(\hat{y}=1|y=0,G=g_1)-P(\hat{y}=1|y=0,G=g_2)|+$ \\$|P(\hat{y}=1|y=1,G=g_1)-P(\hat{y}=1|y=1,G=g_2)|)$ }     \\
    Fair16 & Predictive Parity & $|P(y=1|\hat{y}=1, G=g_1)-P(y=1|\hat{y}=1, G=g_2)|$    \\
    \bottomrule
    \end{tabular}%
   \end{adjustbox}
  \label{tab:Fairness_metrics}%
\end{table}%

Ensemble methods have also been used in the context of mitigating unfairness~\cite{Iosifidis2019FAE,cispa2382,Kenfack2021Impact}. In dealing with class-imbalance tasks, \cite{Iosifidis2019FAE} proposed an ensemble framework at both pre- and post- processing interventions to tackle discrimination class-imbalance tasks in ML. Study~\cite{cispa2382} claimed that an ensemble consisting of randomly selected classifiers is able to behave more fairly than a single classifier in many cases. The recent work~\cite{Kenfack2021Impact} trained different base classifiers to maximise accuracy and determined the weights of base classifiers based on their performance of accuracy and fairness metrics and manually set weights among accuracy and fairness metrics. Then, predicted outcomes are produced through the weighted majority voting method~\cite{Kenfack2021Impact}. Empirical results showed its weight assignment method can have better performance than \cite{cispa2382}. However, using either the random selection~\cite{cispa2382} or weight assignment~\cite{Kenfack2021Impact} has a limited ability to enhance ensemble diversity. During base model training, neither of  \cite{cispa2382,Kenfack2021Impact} considers fairness metrics, resulting in the lower diversity of fairness metrics among ensemble individuals. Multi-objective evolutionary learning can provide the potential strengths to overcome the challenges that the works \cite{cispa2382,Kenfack2021Impact} face since fairness metrics are considered during model training.

\section{Multi-objective Evolutionary Learning and Ensemble Learning for Mitigating Unfairness}\label{sec:framework} 

This section describes the framework of multi-objective evolutionary learning and multi-objective ensemble learning in mitigating unfairness. Then, the details of our designed algorithms based on the framework are provided.

\subsection{Multi-objective Learning Framework for Fairer ML}\label{sec:sub_framework}

Our general framework is presented in Algorithm \ref{algo:framework}, aiming to evolve a population of learning models. Every individual of the population is a fair learning model, e.g., an artificial neural net (ANN). During the evolution, we expect that the population can gradually achieve better tradeoffs among accuracy and multiple fairness measures.

\begin{algorithm}[htbp]
\caption{\label{algo:framework}Multi-objective learning framework.}
\begin{algorithmic}[1]
\Require Initial models $\M_1,\dots,\M_\lambda$, set of model evaluation criteria $\mathcal{E}$, training dataset $\mathcal{D}_{train}$, validation dataset $\mathcal{D}_{validation}$, multi-objective optimiser $\pi$
\Ensure A final model set $\M_1,\dots,\M_\lambda$
\State Partially train~\cite{yao1997new,yao1999evolving} $\M_1,\dots,\M_\lambda$ over $\mathcal{D}_{train}$ \label{line:partial1}
\For{$i \in \{1,\dots,\lambda\}$}
    \State{${\epsilon}_i \leftarrow$ Evaluate ${\M}_i$ with criteria $\mathcal{E}$ on $\D_{validation}$}
\EndFor
\While{terminal conditions are not fulfilled}
    \State $\mathcal{P} \leftarrow$ Select $\mu$ promising models from $\M_1,\dots,\M_\lambda$ \NoNumber{with ``best''  $\epsilon_1,\dots,\epsilon_\mu$ according to $\pi$} \label{line:parentselection}
    \State $\M' \leftarrow$ Generate $\phi$ new models $\M_1',\dots, \M_\phi'$ from $\mathcal{P}$ \NoNumber{according to $\pi$} \label{line:generateoff}
\For{$i \in \{1,\dots,\phi\}$}
    \State{${\M'}_i\leftarrow$ Partially train~\cite{yao1997new,yao1999evolving} $\M'_i$ on $\D_{train}$} \label{line:partial2}
    \State{${\epsilon'}_i \leftarrow$ Evaluate ${\M'}_i$ with criteria $\mathcal{E}$ on $\D_{validation}$}
\EndFor
    \State{$<\M_1,\epsilon_1>,\dots, <\M_\lambda,\epsilon_\lambda> \leftarrow$ Select $\lambda$ promising \NoNumber{models from $\{\M_1,\dots,\M_\lambda\} \bigcup \{\M_1',\dots, \M_\phi'\}$} by $\pi$} \NoNumber{based on $\epsilon_1,\dots,\epsilon_\lambda$ and $\epsilon'_1,\dots,\epsilon'_\phi$, and then update} \NoNumber{$\M_1,\dots,\M_\lambda$ and $\epsilon_1,\dots,\epsilon_\lambda$ accordingly} \label{line:envselection}
\EndWhile
\end{algorithmic}
\end{algorithm}

The inputs to our framework include a number of initial models $\M$ as a population, a set of model evaluation criteria $\mathcal{E}$, a set of training data $\mathcal{D}_{train}$, a set of validation data $\mathcal{D}_{validation}$ and a multi-objective optimiser $\pi$. More specifically, criteria $\mathcal{E}$ are used to calculate the optimised objective values (e.g., accuracy, fairness metrics in Table \ref{tab:Fairness_metrics}) according to the predictions of models $\M$ on validation data $\mathcal{D}_{validation}$. Training data $\mathcal{D}_{train}$ is used for local search strategies (e.g., partial training~\cite{yao1997new,yao1999evolving}) to update parameters of models $\M$. 

A multi-objective optimiser $\pi$ mainly contains three strategies, reproduction, mating selection, and survival selection. In our framework, every time a new model is initialised or generated (lines \ref{line:partial1} and 9 in Algorithm \ref{algo:framework}), partial training~\cite{yao1997new,yao1999evolving} is always adopted on $\mathcal{D}_{train}$. The objective values of each model are obtained through criteria $\mathcal{E}$. In the main loop, first, the mating selection strategy of $\pi$ is applied to select $\mu$ promising models as parent models $\mathcal{P}$ (line \ref{line:parentselection} in Algorithm \ref{algo:framework}). Next, $\lambda$ new models as ${\M}'$ are created with the aim of inheriting information from $\mathcal{P}$ (line \ref{line:generateoff} in Algorithm \ref{algo:framework}) through the reproduction strategy of $\pi$. Specifically, the reproduction strategy is applied to generate new models as offspring by modifying the parameters of parent models, where crossover and mutation are two widely used operators. After partial training and model evaluation (lines 9-10 in Algorithm \ref{algo:framework}), $\lambda$ candidate models are selected from the combination of $\M$ and $\mathcal{M}'$ by the survival selection of $\pi$ as new $\M$ for the next generation (line \ref{line:envselection} in Algorithm \ref{algo:framework}). The above steps repeat until a termination criterion is reached.

The core steps of our framework are the model evaluation based on multiple criteria and generation of new models. Multi-objective evolutionary algorithms (MOEAs)~\cite{li2015many} as $\pi$ are ideal to generate a learning model set with better convergence and diversity. The output models can be further selected by decision-makers or used as an ensemble~\cite{chandra2006ensemble,chen2010multiobjective,onan2017hybrid}.

\subsection{Multi-objective Ensemble Learning Framework for Fairer ML}\label{sec:ensembleframework}

When adopting ensemble learning, the final model set obtained by Algorithm \ref{algo:framework} can be used to construct an ensemble. Algorithm \ref{algo:ensembleframework} shows our proposed multi-objective ensemble learning framework. The inputs of our ensemble framework include trained models $\M_1,\dots,\M_\lambda$ obtained by Algorithm \ref{algo:framework}, a set of model evaluation criteria $\mathcal{E}$, an ensemble training dataset $\mathcal{D}_{ensemble}$ and a multi-objective ensemble selection strategy $\pi_{ens}$. First, the objective values of $\M_1,\dots,\M_\lambda$ are computed through model evaluation criteria $\mathcal{E}$ on the ensemble training dataset $\mathcal{D}_{ensemble}$. Then, various model selection strategies can be applied to select a subset of models from $\mathcal{M}_1,\dots,\mathcal{M}_\lambda$, denoted as $\M_1,\dots,\M_e$, according to $\pi_{ens}$ and the obtained objective values.

\begin{algorithm}[t]
\caption{\label{algo:ensembleframework}Multi-objective ensemble learning framework.}
\begin{algorithmic}[1]
\Require Trained models $\M_1,\dots,\M_\lambda$ obtained by Algorithm~\ref{algo:framework}, set of model evaluation criteria $\mathcal{E}$, ensemble training dataset $\mathcal{D}_{ensemble}$, multi-objective ensemble selection strategy $\pi_{ens}$
\Ensure  A base model set $\M_1,\dots,\M_e$
\For{$i \in \{1,\dots,\lambda\}$}
    \State{${\epsilon}_i \leftarrow$ Evaluate ${\M}_i$  with criteria $\mathcal{E}$ on the ensemble \NoNumber{training dataset $\D_{ensemble}$}}
\EndFor
\State $\M_1,\dots,\M_e \leftarrow$ Select a subset (with size $e$) of $\M_1,\dots,\M_\lambda$ according to $\pi_{ens}$ and $\epsilon_1,\dots,\epsilon_\lambda$
\end{algorithmic}
\end{algorithm}

\subsection{Proposed Algorithms based on Our Framework}\label{sec:sub_implementation}

The choices of the model set, evaluation criteria, multi-objective optimisation algorithm and ensemble selection strategy in our proposed framework can vary according to the prediction tasks and actual preferences. We designed algorithms based on our framework using the following core ingredients.

\subsubsection{Model set} Various ML models can be used. In this work, a set of ANNs with an identical architecture are used as individuals. The weights and biases of each ANN are encoded as a real-value vector and represented as an individual~\cite{yao1997new}.

\subsubsection{Evaluation criteria} 
In this work, we consider 11 measures in total including accuracy, individual unfairness $f_I$ (Eq. \ref{eq:individual}), group unfairness $f_G$ (Eq. \ref{eq:group}), and Fair1--Fair8 (Table \ref{tab:Fairness_metrics}). Cross entropy ($CE$) is widely used to measure the accuracy of classifiers and is minimised. $f_I$ and $f_G$ are also minimised~\cite{speicher2018unified}. The measures of Fair1--Fair8 based on absolute differences are minimised. For Fair1--Fair8 using ratios, taking Fair3 as an example, the objective is calculated as $1 - \min{ \{\frac{P(y=0|\hat{y}=1, G=g_1)}{P(y=0|\hat{y}=1, G=g_2)}, \frac{P(y=0|\hat{y}=1, G=g_2)}{P(y=0|\hat{y}=1, G=g_1)}} \}$, which is to be minimised. In this paper, when observing the values of Fair9--Fair16, the transformation introduced above is applied to these fairness measures. So, the optimal values of Fair1--Fair16 are all zeros.

\subsubsection{Multi-objective optimiser}
Concerning that some criteria from $\mathcal{E}$ can be directly used as loss functions to update ANNs, denoted as $\mathcal{E}_{loss}$, such as $f_I$ and $f_G$, we design an effective strategy including mating selection and reproduction strategies, shown in Algorithm~\ref{algo:reproductionframework}. Survival selection of $\pi$ can be adopted from any MOEAs. In our algorithm, we choose the survival selection of stochastic ranking algorithm (SRA)~\cite{SRA2016}, a well-known multi-indicator-based MOEA. SRA uses the stochastic ranking~\cite{Runarsson2000Stochastic} to balance different search biases of different indicators and achieved the superior performance in dealing with many-objective optimisation problems. The proposed strategy aims to better balance exploration (lines 4-14 in Algorithm~\ref{algo:reproductionframework}) and exploitation (lines 15-22 in Algorithm~\ref{algo:reproductionframework}).

\begin{algorithm}[htbp]
\caption{\label{algo:reproductionframework}Selection and reproduction strategy.}
\begin{algorithmic}[1]
\Require Current population $\M_1,\dots,\M_\lambda$, set of model evaluation criteria $\mathcal{E}$, training dataset $\mathcal{D}_{train}$, validation dataset $\mathcal{D}_{validation}$, multi-objective optimiser $\pi$, parameter $K$ for extreme models in partial training~\cite{yao1997new,yao1999evolving}
\Ensure An offspring model set $\s$
\State $\mathcal{E}_{loss} \leftarrow$ Select from $\mathcal{E}$ the criteria that can be directly used as loss functions to update the parameters of models
\State $m$ = $|\mathcal{E}_{loss}|$
\State Initialise the offspring model set $\s=\emptyset$
\Statex $//$ \textit{Improve exploration ability in each $\mathcal{E}_{loss}$}
\State $\mathcal{S}_{best} \leftarrow$ Select $m$ models from $\M_1,\dots,\M_\lambda$ with the best separately on each criteria $\mathcal{E}_{loss}$ \label{line:parentselection1}
\State $\s' \leftarrow$ Generate new models $\M_1',\dots, \M_m'$ from $\mathcal{S}_{best}$ according to reproduction strategy of $\pi$
\For{$i \in \{1,\dots,m\}$}
\State $k=0$   
   \Repeat
    \State{$\M \leftarrow$ Partially train~\cite{yao1997new,yao1999evolving} the $i$-th model of \NoNumber{$\s'$ on $\D_{train}$, where the loss is calculated based } \NoNumber{ on the corresponding criterion $\mathcal{E}_{loss}^{i}$}} 
    \State{${\epsilon'} \leftarrow$ Evaluate $\M$ on $\D_{validation}$ with criteria $\mathcal{E}$}
    \State Update $\s\leftarrow\s \cup \{<\M, {\epsilon'}>\}$
    \State $k = k + 1$
     \Until $k=K$
\EndFor
\Statex $//$ \textit{Improve exploitation ability in each $\mathcal{E}_{loss}$}
\State $\kappa = \lambda - m*K$
\State $\mathcal{S}_{\pi} \leftarrow$ Select $\kappa$ promising models from $\M_1,\dots,\M_\lambda$ with ``best''  $\epsilon_1,\dots,\epsilon_\kappa$ according to the mating selection of $\pi$ \label{line:parentselection2}
\State $\s''\leftarrow$ Generate $\lambda$ new models $\M_{1}',\dots, \M_{\lambda}'$ from $\mathcal{S}_{\pi}$ according to reproduction strategy of $\pi$ 
\For{$i \in \{1,\dots,\lambda\}$}
    \State {${\M}\leftarrow$ Partially train~\cite{yao1997new,yao1999evolving} the $i$-th model of $\s''$ \NoNumber{ on $\D_{train}$, where the loss is randomly selected} \NoNumber{ from criteria $\mathcal{E}_{loss}$}}
    \State {${\epsilon'} \leftarrow$ Evaluate $\M$ with criteria $\mathcal{E}$ on $\D_{validation}$}
    \State Update $\s\leftarrow \s \cup \{<\M, {\epsilon'}>\}$
\EndFor
\end{algorithmic}
\end{algorithm}

For the exploration improvement part, $m$ best models according to each criteria $\mathcal{E}_{loss}$ are selected from $\M_1,\dots,\M_\lambda$ and denoted as $\mathcal{S}_{best}$. After applying the reproduction strategy to $\mathcal{S}_{best}$, partial training~\cite{yao1997new,yao1999evolving} is performed to each of $\mathcal{S}_{best}$ for $K$ times, where the loss is the same as the corresponding criterion $\mathcal{E}^i_{loss}$ (line 9 in Algorithm~\ref{algo:reproductionframework}). All the generated models are stored in the set $\s$. For the exploitation improvement part, $\kappa$ promising models are selected from $\M_1,\dots,\M_\lambda$ based on SRA's mating selection strategy and denoted as $\mathcal{S}_{\pi}$. Next, $\lambda$ new models are generated after the reproduction strategy is applied to $\mathcal{S}_{\pi}$, where $\kappa$ is equal to $\lambda - m*K$. Then, partial training is performed to each of $\mathcal{S}_{\pi}$, where the loss is randomly selected from criteria $\mathcal{E}_{loss}$. 

Both crossover and mutation are applied in the reproduction strategy of our algorithms. When mutation, isotropic Gaussian perturbation~\cite{minku2013software} is performed, formulated as $r_i = r_i + \delta_i$, where $r_i$ is the $i$-th weight of an ANN, isotropic Gaussian perturbation $\delta_i \sim \mathcal{N}(0,\sigma^2)$ and $\sigma$ is the mutation strength. Given parents $p$ and $q$, the variant of weight crossover~\cite{gong2018multiobjective} is applied and defined as $r_i^{o_1} =u_i r_i^p  +  (1 - u_i)r_i^q$ and $r_i^{o_2} = u_i r_i^q  +  (1 - u_i)r_i^p$, where $u_i$ is the uniformly random value in $[0,1]$, $r_i^{p}$, $r_i^{q}$, $r_i^{o_1}$ and $r_i^{o_2}$ are $i$-th weight of parent $p$, parent $q$, offspring $o_1$ and offspring $o_2$, respectively.

In regard to partial training~\cite{yao1997new,yao1999evolving}, model parameters are updated by stochastic gradient descent (SGD) optimiser~\cite{ruder2016overview}.

\subsubsection{Multi-objective ensemble selection strategy}\label{sec:ensemblestrategy}

When implementing our ensemble learning framework shown in Algorithm \ref{algo:ensembleframework}, four algorithms using different multi-objective ensemble selection strategies~\cite{minku2013nalysis,zhang2015Knee,wang2015twoarch2} are implemented, referred to as EnsAll, EnsBest, EnsKnee and EnsDiv. In EnsAll, all the non-dominated models are selected~\cite{minku2013nalysis}, while in EnsBest, only the best models from the non-dominated models are selected according to their performance~\cite{minku2013nalysis}. In EnsKnee, a knee point (model) subset from the non-dominated models is selected according to the strategy of finding a knee point subset in \cite{zhang2015Knee}. In EnsDiv, a diverse model subset from the non-dominated models is selected according to the selection method in an overflowed diversity archive in the work of \cite{wang2015twoarch2}.

\section{Experimental Studies}\label{sec:xp}
In this section, two studies are used to answer $\mathcal{Q}1$-$\mathcal{Q}4$ through extensive experiments. First, the overview of the two studies is introduced in Section \ref{sec:overview}, including the motivation and details of the two studies. Then, $\mathcal{Q}1$-$\mathcal{Q}2$, and $\mathcal{Q}3$-$\mathcal{Q}4$ will be answered in Section \ref{sec:Q1_Q2} and Section \ref{sec:Q3_Q4}, respectively.

\subsection{An Overview of the Two Studies}\label{sec:overview}
To adequately answer $\mathcal{Q}1$-$\mathcal{Q}4$, two studies are adopted, formulated as tri- and 9-objective optimisation problems.

To achieve a comprehensive investigation of $\mathcal{Q}1$ and $\mathcal{Q}2$, we will compare the performance of the methods that are based on our framework but considering different measures. For convenience, we only consider two unfairness metrics, but the conclusion can be generalised to any case with more than two unfairness measures. To answer $\mathcal{Q}1$ and $\mathcal{Q}2$, the tri-objective case is considered, where the evaluation criteria involve the cross entropy, individual unfairness~\cite{speicher2018unified} and group unfairness~\cite{speicher2018unified}, introduced in Section \ref{sec:fair_indicator}.

\begin{table}[htbp]
  \centering
  \caption{Datasets used in our study. ``Sensitive'' indicates the sensitive attributes and ``$|G|$'' (as in Eq. \ref{eq:group}) is the number of groups.}  
  \setlength{\tabcolsep}{2pt}
    \begin{tabular}{lccl}
    \toprule
    Dataset  & Sensitive & $|G|$ & Description of Prediction Task  \\
    \midrule
    \emph{Student}    & Gender, Age & 8     & Whether a student will pass the exam\\
    \emph{German}   & Gender, Age & 4     & Whether a person has an acceptable credit risk\\
    \multirow{2}{*}{\emph{COMPAS}}   & Gender & \multirow{2}{*}{4}     & Whether an arrested offender will be rearrested  \\
    & Race &  &   within two years counting from taking the test\\
    \emph{LSAT}   & Gender, Race & 16    & Whether a student will pass the exam\\
    \emph{Default} & Gender & 2     & Whether a customer will default on payments \\
    \multirow{2}{*}{\emph{Adult}}  & Gender & \multirow{2}{*}{10}    & Whether a person can get income higher \\
     & Race &    & than \$50,000 per year\\
    \multirow{2}{*}{\emph{Bank}}   & \multirow{2}{*}{Age}   & \multirow{2}{*}{2}     & Whether a client will subscribe to a term\\
    &    &      & deposit service \\
    \multirow{2}{*}{\emph{Dutch}}  & \multirow{2}{*}{Gender}   & \multirow{2}{*}{2}     & Whether a person has a highly prestigious  \\
    &    &  &  occupation \\
    \bottomrule
    \end{tabular}%
  \label{tab:dataset}%
\end{table}%

To answer $\mathcal{Q}3$ and $\mathcal{Q}4$, the 9-objective case that simultaneously optimises the cross entropy and Fair1--Fair8 (Table \ref{tab:Fairness_metrics}) is considered. The conclusion of \cite{anahideh2021choice} can be directly utilised in answering $\mathcal{Q}3$, indicating that the metrics Fair1--Fair8 can well represent Fair1--Fair16. Then, we make full use of the final population to construct an ensemble to answer $\mathcal{Q}4$.

\subsection{Answering $\mathcal{Q}1$ and $\mathcal{Q}2$}\label{sec:Q1_Q2}

In this section, the experimental results of the tri-objective case are used to answer both $\mathcal{Q}1$ and $\mathcal{Q}2$.

\subsubsection{Compared methods} We denote our framework as $F_{*}$, where $*$ means the optimised objectives. So the tri-objective case is $F_{EIG}$, where $E$, $I$, $G$ refer to the cross entropy $CE$, individual unfairness $f_I$ and group unfairness $f_G$, respectively. Three ablation studies are performed, which are all based on our framework $F_{*}$ but consider one or two measures $*$. What's more, we compare $F_{EIG}$ with the state-of-the-art algorithm Multi-FR~\cite{wu2021multifr} that does not use an MOEA. More specifically, four variants of Multi-FR according to the method of gradient normalisation are considered. Although the work of~\cite{wu2021multifr} pointed out that the normalisation of the gradient is optional, the gradient normalisation method can affect the performance of Multi-FR significantly. Three widely used normalisation methods~\cite{Sener2018Multi} are applied in our experiments.

The seven compared methods are summarised as follows. $F_{EI}$ is the bi-objective case that considers both $CE$ and individual unfairness $f_I$ based on our framework. $F_{EG}$ is another bi-objective case that considers both $CE$ and group unfairness $f_G$ based on our framework. $F_{E}$ is a single objective case that only considers $CE$. Multi-FR-$l_2$, Multi-FR-$loss$, and Multi-FR-$loss+$ are variants of Multi-FR approach~\cite{wu2021multifr} using $l_2$ normalisation, $loss$ normalisation, and $loss+$ normalisation, respectively. Multi-FR-no-norm refers to the Multi-FR approach without gradient normalisation.

\subsubsection{Datasets}
Eight well-known benchmark datasets widely used in the literature of algorithmic fairness~\cite{pessach2020algorithmic}, namely
\emph{Student}~\cite{cortez2008using,Kearns2019An},
\emph{German}~\cite{kamiran2009classifying}, \emph{COMPAS}~\cite{compasdataset},  \emph{LSAT}~\cite{sander2004systemic},  \emph{Default}~\cite{YEH20092473},
\emph{Adult}~\cite{adultdataset},
\emph{Bank}~\cite{zafar2017fairness2} and
\emph{Dutch}~\cite{kamiran2012data},
are used in our experimental study. Table \ref{tab:dataset} summarises these datasets. The pre-processing on \emph{German}, \emph{COMPAS} and \emph{Adult} datasets is the same as in~\cite{friedler2019comparative}. Each dataset is randomly split into 3 partitions, with a ratio of 6:2:2, as the training, validation, and test sets. The sensitive features of each dataset in Table \ref{tab:dataset} are all considered in calculating $f_G$. The difficulty of $f_G$ being optimised increases as the value of $|G|$ increases.

\subsubsection{Parameter setting}

\begin{table}[htbp]
\centering
  \caption{Settings of the 3-objective algorithm in terms of mutation strength and batch size in answering $\mathcal{Q}1$ and $\mathcal{Q}2$.}
\begin{tabular}{ccccc}
\toprule
                  & \emph{Student} & \emph{German} & \emph{COMPAS} & \emph{LSAT} \\
\midrule     
Mutation strength & 0.005   & 0.01   & 0.005  & 0.01 \\
Batch size        & 120     & 40     & 200    & 1000 \\
\midrule 
&\emph{Default} & \emph{Adult} & \emph{Bank} & \emph{Dutch} \\
\midrule 
Mutation strength & 0.01    & 0.005 & 0.01 & 0.01  \\
Batch size        & 1000    & 400   & 1000 & 1000 \\
\bottomrule
\end{tabular}
\label{tab:setting}%
\end{table}

All ANN models are fully connected with one hidden layer of 64 nodes. The weights are initialised as in \cite{glorot2010understanding}, which is commonly used. The learning rate is set as $0.004$ for all experiments. The gradient descent optimiser is based on the SGD~\cite{ruder2016overview}. For the experiments with algorithms $F_{*}$s, the $\mu$ and $\lambda$ in Algorithm \ref{algo:framework} are $100$. $K$ in Algorithm \ref{algo:reproductionframework} is set as 10. The $\mathcal{E}_{loss}$ in Algorithm \ref{algo:reproductionframework} contains $CE$, $f_I$ and $f_G$ since all the three objectives are differentiable and are directly used as losses. The settings of mutation strength and batch size on different datasets are shown in Table \ref{tab:setting}, where the parameter values are determined through the grid search. Since $F_{E}$ considers one single objective, the top $\lambda$ models considering $CE$ are directly selected as the new population for the next generation. The probabilities of crossover and mutation are all set as to 1. The termination condition is set as a maximum number of $200$ generations. The four variants of Multi-FR use the same batch size as in Table \ref{tab:setting}. Five-fold cross-validation is applied. For each compared method, 30 independent trials are performed.

\subsubsection{Performance measures}\label{sec:performan_Q1_Q2}
Considering that a set of models will be generated by $F_{*}$s but only one model will be created by Multi-FR, for fair and comprehensive comparisons, two groups of performance measures are introduced.

When comparing population-based algorithms including $F_{E}$, $F_{EI}$, $F_{EG}$ and $F_{EIG}$, two popular indicators~\cite{li2019quality}, hypervolume (HV)~\cite{Shang2021Survey} and coverage over Pareto front (CPF)~\cite{Tian2019Diversity}, are used to evaluate the solution set. A larger HV value indicates that the set has better performance. CPF emphasises more the diversity and a larger value indicates better diversity~\cite{li2019quality,Tian2019Diversity}. In this work, since the true Pareto front is unknown, when calculating HV and CPF, all the non-dominated solutions found in all the experimental trials considering the same objectives on the same dataset are collected as a pseudo Pareto front. After normalising solution set into the closed intervals $[0,1] \times \cdots \times [0,1]$ based on the pseudo Pareto front, $\{1.1,\dots, 1.1\}$ is set as the nadir point in HV.

To compare $F_{EIG}$ with Multi-FR, three metrics are used to consider the domination relationship~\cite{li2019quality}. Given two solutions $p$ and $q$, we denote that~\cite{li2019quality} $p \prec q$, $p \nprec q$ mean $p$ dominates $q$, and $p$ does not dominate $q$, respectively. 

Given a solution $s$ generated by Multi-FR in one trial and a set of model sets $\mathcal{P}$ obtained by $F_{EIG}$ in all $30$ trials, the metric $Dominate$ records if $F_{EIG}$ generates solutions that dominate the solution obtained by Multi-FR as follows.
\begin{equation}\label{eq:HVequ1}\small
    Dominate (P, s) = 
   \frac{1}{30} \sum_{i=1}^{30} { sign \left[ \sum_{j=1}^{|P^i|} { sign \left[P_j^i \prec s \right] } > 0 \right]},
\end{equation}
where $P^i$ is the model set in the $i$-trial, $P^i_j$ is the $j$-th model in $P^i$. $sign[ \cdot ]$ is equal to 1 if $[\cdot]$ is true, otherwise 0. The larger value means there are more trials where $s$ is dominated by some solutions of the model set obtained by $F_{EIG}$.

The metric $Incomparable$ calculates the average proportion of solutions that are obtained by $F_{EIG}$ and incomparable with $s$ over all $30$ trials of $F_{EIG}$.
\begin{equation}\label{eq:HVequ2}\small
    Incomparable (P,s) = 
    \frac{1}{30} \sum_{i=1}^{30} { \frac{1}{|P^i|} \sum_{j=1}^{|P^i|} { sign \left[(P_j^i \nprec s) \& (s \nprec P_j^i)   \right] }},
\end{equation}
A larger $Incomparable$ value means that more models obtained by $F_{EIG}$ are incomparable with $s$.

For the third metric $Dominated$, we calculate the average proportion of solutions that are obtained by $F_{EIG}$ and are dominated by $s$ over all $30$ trials of $F_{EIG}$.
\begin{equation}\label{eq:HVequ3}\small
    Dominated (P,s) = \frac{1}{30} \sum_{i=1}^{30} { \frac{1}{|P^i|} \sum_{j=1}^{|P^i|} { sign \left[s \prec P_j^i \right] }},
\end{equation}
A larger $Dominated$ value means that more models obtained by $F_{EIG}$ are dominated by the model $s$.

\noindent\textbf{($\mathcal{Q}1$) \emph{Can multi-objective learning simultaneously optimise several fairness measures?}}

We answer $\mathcal{Q}1$ from four perspectives on the \emph{test set}: (i) visualisation of optimisation process, (ii) convergence curves of HV values, (iii) HV performance of the final generation, (iv) comparison with state-of-the-art algorithm Multi-FR~\cite{wu2021multifr} according to $Dominate$, $Incomparable$ and $Dominated$.

\begin{figure}[htbp]
    \centering
    \includegraphics[width=.13\textwidth]{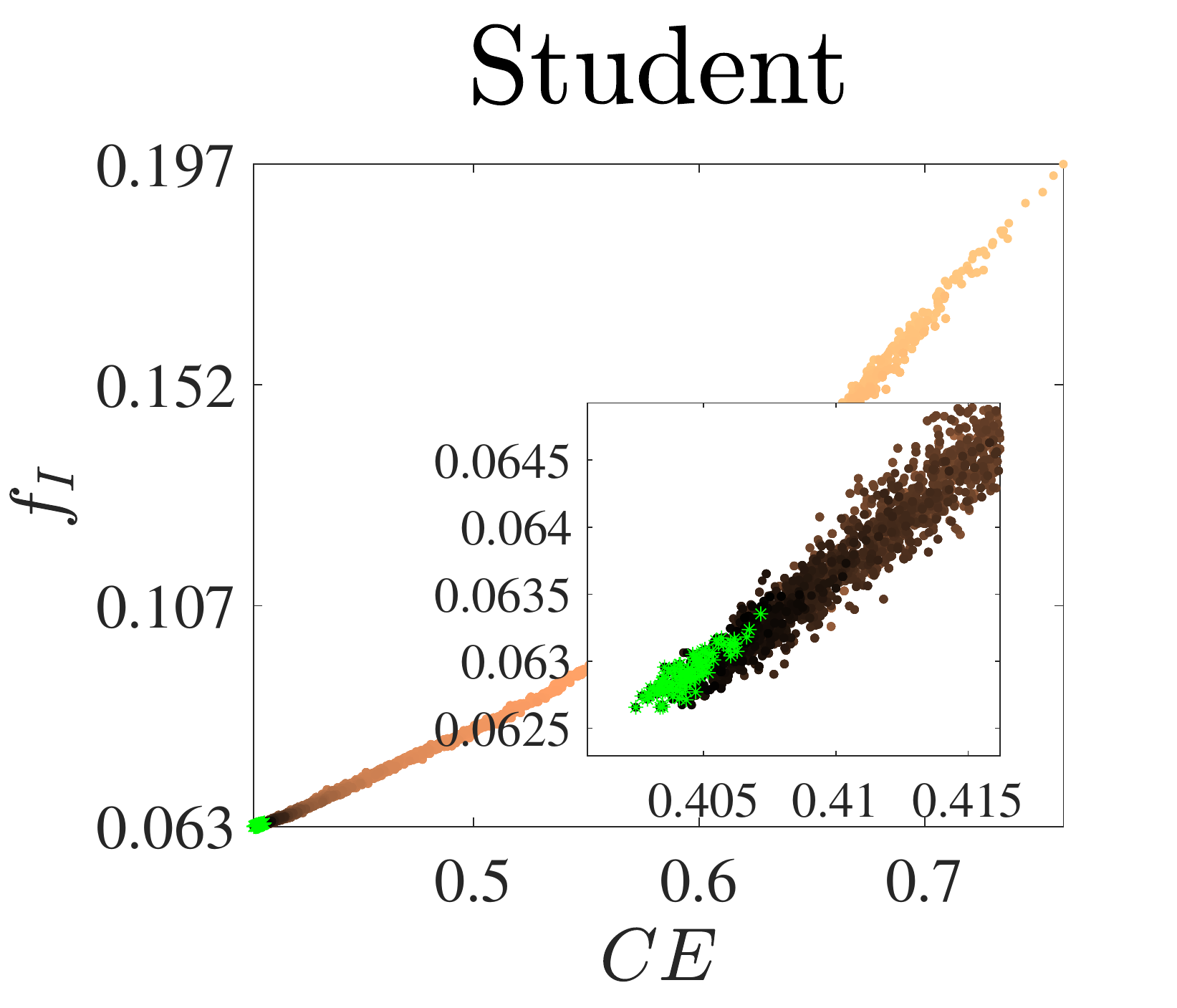}\hspace{1em}
    \includegraphics[width=.13\textwidth]{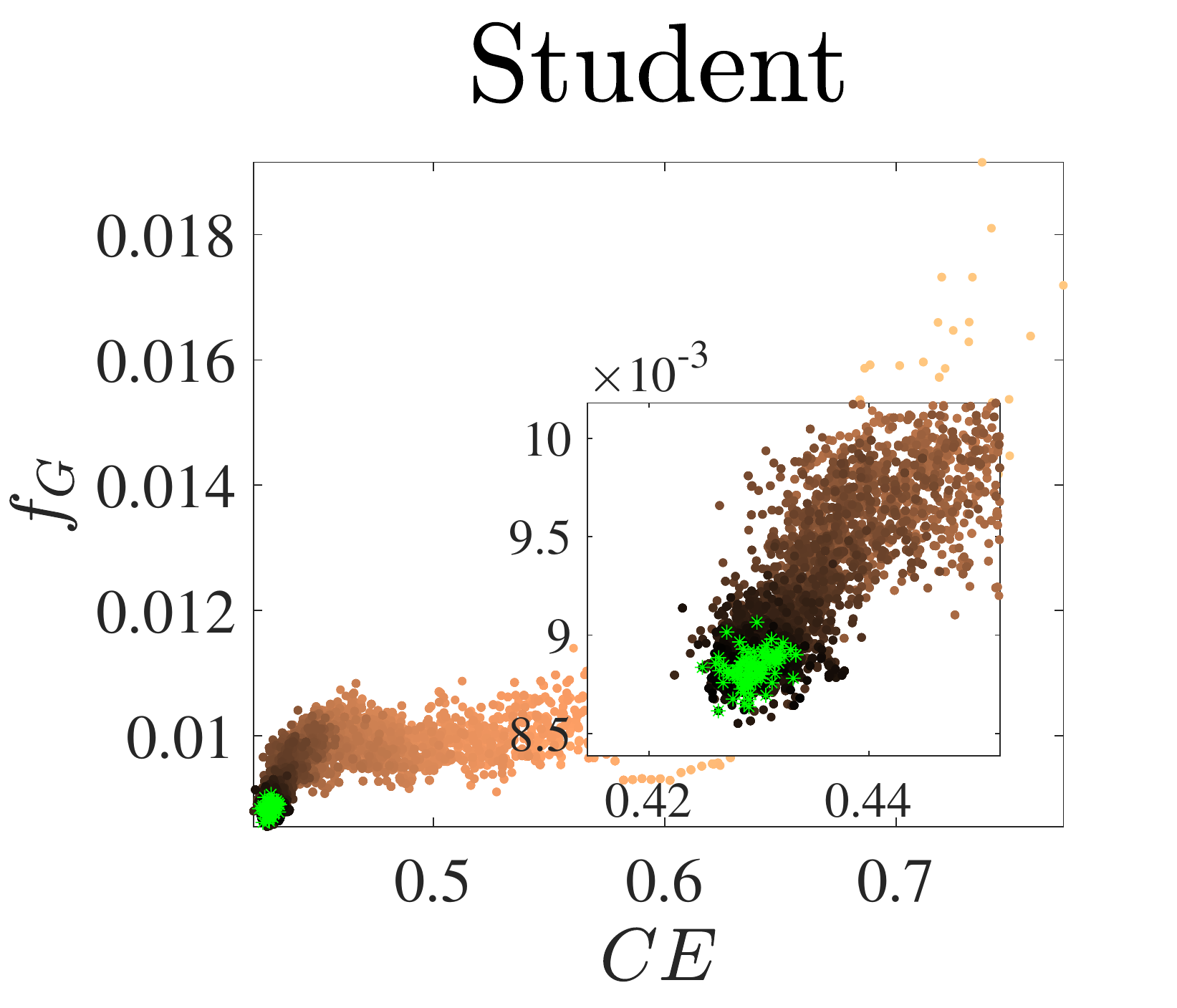}\hspace{1em}
    \includegraphics[width=.13\textwidth]{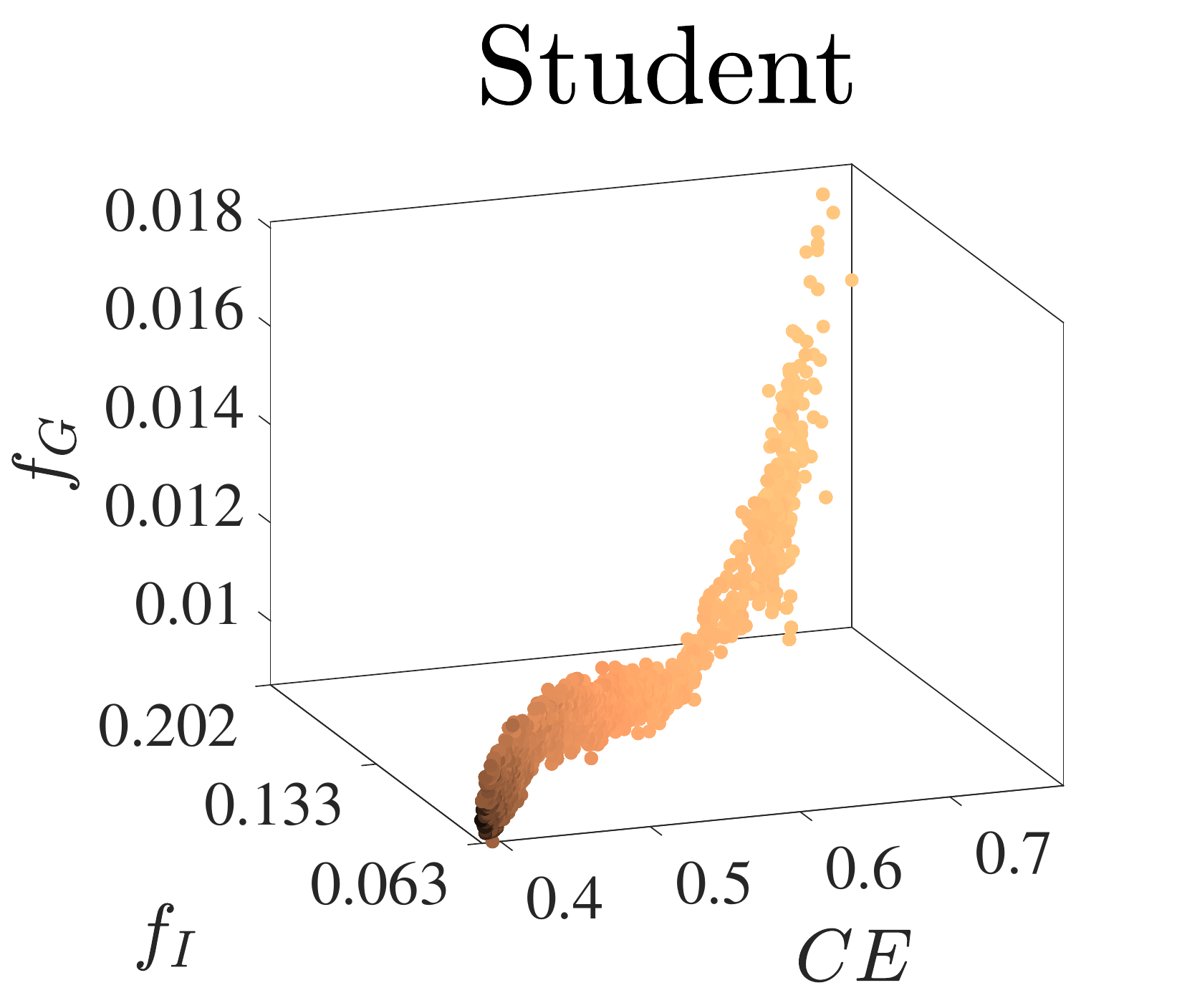}\hspace{1em}\\
    \includegraphics[width=.13\textwidth]{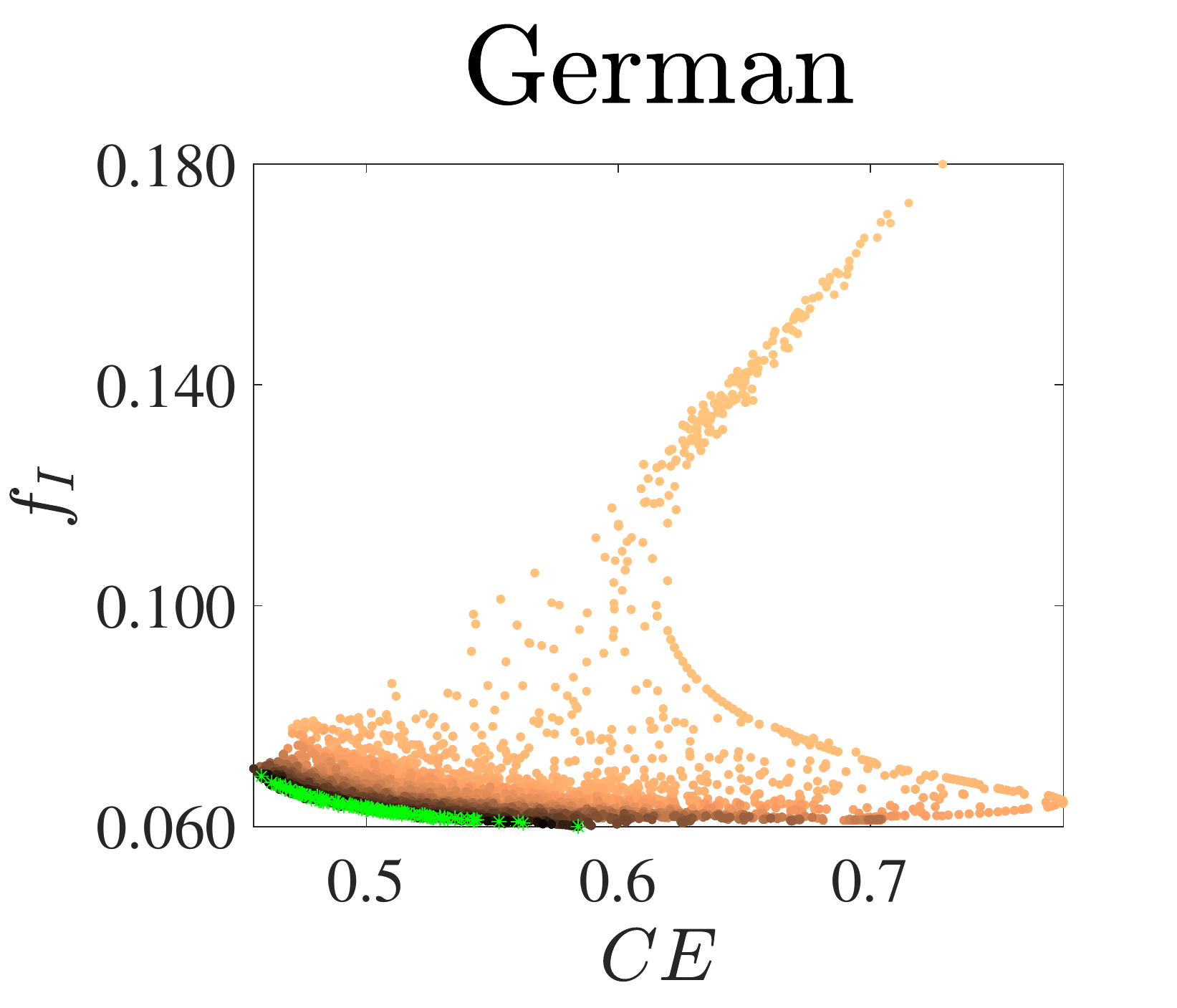}\hspace{1em}
    \includegraphics[width=.13\textwidth]{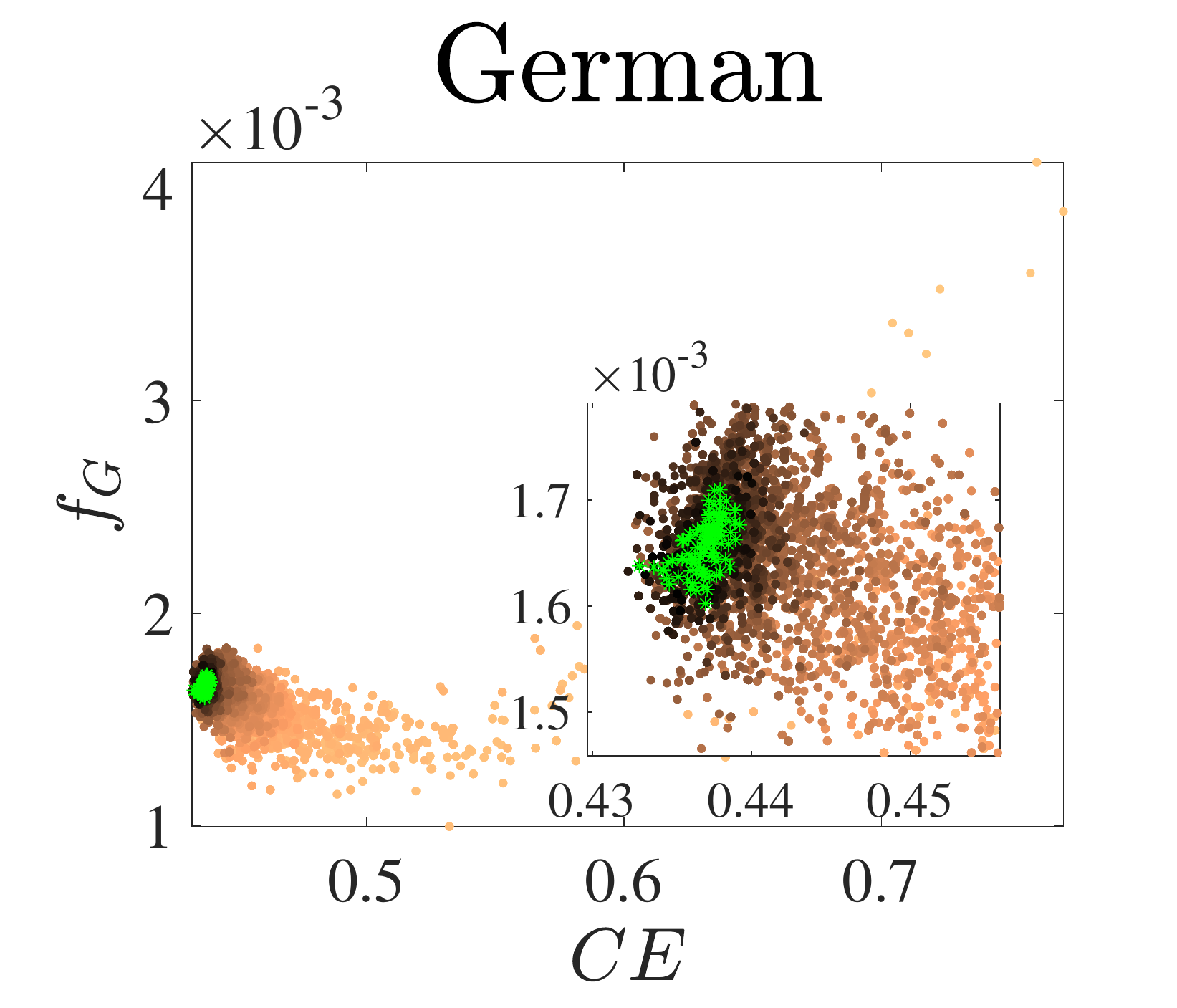}\hspace{1em}
    \includegraphics[width=.13\textwidth]{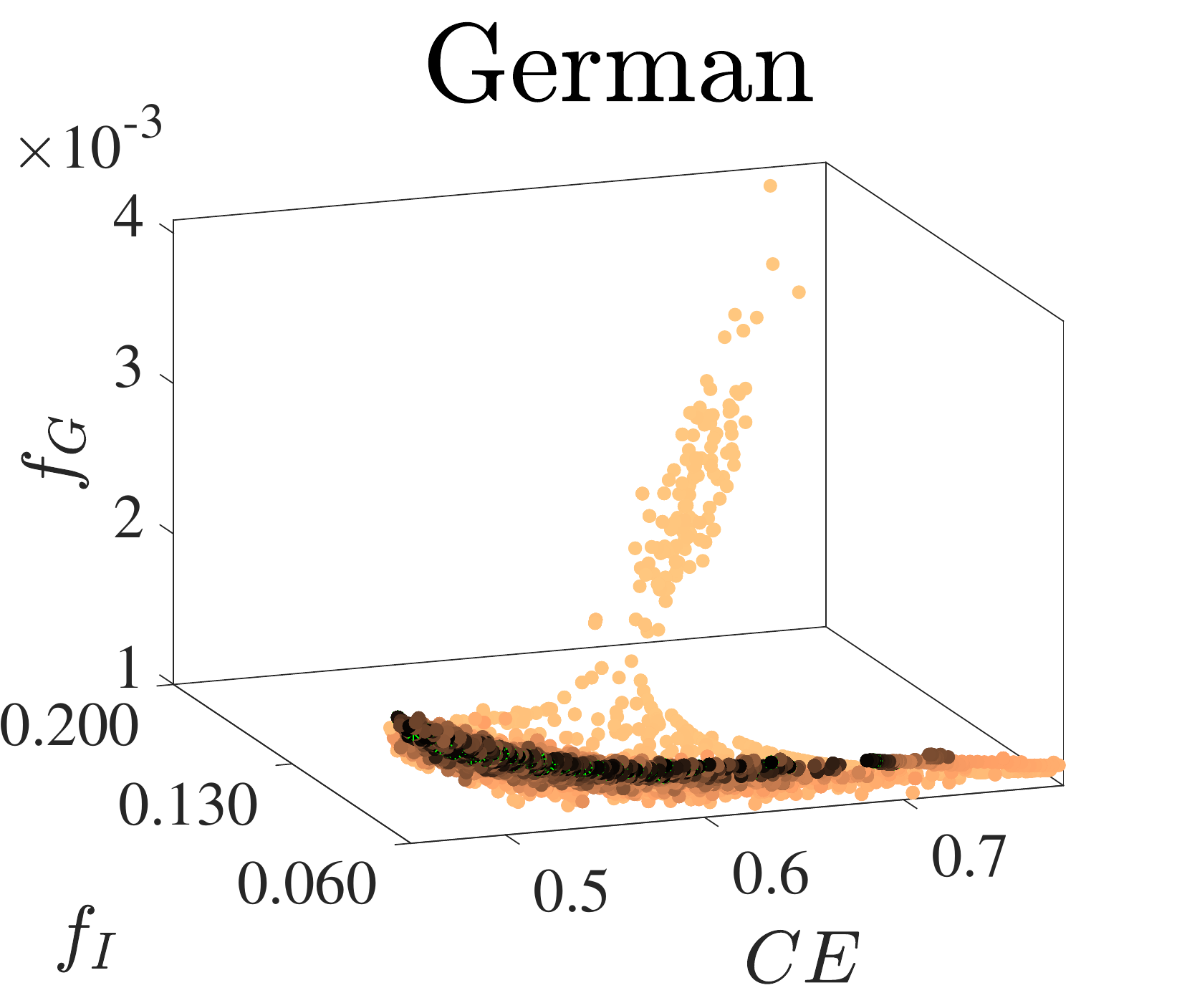}\hspace{1em}\\
    
    \includegraphics[width=.13\textwidth]{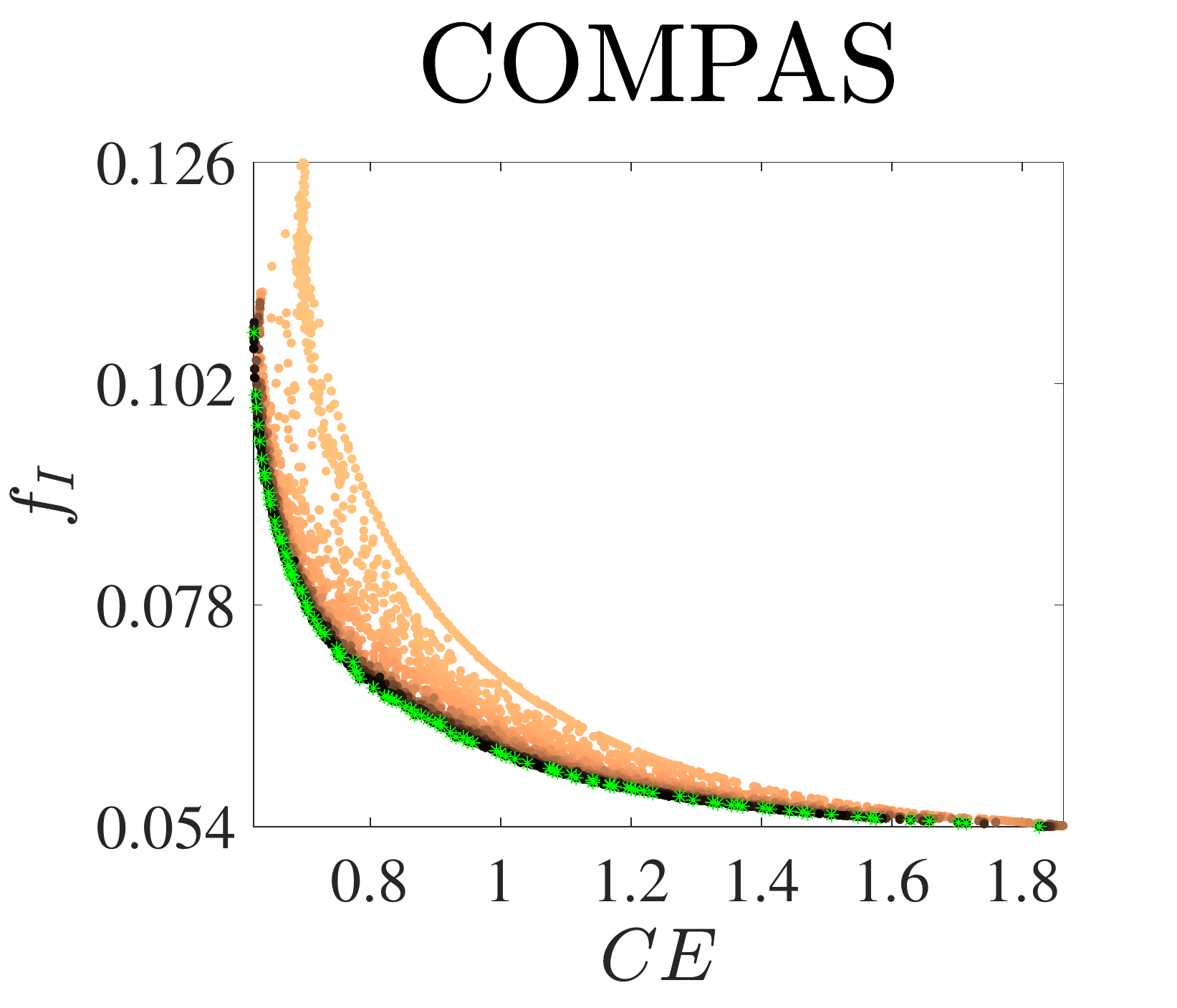}\hspace{1em}
    \includegraphics[width=.13\textwidth]{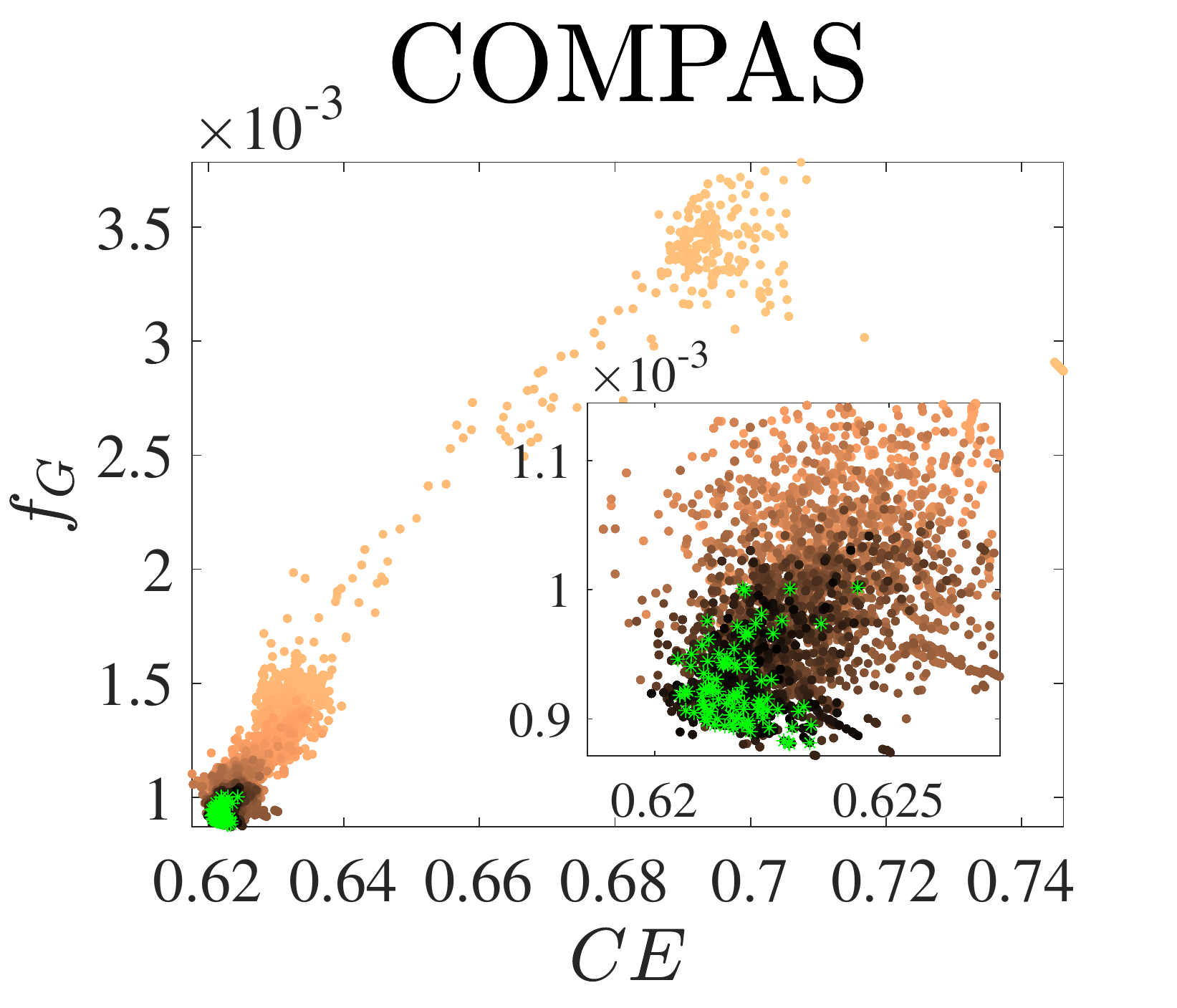}\hspace{1em}
    \includegraphics[width=.13\textwidth]{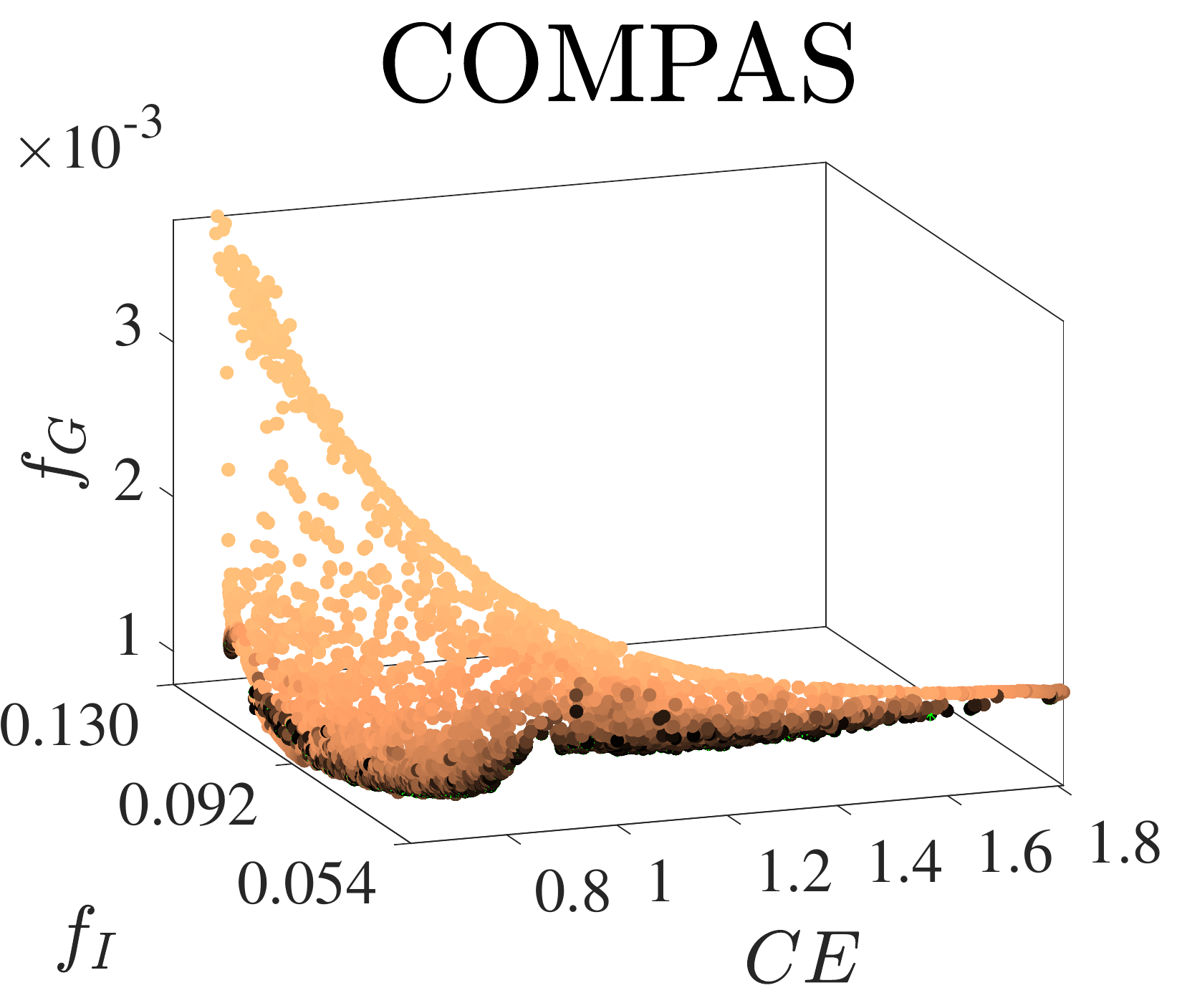}\hspace{1em}\\
    
    \includegraphics[width=.13\textwidth]{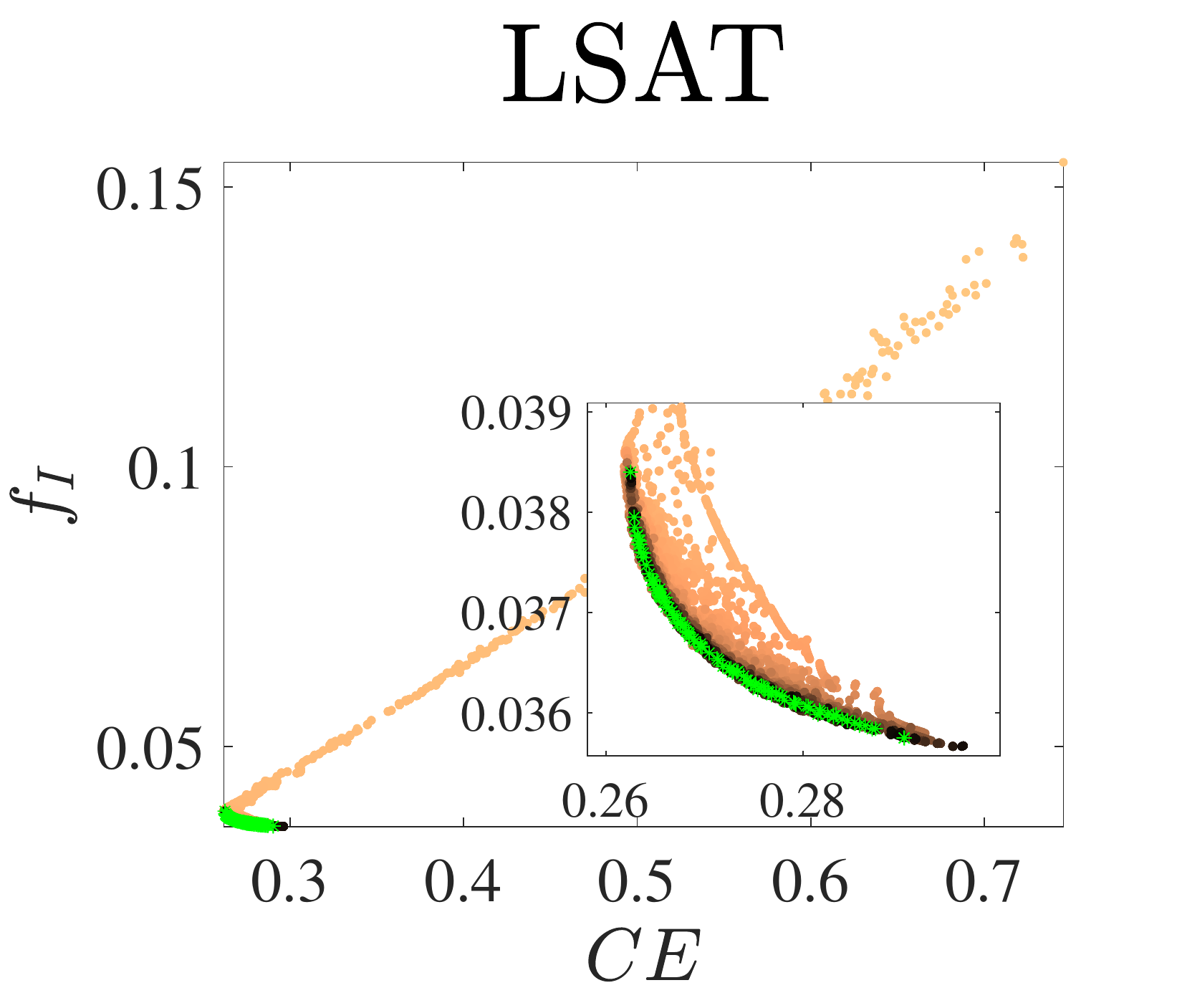}\hspace{1em}
    \includegraphics[width=.13\textwidth]{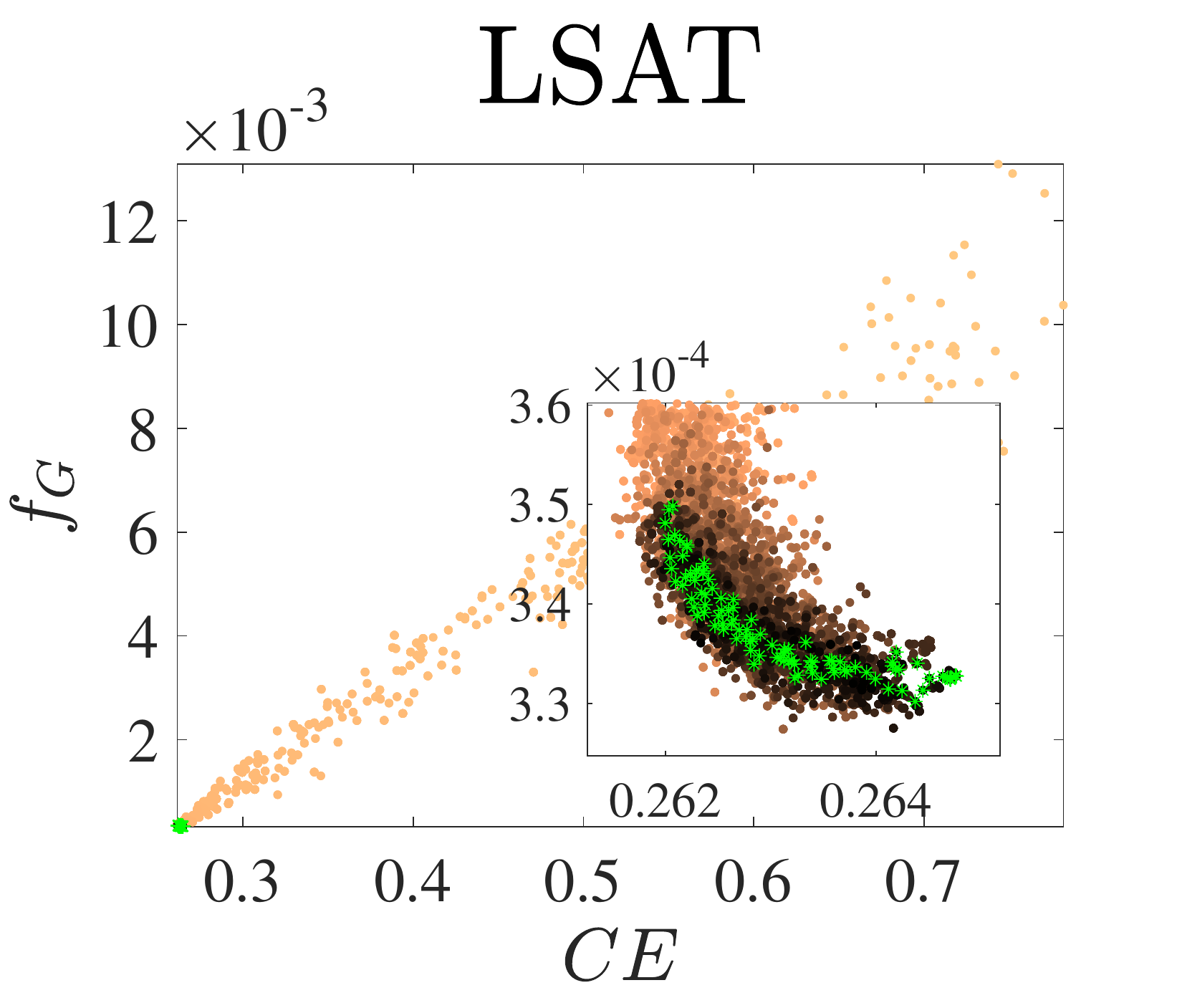}\hspace{1em}
    \includegraphics[width=.13\textwidth]{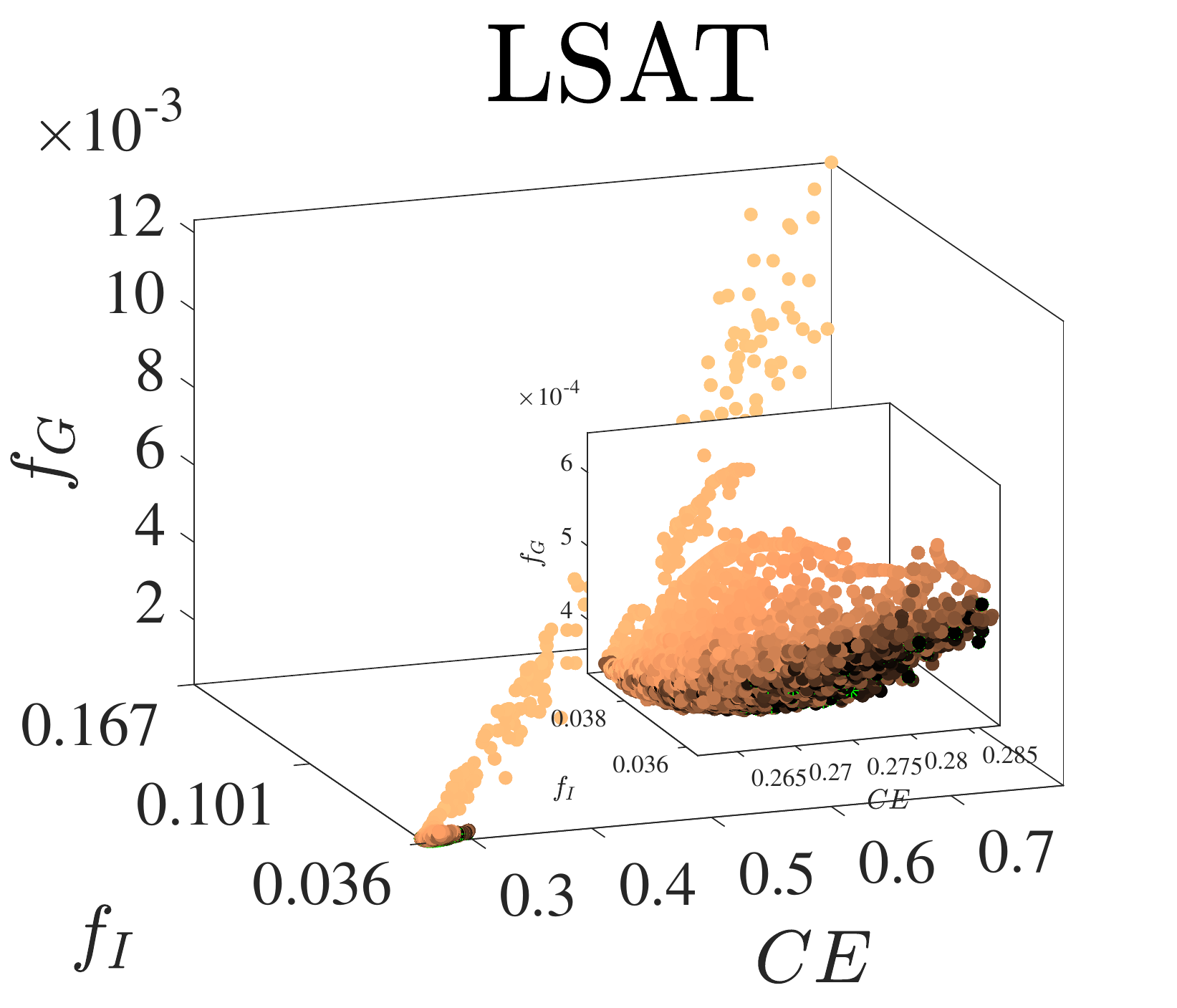}\hspace{1em}\\
    
    \includegraphics[width=.13\textwidth]{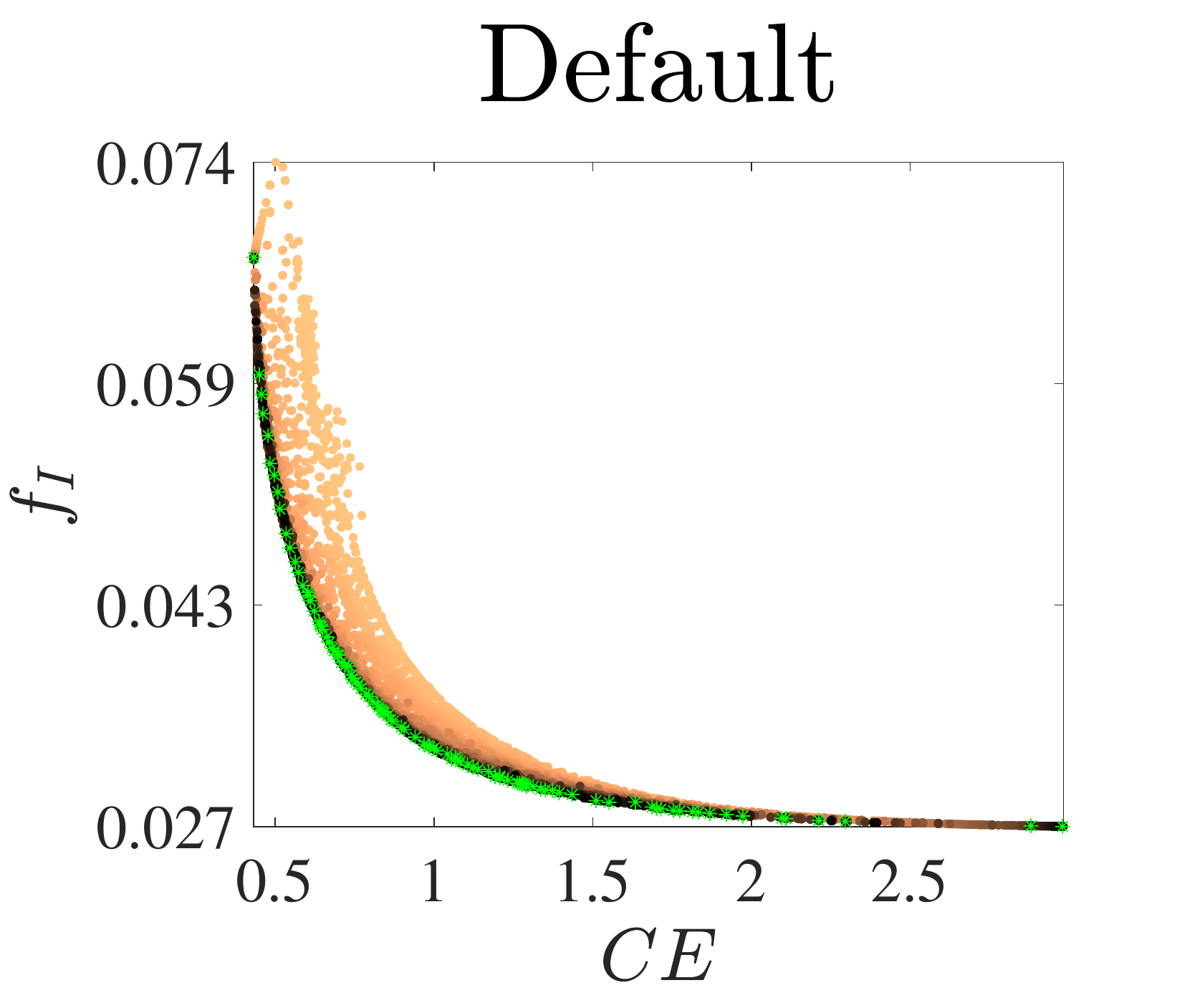}\hspace{1em}
    \includegraphics[width=.13\textwidth]{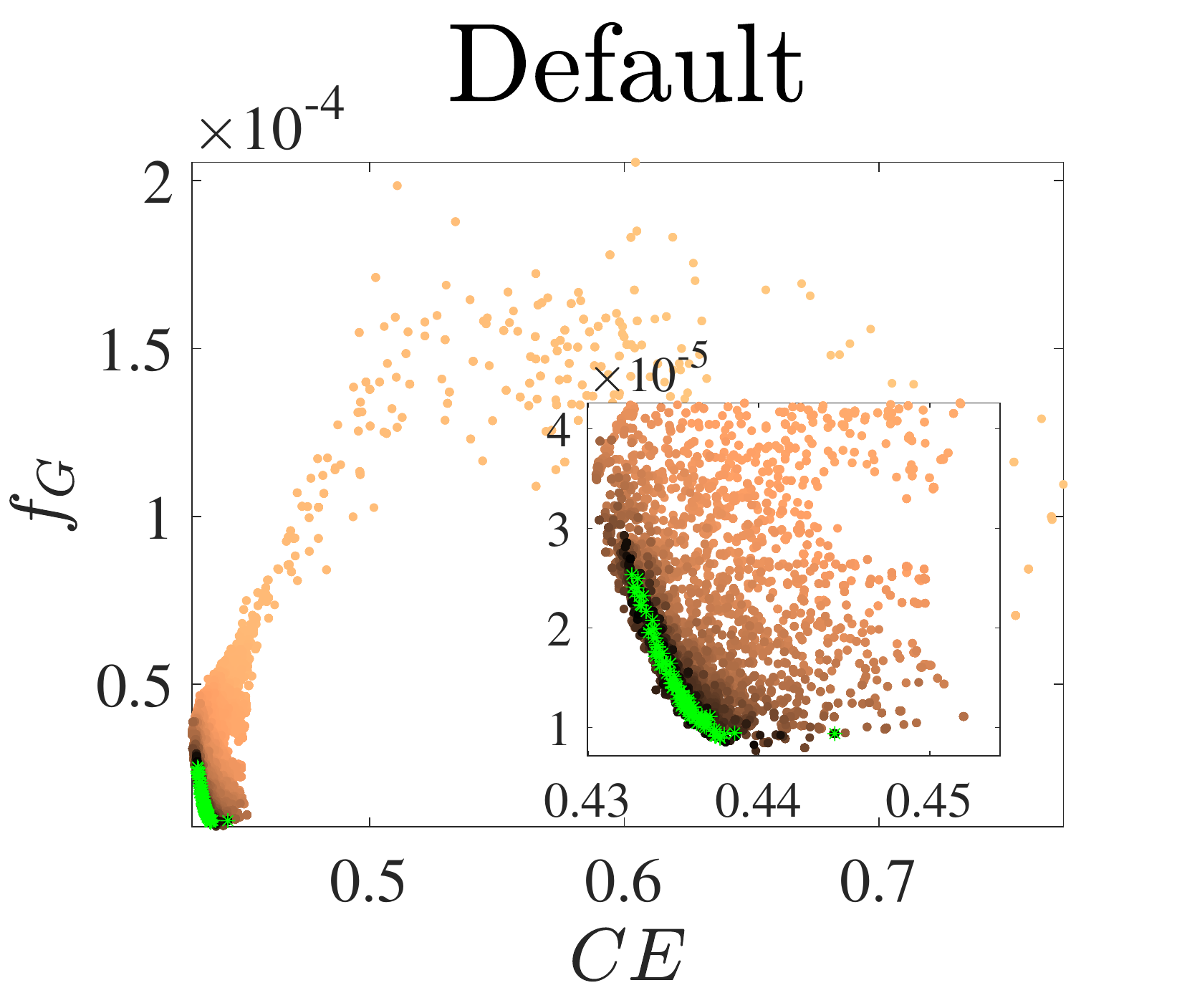}\hspace{1em}
    \includegraphics[width=.13\textwidth]{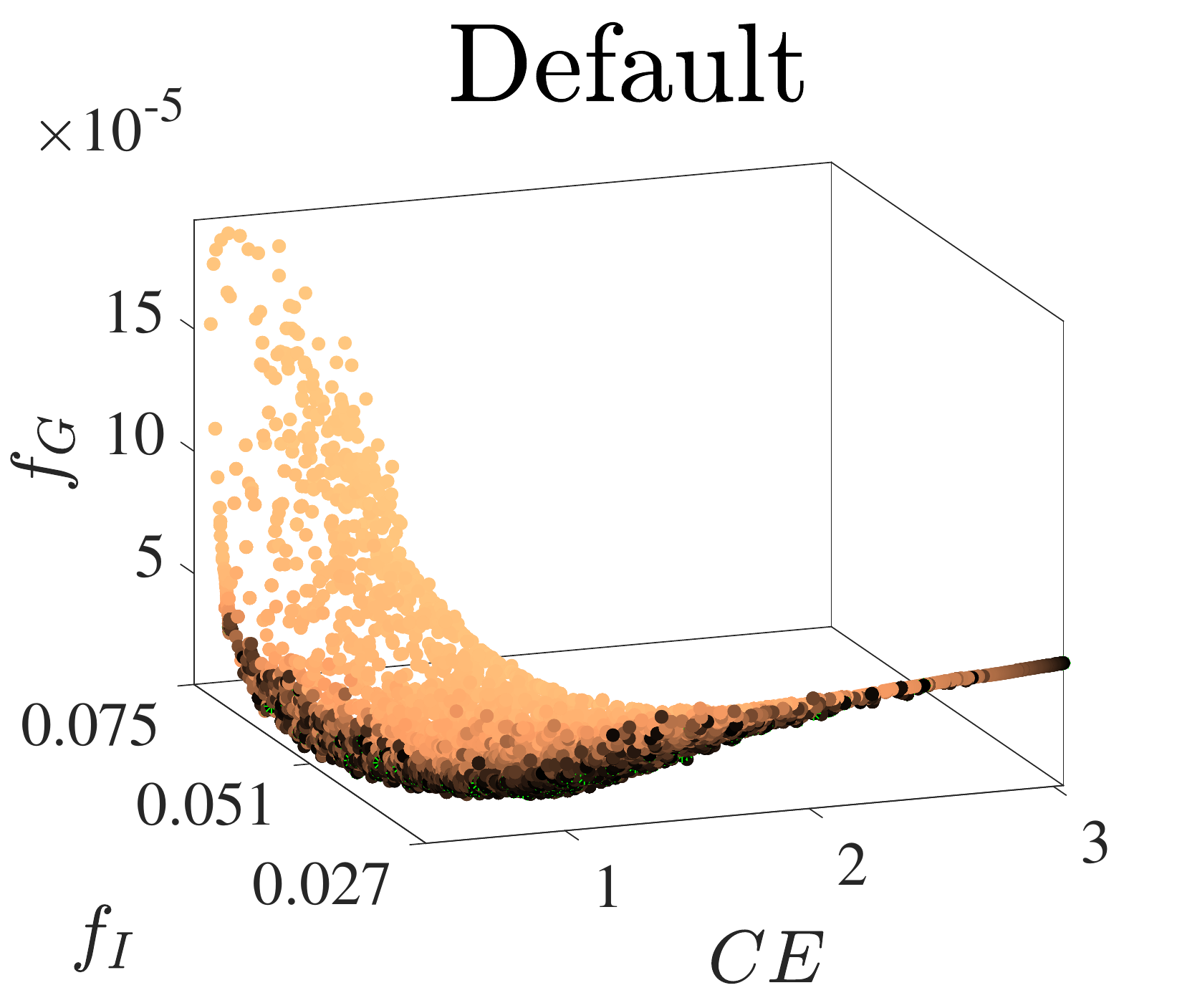}\hspace{1em}\\
    
    \includegraphics[width=.13\textwidth]{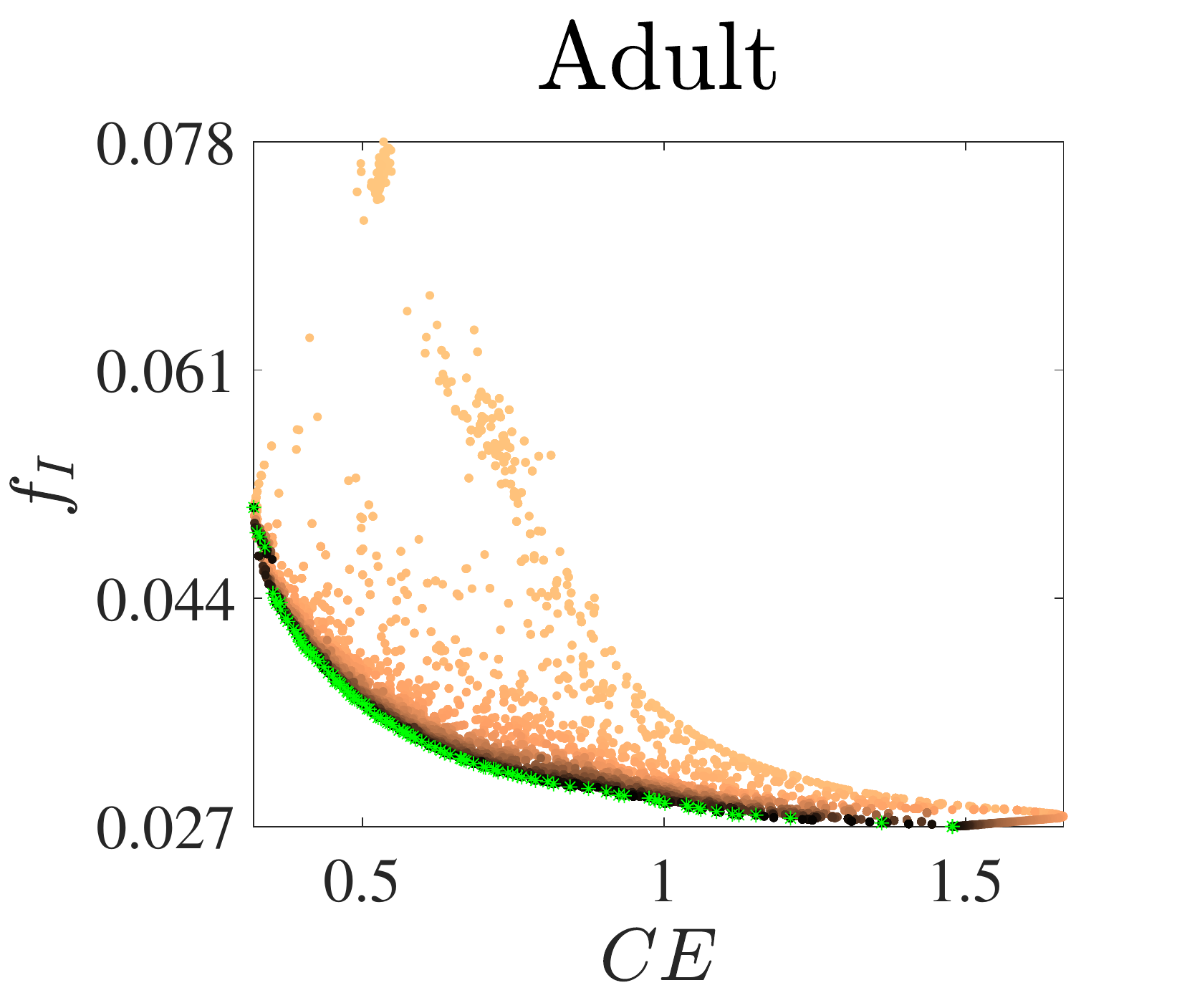}\hspace{1em}
    \includegraphics[width=.13\textwidth]{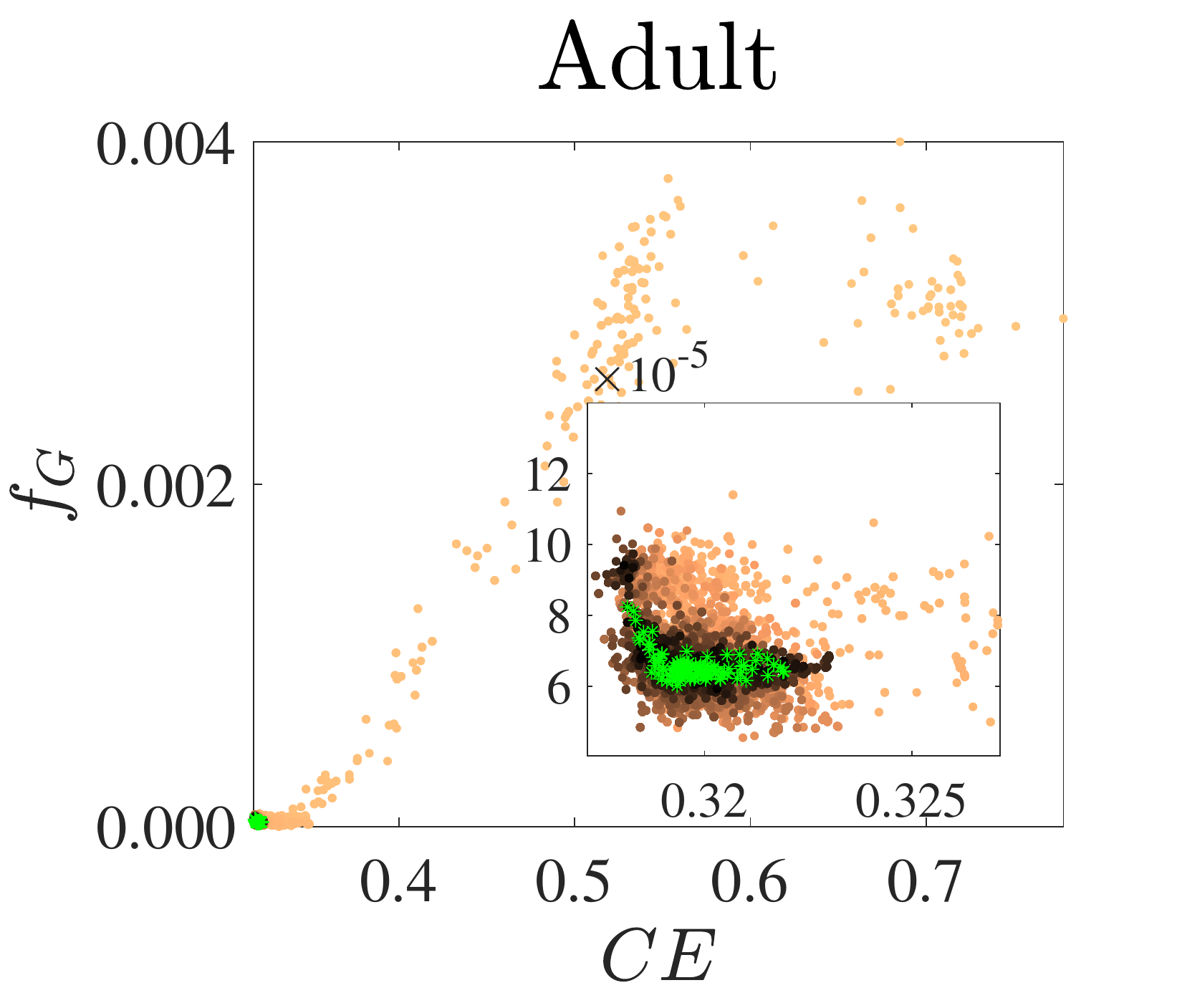}\hspace{1em}
    \includegraphics[width=.13\textwidth]{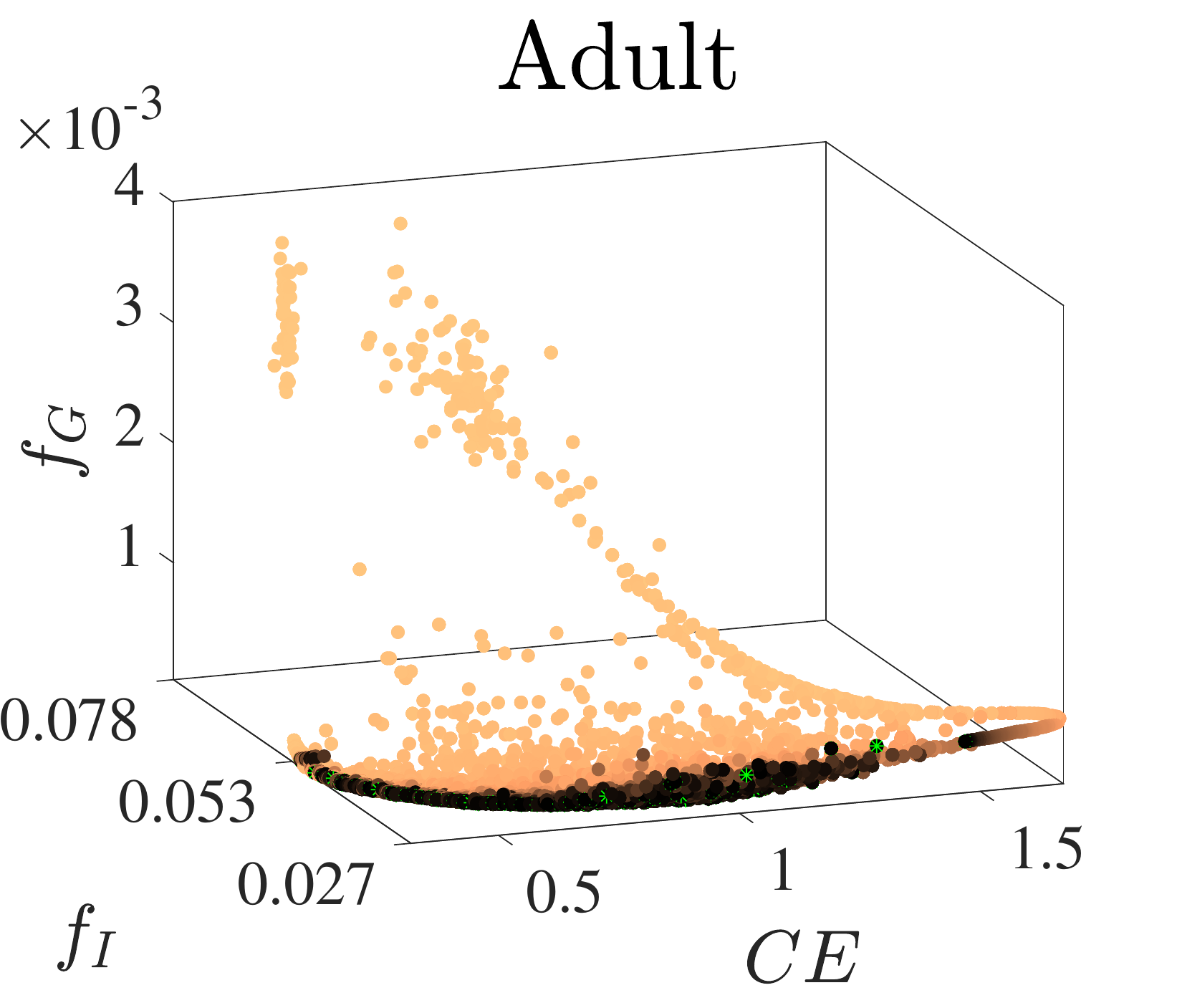}\hspace{1em}\\
    
    \includegraphics[width=.13\textwidth]{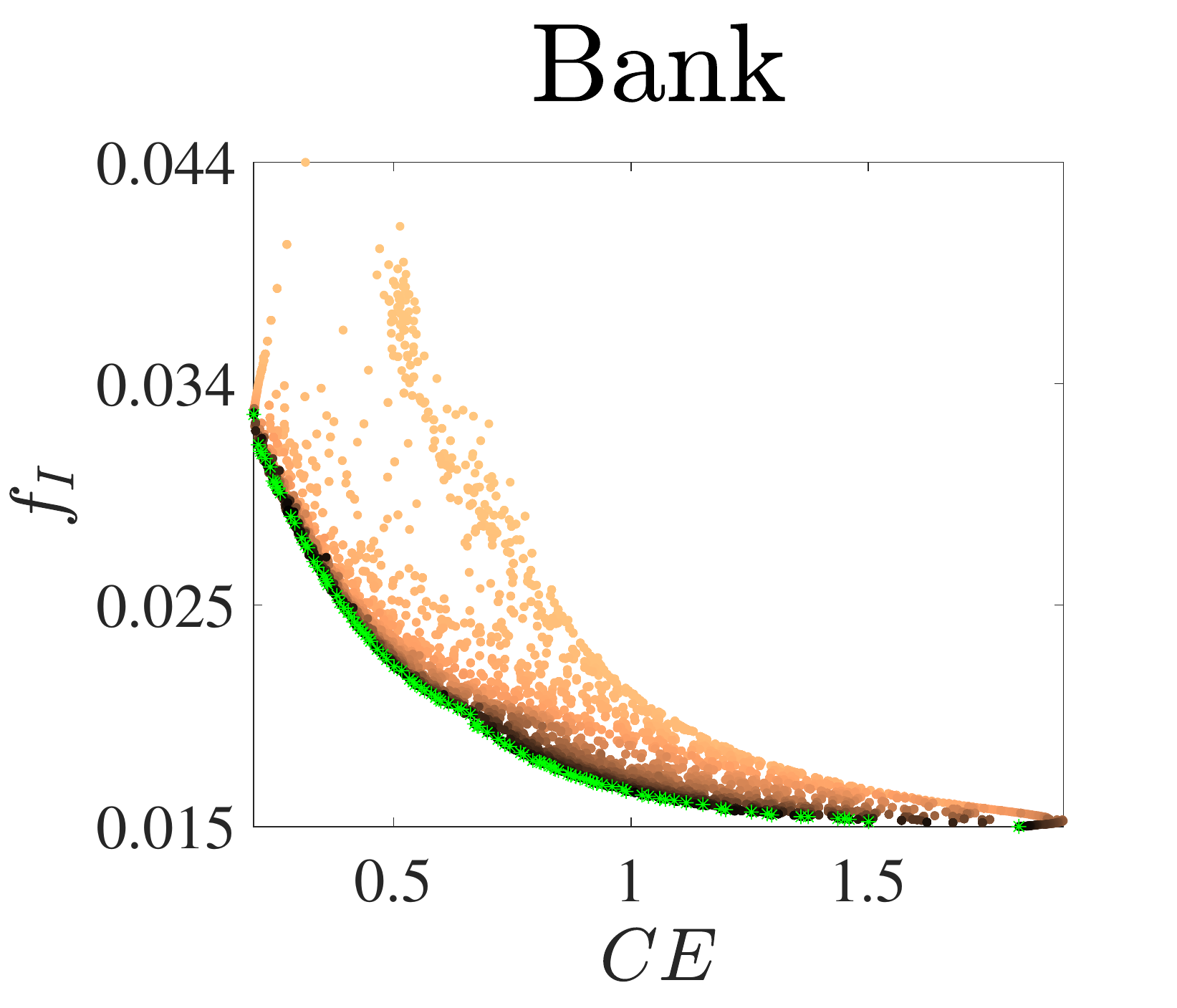}\hspace{1em}
    \includegraphics[width=.13\textwidth]{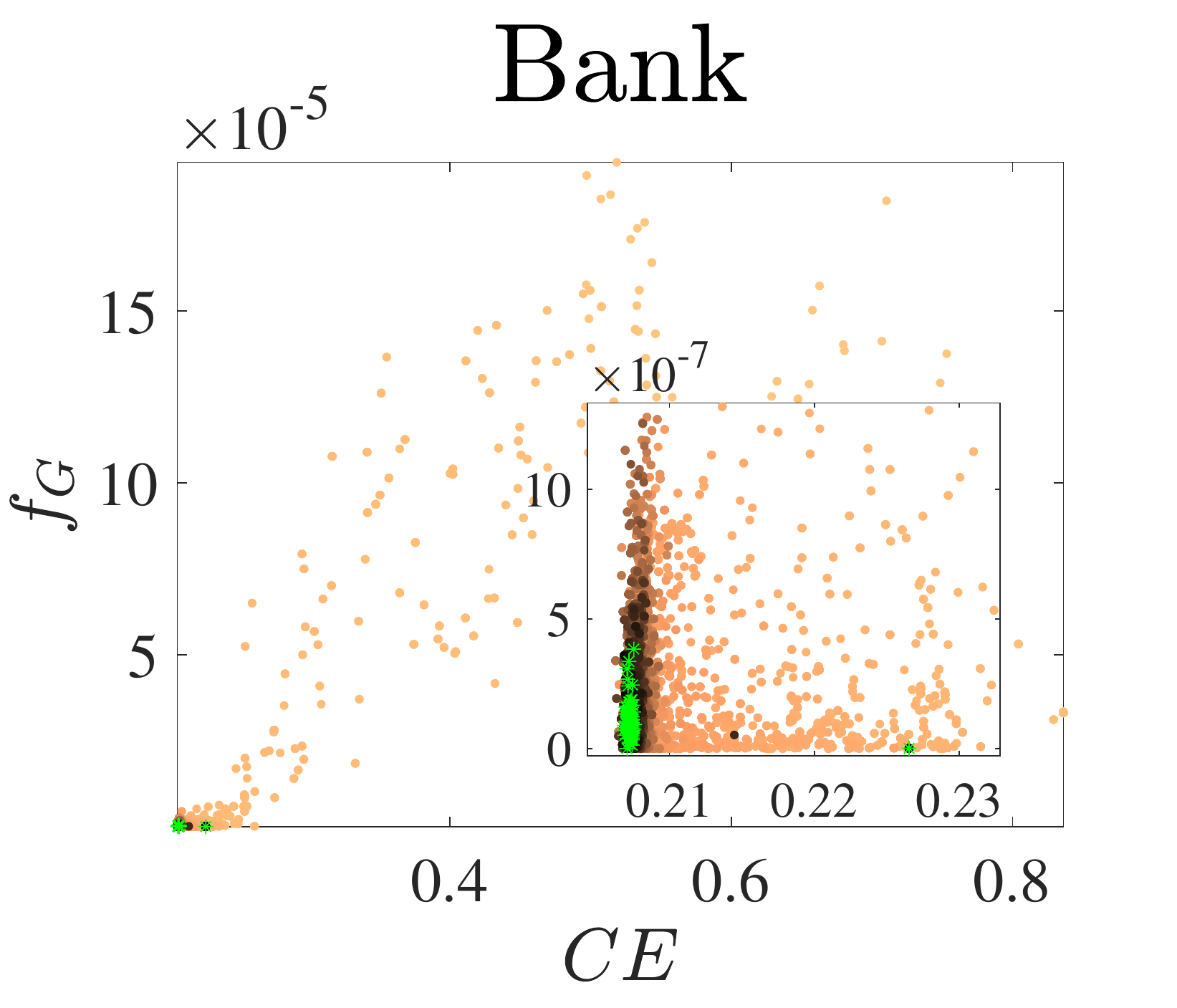}\hspace{1em}
    \includegraphics[width=.13\textwidth]{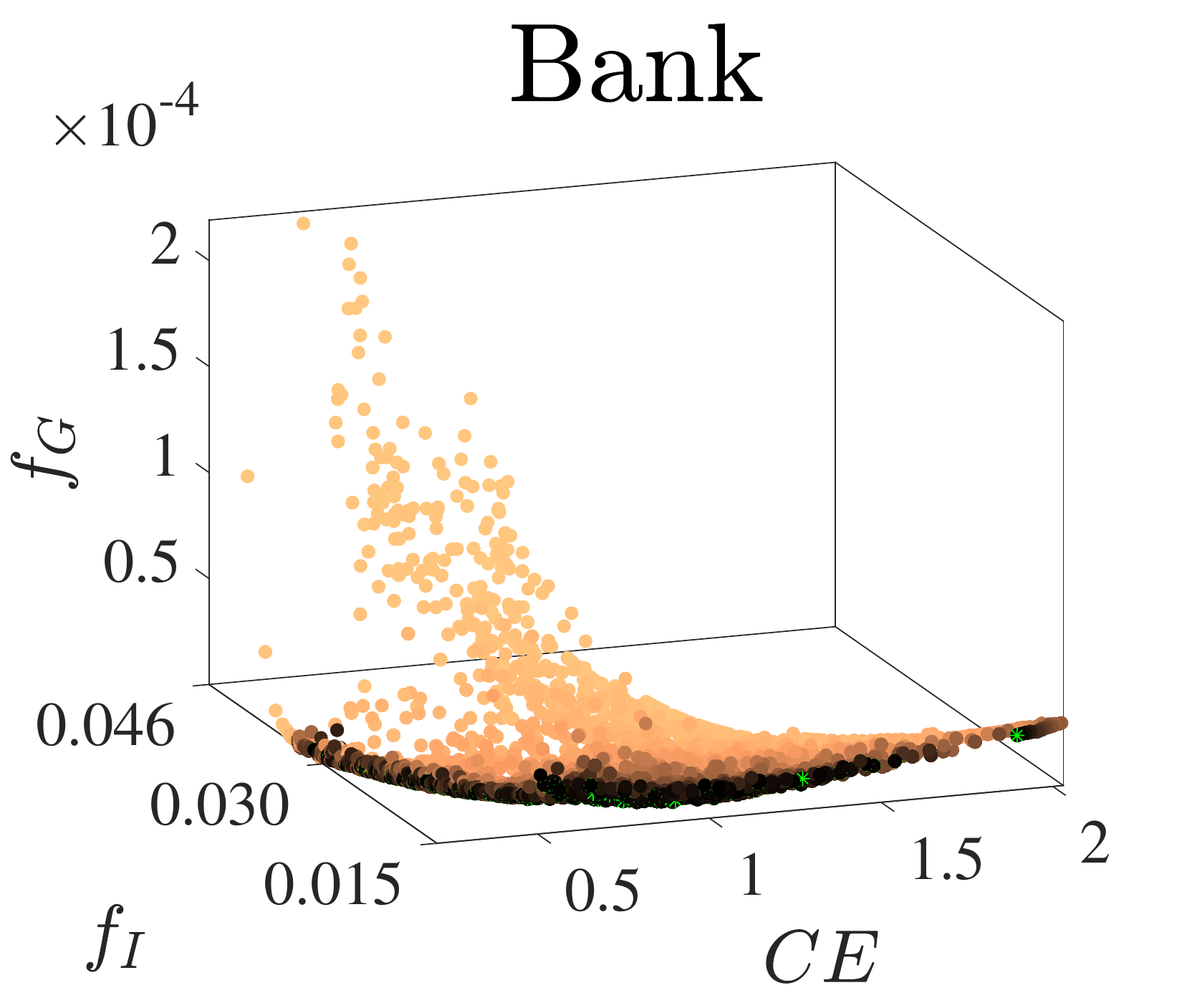}\hspace{1em}\\
    
    \includegraphics[width=.13\textwidth]{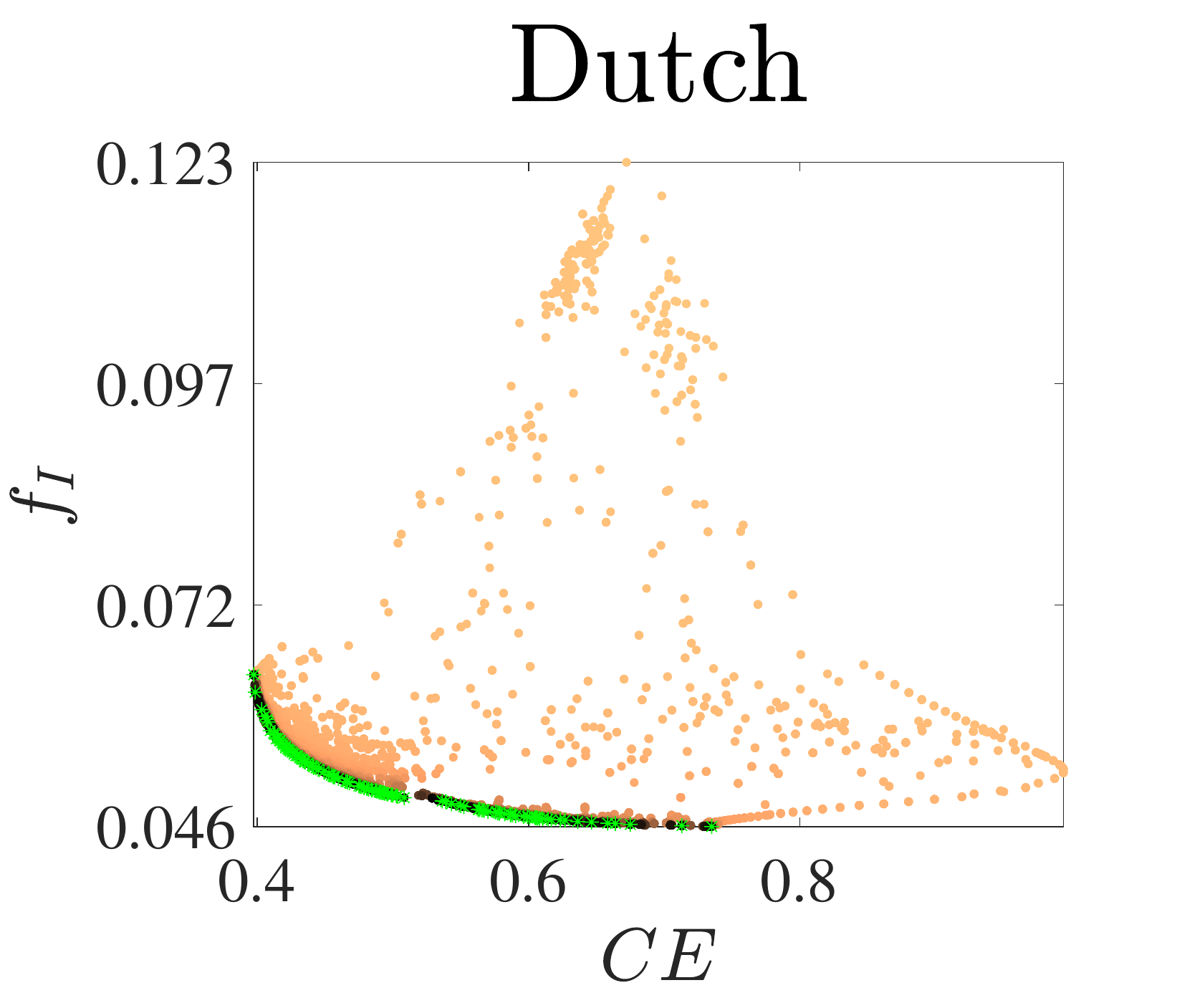}\hspace{1em}
    \includegraphics[width=.13\textwidth]{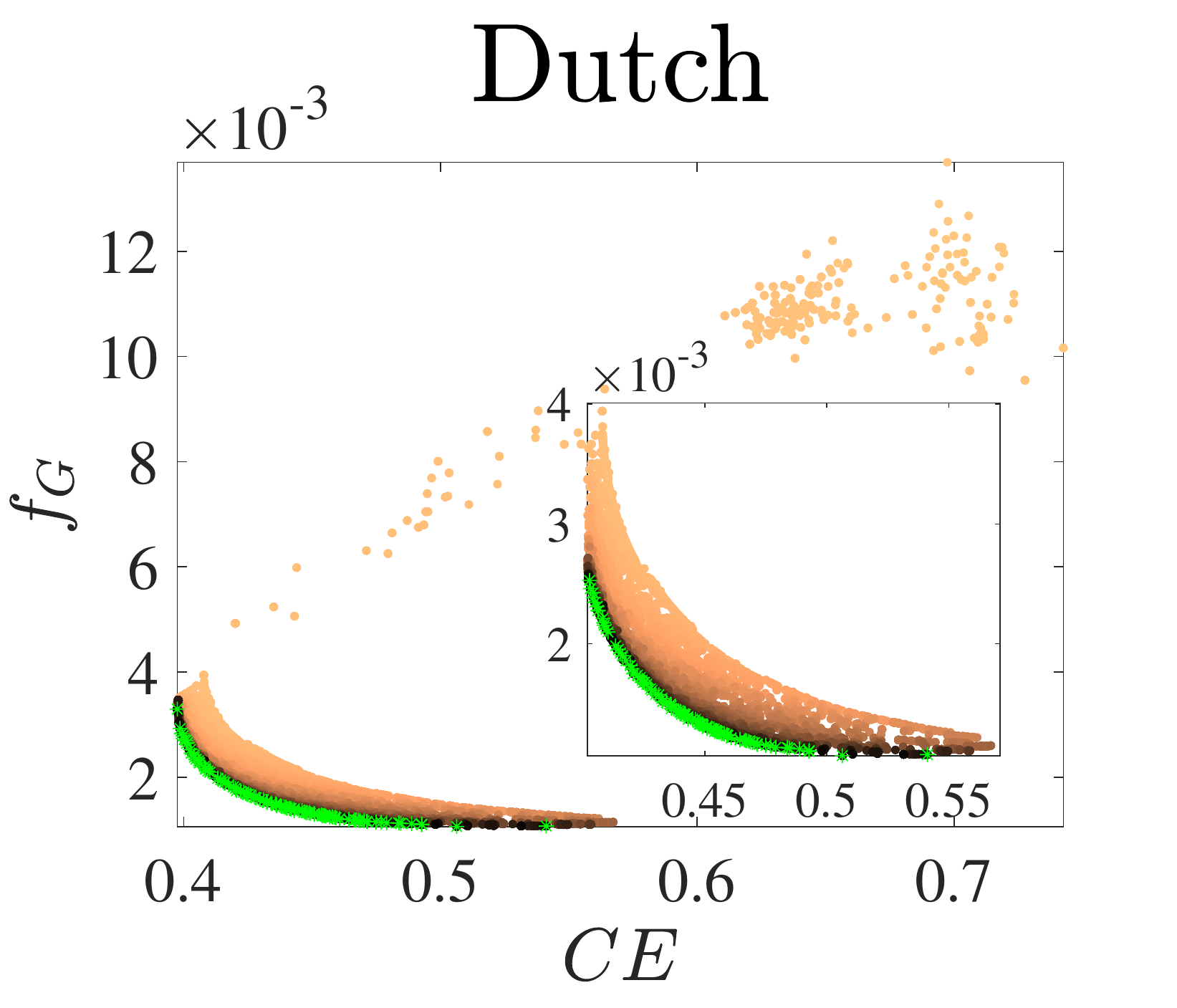}\hspace{1em}
    \includegraphics[width=.13\textwidth]{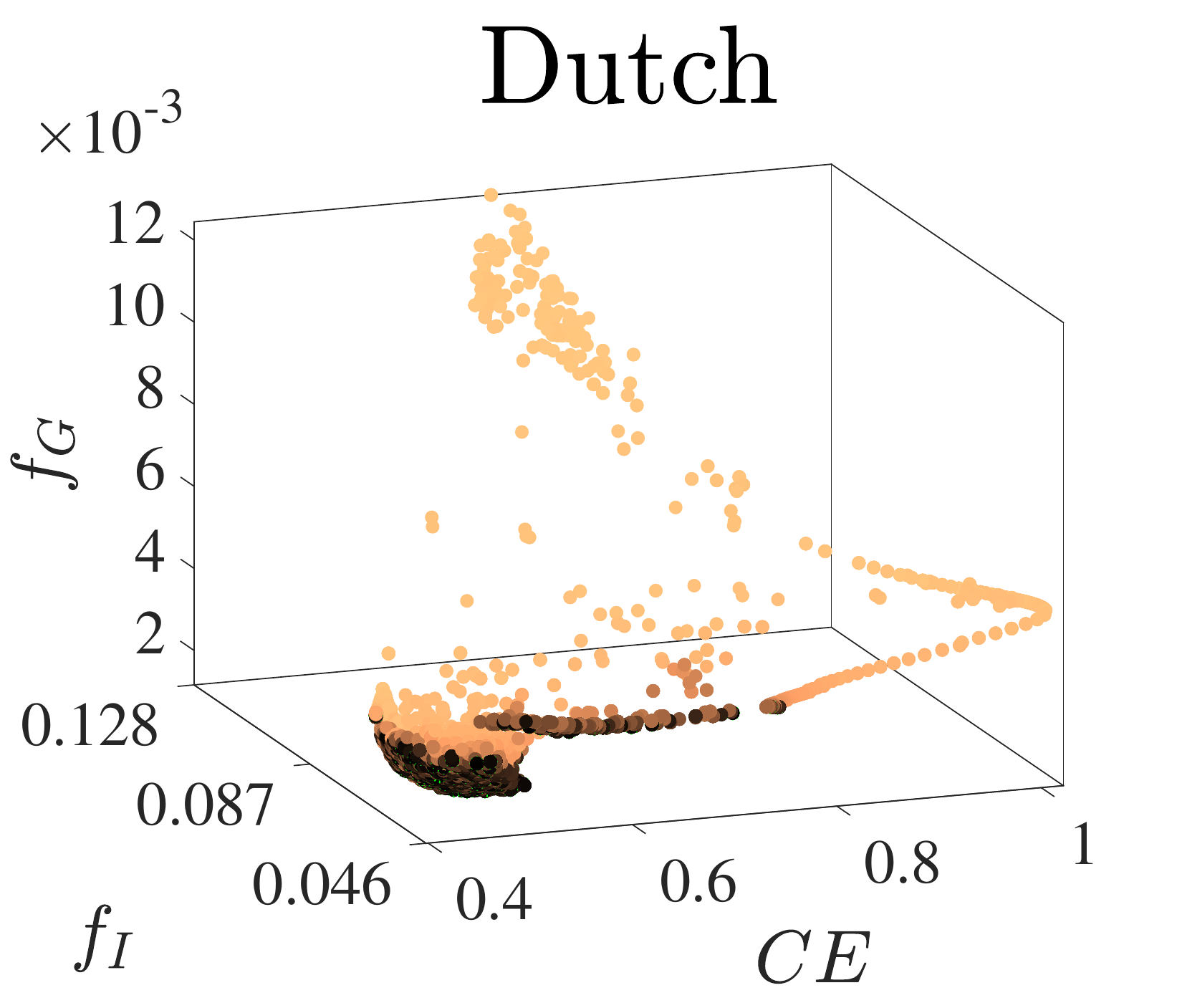}
    
    \caption{\label{fig:process}Evaluated $F_{EI}$, $F_{EG}$ and $F_{EIG}$ values (left to right) on the test set. Different colours indicate solutions at different generations. Green stars highlight the non-dominated solutions in the final generation.}
\end{figure}

Fig. \ref{fig:process} illustrates the optimisation process of arbitrarily selected trials of $F_{EI}$, $F_{EG}$ and $F_{EIG}$ on the test sets, where the non-dominated solutions of each generation are drawn with colour darken as the evolution progresses. It's clearly shown that model error and one or two unfairness measures converge simultaneously towards Pareto fronts (green stars).

\begin{figure}[htbp]
	\centering  
	\includegraphics[width=0.9\linewidth]{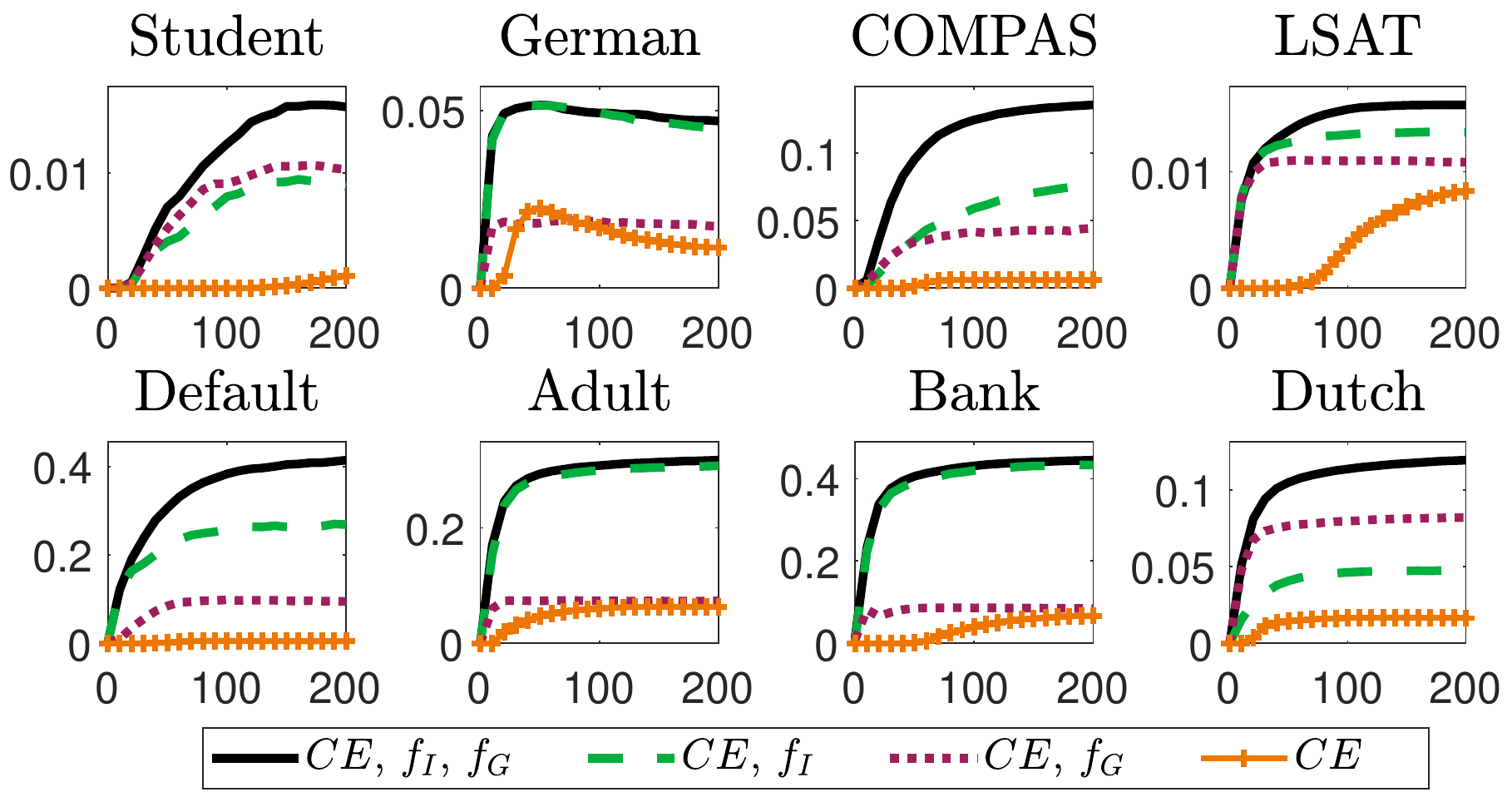}
    \caption{\label{fig:HV}HV values averaged over 30 trials considering their corresponding objectives on the \emph{set set}. \emph{x-axis}: generation number; \emph{y-axis}: HV value. }
\end{figure}

The convergence curves of HV values that quantify the optimisation process over all the independently repeated trials are shown in Fig. \ref{fig:HV}. Specifically, for each dataset, the pseudo Pareto front based on the three objectives is determined considering all the solutions from all the generations of 30 trials of $F_{E}$, $F_{EI}$, $F_{EG}$ and $F_{EIG}$. Then, based on the pseudo Pareto front, we record the average HV values of every 10 generations in each experiment over all the 30 trials. It's worth noting that $CE$, $f_{I}$ and $f_{G}$ \emph{are all involved in the calculation of HV}, which makes the HVs of different algorithms comparable with each other, although their optimised objectives are different. Therefore, the HV results can represent the performance of $CE$, $f_{I}$, and $f_{G}$ in terms of convergence and diversity.

As shown in Fig. \ref{fig:HV}, in all the three experiments considering two or three objectives, the HV values increase along with evolution, implying that the diversity and convergence become significantly better and that the model error, individual and group unfairness decrease along with evolution. During the evolution, since $F_{E}$ only optimises $CE$, the improvement of $CE$ may lead to worse $f_{I}$ and $f_{G}$, which makes little improvement in terms of HV on most datasets. The HV values of $F_{EIG}$ (black curve) are always larger than others, indicating that $f_{I}$ and $f_{G}$ both become better, as well as the accuracy.

We record the HV values of the model sets in the final generation of the four comparison algorithms on the test data in 30 trials in Table \ref{tab:HVs_cmp}, where ``$+/\approx/-$'' indicates the averaged HV values of corresponding algorithms are statistically better/similar/worse than that of $F_{EIG}$ according to the Wilcoxon rank sum test with a 0.05 significance level. Table \ref{tab:HVs_cmp} indicates that except that $F_{EIG}$ and $F_{EI}$ have the same performance on the \emph{German} dataset, $F_{EIG}$ is statistically superior to others in terms of HV values, which shows that $F_{EIG}$ can optimise accuracy, individual and group unfairness measures simultaneously without sacrificing any of them.

\begin{table}[htbp]
  \centering
  \caption{HV values of final solutions averaged over 30 trials. ``+/$\approx$/-'' indicates that the average HV value of the corresponding algorithm (specified by column header) is statistically better/similar/worse than the one of $F_{EIG}$ according to the Wilcoxon rank sum test with a 0.05 significance level. The best averaged HV values are highlighted in grey backgrounds.}
  \begin{adjustbox} {max width=\linewidth}
    \begin{tabular}{lllll}
    \toprule
     Dataset   & \multicolumn{1}{c}{$F_{E}$} & \multicolumn{1}{c}{$F_{EG}$}& \multicolumn{1}{c}{$F_{EI}$} & \multicolumn{1}{c}{$F_{EIG}$} \\
    \midrule
    \emph{Student} &  0.00106 (1.013e-03)- & 0.01032 (1.066e-02)- & 0.00884 (1.127e-02)- & \cellcolor[rgb]{ .751,  .751,  .751}0.01574 (8.184e-03) \\
    \emph{German} & 0.01134 (7.364e-03)- & 0.01736 (5.404e-03)- & 0.04498 (1.066e-02)$\approx$ & \cellcolor[rgb]{ .751,  .751,  .751}0.04701 (7.900e-03) \\
    \emph{COMPAS} &  0.00581 (1.655e-03)- & 0.04450 (6.052e-03)- & 0.07626 (1.038e-02)- & \cellcolor[rgb]{ .751,  .751,  .751}0.13603 (1.171e-02) \\
    \emph{LSAT}  & 0.00833 (2.708e-03)- & 0.01083 (3.697e-04)- & 0.01339 (5.811e-04)- & \cellcolor[rgb]{ .751,  .751,  .751}0.01574 (4.291e-04) \\
    \emph{Default} &  0.00472 (2.730e-03)- & 0.09464 (9.910e-03)- & 0.26846 (3.250e-02)- & \cellcolor[rgb]{ .751,  .751,  .751}0.41414 (2.385e-02) \\
    \emph{Adult} & 0.06335 (1.722e-03)- & 0.07339 (3.406e-03)- & 0.30749 (4.012e-03)- & \cellcolor[rgb]{ .751,  .751,  .751}0.31719 (5.396e-03) \\
    \emph{Bank}  &  0.06722 (1.519e-02)- & 0.08449 (3.688e-03)- & 0.43374 (1.404e-02)- & \cellcolor[rgb]{ .751,  .751,  .751}0.44513 (5.176e-03) \\
    \emph{Dutch} &  0.01644 (5.373e-04)- & 0.08196 (1.612e-03)- & 0.04734 (1.594e-03)- & \cellcolor[rgb]{ .751,  .751,  .751}0.11945 (1.250e-03) \\
    \bottomrule
    \end{tabular}%
     \end{adjustbox}
  \label{tab:HVs_cmp}%
\end{table}%

Finally, we compare the state-of-the-art Multi-FR with $F_{EIG}$. $Dominate$, $Incomparable$ and $Dominated$ values of models obtained by Multi-FR with $l_2$ or $loss$ or $loss+$ or no normalisation on the test data in 30 trials are recorded in Table \ref{tab:threemetrics}. The $Dominate$ values of Multi-FR without normalisation is always 1 on all datasets, which implies that there are always some models generated by $F_{EIG}$ having better performance in terms of all the three objectives, $CE$, $f_{I}$, and $f_{G}$, than the models obtained by Multi-FR without normalisation in each trial. On \emph{Student}, \emph{German}, \emph{COMPAS}, \emph{Default}, and \emph{Bank} datasets, the $Dominate$ values of Multi-FR with either $loss$ or $loss+$ are larger than 0.65, which indicates that $F_{EIG}$ has a high probability of generating models that are better than Multi-FR in terms of $CE$, $f_{I}$, and $f_{G}$. On \emph{LSAT}, \emph{Adult} and \emph{Dutch} datasets, although the $Dominate$ values of Multi-FR with $l_2$ or $loss$ or $loss+$ is not high, $Dominated$ values of them is high and $Incomparable$ values of them is low, which means there are many models generated by $F_{EIG}$ that are incomparable with the models of Multi-FR. Therefore, compared with four variants of Multi-FR, $F_{EIG}$ can provide the models that have better $CE$, $f_{I}$, and $f_{G}$ on 5 out of 8 datasets and perform no worse on the rest of datasets.

In summary, the experimental results show that $F_{EIG}$ applying multi-objective learning can simultaneously optimise accuracy and multiple fairness measures and outperform the state-of-the-art.

\begin{table}[htbp]
  \centering
  \caption{$Dominate$, $Incomparable$ and $Dominated$ values of models obtained by Multi-FR with $l_2$, $loss$, $loss+$ or no normalisation corresponding to $F_{EIG}$ on test data averaged over 30 trials.}
  \begin{adjustbox} {max width=\linewidth}
    \begin{tabular}{c|l|ccc}
    \toprule
    Dataset & Multi-FR  & $Dominate$ & $Incomparable$  &  $Dominated$\\
    \midrule
    \multirow{4}[2]{*}{\emph{Student}} &  $l_2$ & 0.72111 (3.281e-03) & 0.02977 (2.811e-02) & 0.27979 (3.020e-01) \\
          &  $loss$ & 0.72667 (2.796e-03) & 0.03123 (2.395e-02) & 0.28912 (2.658e-01) \\
          &  $loss+$ & 0.60667 (3.101e-03) & 0.03777 (2.256e-02) & 0.39137 (2.866e-01) \\
          & no norm & 1.00000 (3.469e-18) & 0.00007 (2.494e-04) & 0.00004 (1.663e-04) \\
    \hline
    \multirow{4}[2]{*}{\emph{German}} &  $l_2$ & 0.84111 (2.341e-03) & 0.10833 (1.420e-01) & 0.07904 (1.097e-01) \\
          &  $loss$ & 0.75444 (2.552e-03) & 0.16963 (1.503e-01) & 0.12300 (1.184e-01) \\
          &  $loss+$ & 0.65111 (3.066e-03) & 0.22798 (1.678e-01) & 0.16929 (1.379e-01) \\
          & no norm & 1.00000 (3.469e-18) & 0.00000 (0.000e+00) & 0.00000 (0.000e+00) \\
    \hline
    \multirow{4}[2]{*}{\emph{COMPAS}} &  $l_2$ & 0.78889 (1.726e-03) & 0.22681 (1.601e-01) & 0.00883 (7.003e-03) \\
          &  $loss$ & 0.92333 (9.114e-04) & 0.09312 (8.698e-02) & 0.00402 (3.688e-03) \\
          &  $loss+$ & 0.91889 (1.053e-03) & 0.09984 (1.039e-01) & 0.00436 (4.275e-03) \\
          & no norm & 1.00000 (3.469e-18) & 0.00596 (4.099e-03) & 0.00023 (3.350e-04) \\
    \hline
    \multirow{4}[2]{*}{\emph{LSAT}} &  $l_2$ & 1.00000 (3.469e-18) & 0.00703 (6.520e-03) & 0.00000 (0.000e+00) \\
          &  $loss$ & 0.11889 (1.284e-03) & 0.80802 (1.061e-01) & 0.08166 (1.377e-02) \\
          &  $loss+$ & 0.11333 (1.176e-03) & 0.81427 (9.843e-02) & 0.08116 (1.141e-02) \\
          & no norm & 1.00000 (3.469e-18) & 0.00006 (2.447e-04) & 0.00000 (0.000e+00) \\
    \hline
    \multirow{4}[2]{*}{\emph{Default}} &  $l_2$ & 0.44556 (7.174e-04) & 0.54748 (6.486e-02) & 0.02406 (3.444e-03) \\
          &  $loss$ & 1.00000 (3.469e-18) & 0.02567 (1.735e-17) & 0.00167 (8.674e-19) \\
          &  $loss+$ & 0.86333 (2.012e-03) & 0.15516 (1.878e-01) & 0.00750 (8.327e-03) \\
          & no norm & 1.00000 (3.469e-18) & 0.02220 (9.173e-04) & 0.00067 (2.168e-19) \\
    \hline
    \multirow{4}[2]{*}{\emph{Adult}} &  $l_2$ & 0.37444 (8.679e-04) & 0.61327 (7.977e-02) & 0.02581 (3.537e-03) \\
          &  $loss$ & 0.10111 (1.519e-03) & 0.85743 (1.371e-01) & 0.04531 (9.639e-03) \\
          &  $loss+$ & 0.01889 (4.689e-04) & 0.93110 (4.126e-02) & 0.05074 (6.347e-03) \\
          & no norm & 1.00000 (3.469e-18) & 0.00080 (4.000e-04) & 0.00000 (0.000e+00) \\
    \hline
    \multirow{4}[2]{*}{\emph{Bank}} &  $l_2$ & 0.65889 (1.831e-03) & 0.34613 (1.749e-01) & 0.01239 (5.048e-03) \\
          &  $loss$ & 0.83333 (3.232e-03) & 0.16770 (3.124e-01) & 0.00587 (1.006e-02) \\
          &  $loss+$ & 0.69889 (4.323e-03) & 0.29771 (4.156e-01) & 0.01026 (1.356e-02) \\
          & no norm & 1.00000 (3.469e-18) & 0.00643 (3.676e-03) & 0.00081 (4.188e-04) \\
    \hline
    \multirow{4}[2]{*}{\emph{Dutch}} &  $l_2$ & 0.07111 (5.356e-04) & 0.91911 (5.244e-02) & 0.00990 (6.095e-04) \\
          &  $loss$ & 0.07667 (1.540e-03) & 0.91382 (1.509e-01) & 0.01020 (1.964e-03) \\
          &  $loss+$ & 0.24222 (1.978e-03) & 0.75108 (1.929e-01) & 0.00900 (3.373e-03) \\
          & no norm & 1.00000 (3.469e-18) & 0.00508 (5.283e-04) & 0.00000 (0.000e+00) \\
    \bottomrule
    \end{tabular}%
  \label{tab:threemetrics}%
  \end{adjustbox}
\end{table}%

\noindent\textbf{($\mathcal{Q}2$) \emph{Can we obtain a group of diverse models by applying multi-objective learning?}}

To answer $\mathcal{Q}2$, three perspectives are considered on the \emph{test set}: (i) visualisation of Pareto fronts generated by $F_{EIG}$, (ii) evaluation based on diversity indicator CPF, (iii) comparison with the state-of-the-art algorithm Multi-FR~\cite{wu2021multifr}.

\begin{figure}[htbp]
    \centering
    \includegraphics[width=0.5\textwidth]{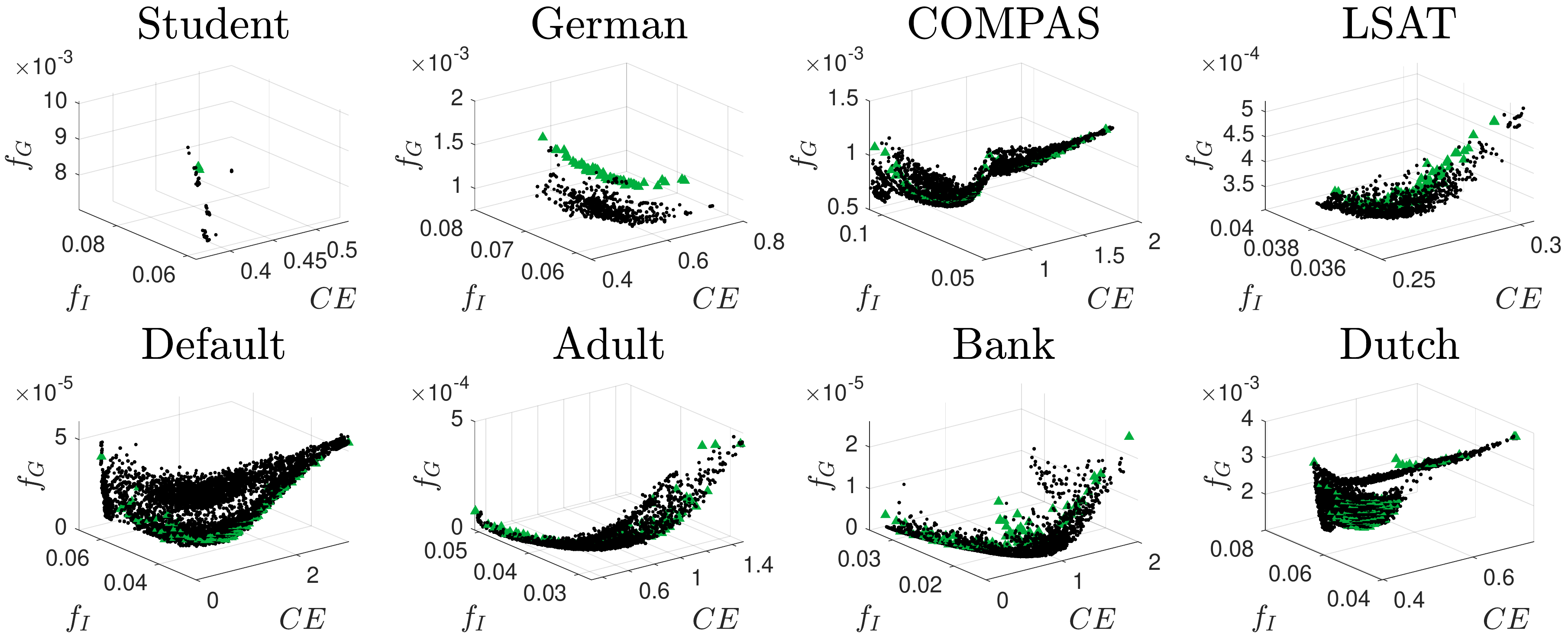}
    \caption{\label{fig:PF_objs3_final} Illustrative examples:
     (i) non-dominated solutions of the models obtained in the whole evolution process of $F_{EIG}$ in 30 trials (black points). (ii) non-dominated solutions of the models in the last generation of $F_{EIG}$ in one arbitrary trial (green triangles).}
\end{figure}

As shown in Fig. \ref{fig:PF_objs3_final}, we select non-dominated solutions of the models during the whole evolution process of $F_{EIG}$ in all trials as the Pareto fronts (black points). We plot non-dominated solutions of the models in the last generation of $F_{EIG}$ in one arbitrary trial (green triangles). Fig. \ref{fig:PF_objs3_final} clearly shows the tradeoffs among $CE$, $f_{I}$, and $f_{G}$ through the obtained Pareto fronts, which helps decision-makers to understand the different behaviours among the three objectives and how much  sacrifice of one certain metric can improve others to what extent. The diverse set of models can be observed from Fig. \ref{fig:PF_objs3_final} clearly except for the \emph{Student} dataset, whose diverse models might be better observed from Fig. \ref{fig:HV}.

We quantify the diversity of models obtained by $F_{E}$, $F_{EI}$, $F_{EG}$ and $F_{EIG}$, respectively. Since HV indicator can evaluate both convergence and diversity of a mode set, the results of Table \ref{tab:HVs_cmp} indicate that $F_{EIG}$ can provide a more diverse model set to some extent. In addition to HV, we choose CPF~\cite{Tian2019Diversity} that only evaluates the diversity~\cite{li2019quality} for further analysis. The pseudo Pareto fronts used in calculating CPF are the same as in calculating HV in $\mathcal{Q}1$ on the test data. Similarly, $CE$, $f_{I}$ and $f_{G}$ \emph{are all involved in the calculation of CPF}. We record the average CPF values in the last generation of $F_{E}$, $F_{EI}$, $F_{EG}$ and $F_{EIG}$, respectively, in all the 30 trials in Table \ref{tab:CPF}. Table \ref{tab:CPF} shows that the tri-objective $F_{EIG}$ can provide more diverse models than others on 5 out of 8 datasets according to the Wilcoxon rank sum test with  a 0.05 significance level. For \emph{Student}, \emph{German} and \emph{Default} datasets, the diversity performance of $F_{EIG}$ is comparable to others.

\begin{table}[htbp]
  \centering
  \caption{\label{tab:CPF}CPF values of final solutions averaged over 30 trials. 
  ``+/$\approx$/-'' indicates that the average CPF values of the corresponding algorithm (specified by column header) is statistically better/similar/worse than the one of $F_{EIG}$ according to the Wilcoxon rank sum test with a 0.05 significance level. Larger CPF values imply better performance. The best averaged CPF values are highlighted in grey backgrounds.}
  \begin{adjustbox} {max width=\linewidth}
    \begin{tabular}{lllll}
    \toprule
    Dataset   & \multicolumn{1}{c}{$F_{E}$} & \multicolumn{1}{c}{$F_{EG}$}& \multicolumn{1}{c}{$F_{EI}$} & \multicolumn{1}{c}{$F_{EIG}$} \\ 
    \midrule
    \emph{Student} &  0.00025 (1.362e-03)$\approx$ & 0.00050 (1.997e-03)$\approx$ & \cellcolor[rgb]{ .751,  .751,  .751}0.00233 (7.608e-03)$\approx$ & 0.00153 (3.449e-03) \\
    \emph{German} & 0.00033 (1.795e-03)- & 0.03347 (3.525e-02)$\approx$ & 0.00143 (4.269e-03)- & \cellcolor[rgb]{ .751,  .751,  .751}0.03980 (2.873e-02) \\
    \emph{COMPAS} &  0.00067 (2.494e-03)- & 0.26997 (7.013e-02)- & 0.00516 (6.391e-03)- & \cellcolor[rgb]{ .751,  .751,  .751}0.52135 (6.884e-02) \\
    \emph{LSAT}  &  0.00182 (4.094e-03)- & 0.06179 (3.495e-02)- & 0.00710 (7.592e-03)- & \cellcolor[rgb]{ .751,  .751,  .751}0.54121 (9.452e-02) \\
    \emph{Default} &  0.00000 (0.000e+00)- & 0.54051 (9.128e-02)$\approx$ & 0.03818 (1.387e-02)- & \cellcolor[rgb]{ .751,  .751,  .751}0.57444 (6.706e-02) \\
    \emph{Adult} & 0.00124 (2.214e-03)- & 0.50121 (1.034e-01)- & 0.00558 (6.906e-03)- & \cellcolor[rgb]{ .751,  .751,  .751}0.71014 (6.253e-02) \\
    \emph{Bank}  & 0.00100 (3.000e-03)- & 0.39488 (7.600e-02)- & 0.00059 (1.853e-03)- & \cellcolor[rgb]{ .751,  .751,  .751}0.50912 (6.095e-02) \\
    \emph{Dutch} &  0.00000 (0.000e+00)- & 0.28123 (7.249e-02)- & 0.25025 (6.642e-02)- & \cellcolor[rgb]{ .751,  .751,  .751}0.85461 (4.012e-02) \\
    \bottomrule
    \end{tabular}%
    \end{adjustbox}
\end{table}%

The ``extreme'' models in the final generation generated by $F_{EIG}$ are selected for analysis. More specifically, the best models according to individual metrics (accuracy, $f_I$, $f_G$) are selected for further analysis. The accuracy, $f_I$ and $f_G$ values of those selected models are averaged respectively over 30 trials and reported in Table \ref{tab:Multi_FR}, where the $\Delta$ values mean the differences between the metric values and the best values in the corresponding metric. Table \ref{tab:Multi_FR} clearly shows the extreme tradeoffs among these metrics. Take \emph{Default} as an example, the best performance of accuracy, $f_I$ and $f_G$ for $F_{EIG}$ are 0.82332, 0.02724, and 0.00001, respectively. Then, it's possible for the model with the best accuracy to optimise $f_I$ from 0.06753 to 0.2724 but the model must sacrifice the high accuracy performance for about 6.022e-1, which helps decision-makers to clearly understand the tradeoffs. Thanks to the diverse tradeoff set, decision-makers can make an appropriate decision depending on the demands in real life.

\begin{figure}[htbp]
    \centering
    \includegraphics[width=0.4\textwidth]{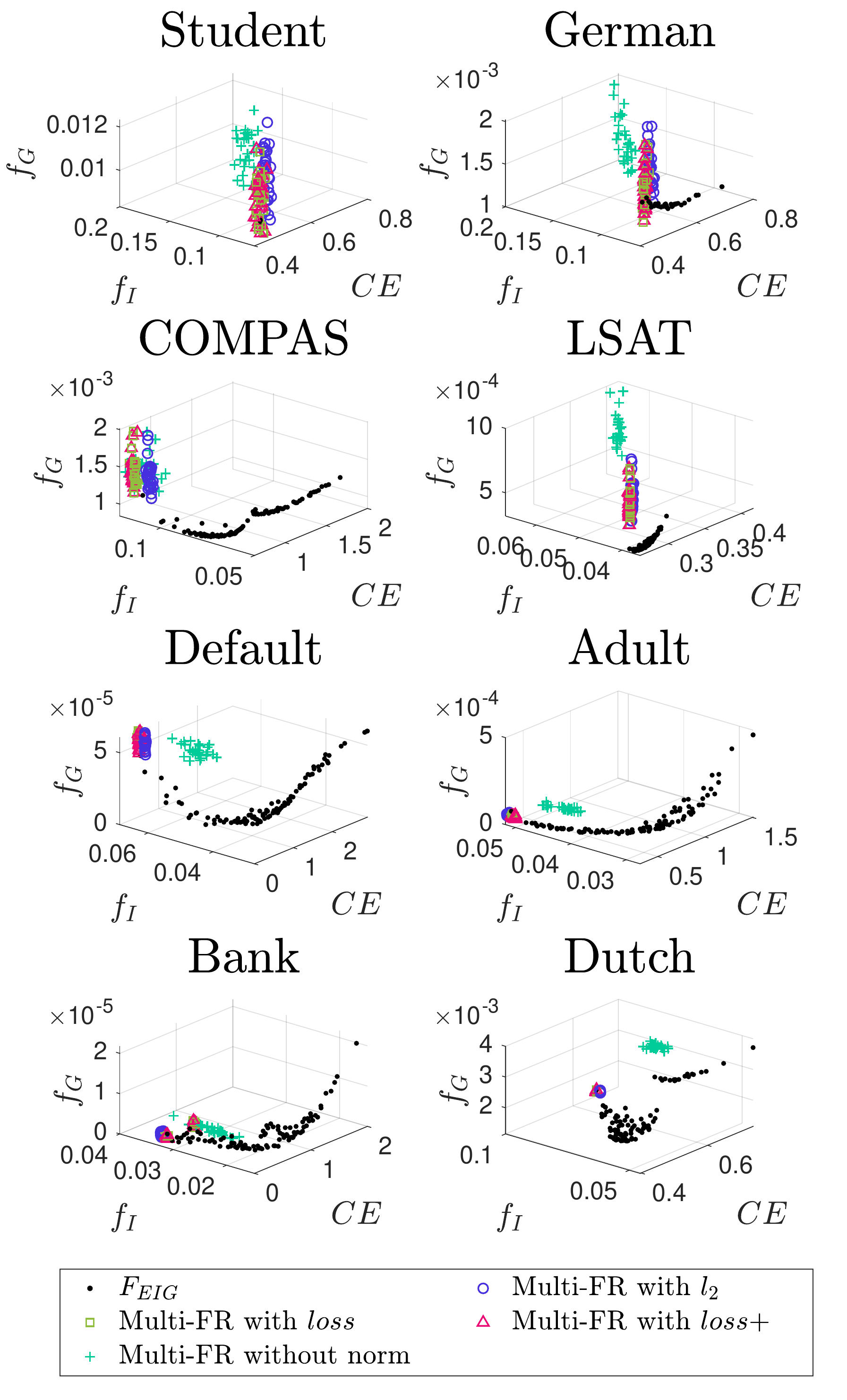}\\
    \caption{\label{fig:cmp_Multi_FR_MOO}
    Illustrative examples:
     (i) the models obtained by different Multi-FRs in all 30 trials (colourful points);
     (ii) non-dominated solutions of the models in the last generation of $F_{EIG}$ in one arbitrary trial (black points).}
\end{figure}

The average values of accuracy, $f_I$, and $f_G$ of models obtained by Multi-FRs using different normalisation methods are also recorded in Table \ref{tab:Multi_FR}. For $f_I$ and $f_G$, $F_{EIG}$ can statistically achieve better performance than all four types of Multi-FRs. Fig. \ref{fig:cmp_Multi_FR_MOO} visualises the model set of $F_{EIG}$ in \emph{one arbitrary} trial and models of four Multi-FRs in \emph{all} trials. Although the weights of the gradients of accuracy, $f_I$, and $f_G$ are adaptively determined by Frank-Wolf Solver method during the training process, the models obtained by Multi-FRs only distribute in a sub-region of those obtained by $F_{EIG}$, which implies that $F_{EIG}$ is able to explore more diverse decision solutions in both $f_I$ and $f_G$ metrics.

$F_{EIG}$ using multi-objective learning can obtain a group of diverse models and outperform the state-of-the-art.

\subsection{Answering $\mathcal{Q}3$ and $\mathcal{Q}4$}\label{sec:Q3_Q4}
In this section, we implement the 9-objective case to answer $\mathcal{Q}3$ and $\mathcal{Q}4$. Three metric sets are considered according to whether the fairness measures belong to the representative measure subset in Table \ref{tab:Fairness_metrics}, Metric Set I ($CE$ $\cup$ Fair1--Fair8), Metric Set II ($CE$ $\cup$ Fair9-Fair16), and Metric Set III ($CE$ $\cup$ Fair1--Fair16). The Metric Set I is the 9 directly optimised objectives. Our ensemble methods are denoted as $F_{Repre}$. The ensemble models obtained by $F_{Repre}$ are denoted as ``Ens''.

\subsubsection{Compared methods}\label{sec:ComparedMethods}
To verify the effectiveness of our ensemble methods in balancing accuracy and multiple fairness metrics, our ensemble learning framework (shown in Algorithm \ref{algo:ensembleframework}) according to four different multi-objective ensemble selection strategies $\pi_{ens}$ are implemented, called EnsBest, EnsAll, EnsKnee and EnsDiv (described in Section \ref{sec:ensemblestrategy}), to optimise the measures in the Metric Set I.

The ensemble method of \cite{Kenfack2021Impact} is used as a baseline for its outstanding performance compared to \cite{cispa2382} according to \cite{Kenfack2021Impact}. In~\cite{Kenfack2021Impact}, different classifiers are considered as base models including Linear Regression (LR), Linear Discriminant Analysis (LDA), K-nearest Neighbors Algorithm (KNN), Naive Bayes (NB), Random Forest (RF), Support Vector Machine Classifier (SVM), and Classification And Regression Trees (CART). Five ways of ensemble combination introduced in~\cite{Kenfack2021Impact} all show their competitive performance in mitigating unfairness and are used as compared methods in our study, described as follows. KCR is an ensemble of KNN, CART and RF. KCS is an ensemble of KNN, CART and SVM. LrKSCR is an ensemble of LR, KNN, SVM, CART and RF. LrKLSCR is an ensemble of LR, KNN, LDA, SVM, CART and RF. KCSRN is an ensemble of KNN, CART, SVM, RF and NB.

Thus, a total of nine algorithms are compared in this work. Fair1--Fair8 are directly optimised based on the formulations in Table~\ref{tab:Fairness_metrics}. Cross entropy is used to measure the accuracy.

\subsubsection{Datasets}
Besides the eight datasets in Table \ref{tab:dataset}, seven new datasets are also considered to answer $\mathcal{Q}3$ and $\mathcal{Q}4$ and to verify the effectiveness of our ensemble method, including \emph{Academics}~\cite{hussain2018educational}, \emph{Heart}~\cite{chicco2020machine}, \emph{Diabetes}~\cite{Dua2019}, \emph{Performance}~\cite{hussain2018classification}, \emph{IBM}~\cite{yang2020ibm}, \emph{Drug}~\cite{fehrman2017five} and \emph{Patient}~\cite{Sadikin2020}. These additional datasets are added to facilitate comparisons with existing work~\cite{Kenfack2021Impact}. In order to construct ensemble models, each dataset is randomly split into 4 partitions, with a ratio of 5:1.25:1.25:2.5, as training, validation, ensemble, and test sets, where the setting is the same as the work~\cite{Yao1998Making}. The use of validation set is the same as that of $\mathcal{D}_{validation}$ in Algorithm \ref{algo:framework}, whilst the ensemble sets are only used in the process of ensemble strategies on the 15 datasets. The selection of sensitive attributes on \emph{German}, \emph{Adult}, \emph{COMPAS}, and \emph{Bank} is, gender and age, gender, race, and age, respectively, which is the same as the work~\cite{anahideh2021choice}. The remainder is all set as gender.

\subsubsection{Parameter settings}
According to the size of datasets, we use two settings for the number of hidden nodes and learning rate. For \emph{Student}, \emph{German}, \emph{COMPAS}, \emph{LSAT}, \emph{Default}, \emph{Adult}, \emph{Bank}, \emph{Dutch} and \emph{Patient}, the individuals of each Ens* are ANNs that are fully connected with one hidden layer of 64 nodes. The learning rate is set as $0.004$. For the remaining datasets, the number of hidden nodes is set as 32 and learning rate is $0.001$. The initialisation method is the same as \cite{glorot2010understanding}. The $\mu$ and $\lambda$ in Algorithm \ref{algo:framework} are $300$. The $\mathcal{E}_{loss}$ in Algorithm \ref{algo:reproductionframework} only contains $CE$ since Fair1--Fair16 cannot be directly used as losses. Larger mutation strength is applied for the 9-objective case to improve the performance of the objectives of Fair1--Fair8. The mutation strength is 0.1 and the batch size is set as 1000. $K$ in Algorithm \ref{algo:reproductionframework} is set as 10. The probabilities of crossover and mutation are all set as to 1. The termination condition is set as a maximum number of $100$ generations. When dealing with the test dataset $\mathcal{D}_{test}$, the four ensemble methods output the arithmetic averaged prediction value over the selected model subset.

For EnsDiv, the 50 base classifiers are selected from the non-dominated solutions in the final population of $F_{Repre}$ using the diversity update strategy of Two$\_$Arch2~\cite{wang2015twoarch2} according to the objectives on the ensemble data. For EnsKnee, the 50 base classifiers are chosen based on the knee point selection of the work~\cite{zhang2015Knee}. For Ens*, the ensemble prediction is the arithmetic average of the selected base ANNs. 

\begin{table*}[htbp]
  \centering
  \caption{Comparing models shown to be non-dominated on the test data and best on at least one of the metrics ($CE$, $f_I$ and $f_G$), in terms of the three metric values averaged over 30 trials. ``NDM'' stands for ``non-dominated models''. ``+/$\approx$/-'' indicates that the average objective value (specified by column header) of corresponding algorithm (specified by row header) is statistically better/similar/worse than the best averaged value (grey backgrounds) according to the Wilcoxon rank sum test with a 0.05 significance level.
  The $\Delta$ values are the differences between the metric values and the best values. }
  \scriptsize
  \setlength{\tabcolsep}{4pt}
    \begin{tabular}{l|ll|ll|ll|ll|ll|ll}
    \toprule
          & \multicolumn{6}{c|}{\emph{Student}}                  & \multicolumn{6}{c}{\emph{German}} \\
          & \multicolumn{1}{c}{Accuracy $\uparrow$} & \multicolumn{1}{c|}{$\Delta$ Accuracy} & \multicolumn{1}{c}{$f_I$ $\downarrow$} & \multicolumn{1}{c|}{$\Delta$ $f_I$} & \multicolumn{1}{c}{$f_G$ $\downarrow$} & \multicolumn{1}{c|}{$\Delta$ $f_G$} & \multicolumn{1}{c}{Accuracy$\uparrow$} & \multicolumn{1}{c|}{$\Delta$ Accuracy} & \multicolumn{1}{c}{$f_I$ $\downarrow$} & \multicolumn{1}{c|}{$\Delta$ $f_I$} & \multicolumn{1}{c}{$f_G$ $\downarrow$} & \multicolumn{1}{c}{$\Delta$ $f_G$} \\
    \midrule
    NDM with best $CE$ & \cellcolor[rgb]{ .751,  .751,  .751}0.78694 &       & 0.06609$\approx$ & 3.47E-04 & 0.00973$\approx$ & 2.90E-04 & \cellcolor[rgb]{ .751,  .751,  .751}0.765 &       & 0.06861- & 9.32E-03 & 0.00167- & 4.07E-04 \\
    NDM with best $f_I$ & 0.78694$\approx$ & -1.11E-16 & \cellcolor[rgb]{ .751,  .751,  .751}0.06574 &       & 0.00956$\approx$ & 1.23E-04 & 0.71100- & -5.40E-02 & \cellcolor[rgb]{ .751,  .751,  .751}0.05929 &       & 0.00129$\approx$ & 3.50E-05 \\
    NDM with best $f_G$ & 0.78243$\approx$ & -4.51E-03 & 0.06614$\approx$ & 3.96E-04 & \cellcolor[rgb]{ .751,  .751,  .751}0.00944 &       & 0.71333- & -5.17E-02 & 0.06108- & 1.79E-03 & \cellcolor[rgb]{ .751,  .751,  .751}0.00126 &  \\
    Multi-FR-$l_2$ & 0.69730- & -8.96E-02 & 0.10487- & 3.91E-02 & 0.01015- & 7.15E-04 & 0.70017- & -6.48E-02 & 0.10344- & 4.42E-02 & 0.00152- & 2.57E-04 \\
    Multi-FR-$loss$ & 0.75901- & -2.79E-02 & 0.07419- & 8.45E-03 & 0.01058- & 1.15E-03 & 0.73833- & -2.67E-02 & 0.07784- & 1.86E-02 & 0.00160- & 3.38E-04 \\
    Multi-FR-$loss+$ & 0.76261- & -2.43E-02 & 0.07417- & 8.43E-03 & 0.01015- & 7.10E-04 & 0.74550- & -1.95E-02 & 0.07699- & 1.77E-02 & 0.00153- & 2.74E-04 \\
    Multi-FR-no-norm & 0.60946- & -1.78E-01 & 0.14922- & 8.35E-02 & 0.01305- & 3.62E-03 & 0.53950- & -2.26E-01 & 0.16099- & 1.02E-01 & 0.00243- & 1.17E-03 \\
    \midrule

    \midrule
    \multicolumn{1}{r|}{} & \multicolumn{6}{c|}{\emph{COMPAS}}                   & \multicolumn{6}{c}{\emph{LSAT}} \\
          & Accuracy $\uparrow$ & \multicolumn{1}{c|}{$\Delta$ Accuracy} & \multicolumn{1}{c}{$f_I$ $\downarrow$} & \multicolumn{1}{c|}{$\Delta$ $f_I$} & \multicolumn{1}{c}{$f_G$ $\downarrow$} & \multicolumn{1}{c|}{$\Delta$ $f_G$} & \multicolumn{1}{c}{Accuracy $\uparrow$} & \multicolumn{1}{c|}{$\Delta$ Accuracy} & \multicolumn{1}{c}{$f_I$ $\downarrow$} & \multicolumn{1}{c|}{$\Delta$ $f_I$} & \multicolumn{1}{c}{$f_G$ $\downarrow$} & \multicolumn{1}{c}{$\Delta$ $f_G$} \\
    \midrule
    NDM with best $CE$ & 0.65046- & -6.10E-03 & 0.10443- & 5.03E-02 & 0.00107- & 2.42E-04 & 0.89959- & -7.80E-04 & 0.03848- & 2.81E-03 & 0.00035- & 2.14E-05 \\
    NDM with best $f_I$ & 0.47109- & -1.86E-01 & \cellcolor[rgb]{ .751,  .751,  .751}0.05416 &       & 0.00136- & 5.23E-04 & 0.89751- & -2.87E-03 & \cellcolor[rgb]{ .751,  .751,  .751}0.03566 &       & 0.00053- & 2.08E-04 \\
    NDM with best $f_G$ & 0.58070- & -7.59E-02 & 0.08173- & 2.76E-02 & \cellcolor[rgb]{ .751,  .751,  .751}0.00083 &       & 0.89849- & -1.89E-03 & 0.03797- & 2.31E-03 & \cellcolor[rgb]{ .751,  .751,  .751}0.00033 &  \\
    Multi-FR-$l_2$ & 0.64651- & -1.01E-02 & 0.10622- & 5.21E-02 & 0.00149- & 6.62E-04 & 0.89225- & -8.12E-03 & 0.04026- & 4.60E-03 & 0.00071- & 3.79E-04 \\
    Multi-FR-$loss$ & 0.65630$\approx$ & -2.53E-04 & 0.11461- & 6.04E-02 & 0.00115- & 3.21E-04 & 0.90013$\approx$ & -2.45E-04 & 0.03826- & 2.59E-03 & 0.00035- & 2.64E-05 \\
    Multi-FR-$loss+$ & \cellcolor[rgb]{ .751,  .751,  .751}0.65656 &       & 0.11555- & 6.14E-02 & 0.00108- & 2.43E-04 & \cellcolor[rgb]{ .751,  .751,  .751}0.90037- &       & 0.03843- & 2.77E-03 & 0.00035- & 2.35E-05 \\
    Multi-FR-no-norm & 0.51798- & -1.39E-01 & 0.11164- & 5.75E-02 & 0.00311- & 2.28E-03 & 0.87291- & -2.75E-02 & 0.06372- & 2.81E-02 & 0.00058- & 2.51E-04 \\
    \midrule

    \midrule
    \multicolumn{1}{r|}{} & \multicolumn{6}{c|}{\emph{Default}}                  & \multicolumn{6}{c}{\emph{Adult}} \\
          & Accuracy $\uparrow$ & \multicolumn{1}{c|}{$\Delta$ Accuracy} & \multicolumn{1}{c}{$f_I$ $\downarrow$} & \multicolumn{1}{c|}{$\Delta$ $f_I$} & \multicolumn{1}{c}{$f_G$ $\downarrow$} & \multicolumn{1}{c|}{$\Delta$ $f_G$} & \multicolumn{1}{c}{Accuracy $\uparrow$} & \multicolumn{1}{c|}{$\Delta$ $f_G$} & \multicolumn{1}{c}{$f_I$ $\downarrow$} & \multicolumn{1}{c|}{$\Delta$ $f_I$} & \multicolumn{1}{c}{$f_G$ $\downarrow$} & \multicolumn{1}{c}{$\Delta$ $f_G$} \\
    \midrule
    NDM with best $CE$ & \cellcolor[rgb]{ .751,  .751,  .751}0.82332 &       & 0.06753- & 4.03E-02 & 0.00004- & 3.08E-05 & 0.85684- & -1.33E-03 & 0.05066- & 2.32E-02 & 0.00008- & 5.31E-05 \\
    NDM with best $f_I$ & 0.22117- & -6.02E-01 & \cellcolor[rgb]{ .751,  .751,  .751}0.02724 &       & 0.00006- & 4.85E-05 & 0.24902- & -6.09E-01 & \cellcolor[rgb]{ .751,  .751,  .751}0.02745 &       & 0.00046- & 4.28E-04 \\
    NDM with best $f_G$ & 0.73987- & -8.35E-02 & 0.04736- & 2.01E-02 & \cellcolor[rgb]{ .751,  .751,  .751}0.00001 &       & 0.80414- & -5.40E-02 & 0.04095- & 1.35E-02 & \cellcolor[rgb]{ .751,  .751,  .751}0.00003 &  \\
    Multi-FR-$l_2$ & 0.82193- & -1.39E-03 & 0.06623- & 3.90E-02 & 0.00005- & 4.51E-05 & 0.85130- & -6.86E-03 & 0.05144- & 2.40E-02 & 0.00006- & 2.53E-05 \\
    Multi-FR-$loss$ & 0.82008- & -3.24E-03 & 0.06833- & 4.11E-02 & 0.00005- & 4.61E-05 & 0.85721- & -9.56E-04 & 0.05048- & 2.30E-02 & 0.00008- & 4.82E-05 \\
    Multi-FR-$loss+$ & 0.82261- & -7.11E-04 & 0.06782- & 4.06E-02 & 0.00005- & 4.04E-05 & \cellcolor[rgb]{ .751,  .751,  .751}0.85816- &       & 0.05027- & 2.28E-02 & 0.00008- & 4.78E-05 \\
    Multi-FR-no-norm & 0.52346- & -3.00E-01 & 0.05327- & 2.60E-02 & 0.00012- & 1.07E-04 & 0.65512- & -2.03E-01 & 0.04723- & 1.98E-02 & 0.00020- & 1.66E-04 \\
    \midrule

    \midrule
    \multicolumn{1}{r|}{} & \multicolumn{6}{c|}{\emph{Bank}}                     & \multicolumn{6}{c}{\emph{Dutch}} \\
          & Accuracy $\uparrow$ & \multicolumn{1}{c|}{$\Delta$ Accuracy} & \multicolumn{1}{c}{$f_I$ $\downarrow$} & \multicolumn{1}{c|}{$\Delta$ $f_I$} & \multicolumn{1}{c}{$f_G$ $\downarrow$} & \multicolumn{1}{c|}{$\Delta$ $f_G$} & \multicolumn{1}{c}{Accuracy $\uparrow$} & \multicolumn{1}{c|}{$\Delta$ Accuracy} & \multicolumn{1}{c}{$f_I$ $\downarrow$} & \multicolumn{1}{c|}{$\Delta$ $f_I$} & \multicolumn{1}{c}{$f_G$ $\downarrow$} & \multicolumn{1}{c}{$\Delta$ $f_G$} \\
    \midrule
    NDM with best $CE$ & \cellcolor[rgb]{ .751,  .751,  .751}0.90049 &       & 0.03344- & 1.86E-02 & 0.00000- & 1.27E-06 & 0.82605- & -8.17E-04 & 0.06346- & 1.78E-02 & 0.00343- & 2.30E-03 \\
    NDM with best $f_I$ & 0.12660- & -7.74E-01 & \cellcolor[rgb]{ .751,  .751,  .751}0.0148 &       & 0.00002- & 2.17E-05 & 0.57309- & -2.54E-01 & \cellcolor[rgb]{ .751,  .751,  .751}0.04566 &       & 0.00371- & 2.59E-03 \\
    NDM with best $f_G$ & 0.82482- & -7.57E-02 & 0.02541- & 1.06E-02 & \cellcolor[rgb]{ .751,  .751,  .751}0.00000 &       & 0.70398- & -1.23E-01 & 0.08550- & 3.99E-02 & \cellcolor[rgb]{ .751,  .751,  .751}0.00113 &  \\
    Multi-FR-$l_2$ & 0.89686- & -3.64E-03 & 0.03436- & 1.96E-02 & 0.00000- & 7.11E-07 & 0.82686$\approx$ & -1.38E-05 & 0.06219- & 1.65E-02 & 0.00340- & 2.27E-03 \\
    Multi-FR-$loss$ & 0.89967$\approx$ & -8.20E-04 & 0.03343- & 1.86E-02 & 0.00000- & 2.00E-06 & \cellcolor[rgb]{ .751,  .751,  .751}0.82687 &       & 0.06327- & 1.76E-02 & 0.00346- & 2.34E-03 \\
    Multi-FR-$loss+$ & 0.89984$\approx$ & -6.56E-04 & 0.03332- & 1.85E-02 & 0.00000- & 1.65E-06 & 0.82675$\approx$ & -1.16E-04 & 0.06338- & 1.77E-02 & 0.00346- & 2.33E-03 \\
    Multi-FR-no-norm & 0.51696- & -3.84E-01 & 0.02963- & 1.48E-02 & 0.00006- & 5.92E-05 & 0.61590- & -2.11E-01 & 0.07467- & 2.90E-02 & 0.00275- & 1.62E-03 \\
    \bottomrule
    \end{tabular}%
  \label{tab:Multi_FR}%
\end{table*}%

For KCR, KCS, LrKSCR, LrKLSCR, and KCSRN, the setting in the original study~\cite{Kenfack2021Impact} is used. All the base models are implemented by scikit-learn~\cite{pedregosa2011scikit}. The weights among base classifiers are determined by two types of weights, manual-based weights $e$ and metric-based weights $t$ on the ensemble data. For manual-based weights, all the objectives are treated equally ($e=1/9$), which is the same as the original setting~\cite{Kenfack2021Impact}. Soft majority voting is applied to determine metric-based weights among the base classifiers since the experiment results of \cite{Kenfack2021Impact} show that soft majority voting performed better than hard majority voting to balance different metrics. Four-fold cross-validation is applied. For each compared method, 30 independent trials are performed.

\subsubsection{Performance measures}
HV is used to evaluate the overall performance of a model set in terms of convergence and diversity, and the detail is introduced in Section~\ref{sec:performan_Q1_Q2}. G-mean (Geometric mean)~\cite{derringer1994balancing} is used to measure the performance of an ensemble in terms of accuracy and Fair1--Fair8 since G-mean can measure a solution considering multiple objectives with different units and is widely used in many applications~\cite{derringer1994balancing, vieira2012optimisation}. In the calculation of G-mean, the ``Accuracy'' metric is equal to (1-accuracy) in order to make all the measures in accuracy and Fair1--Fair8 be minimised. The smaller G-mean value means the better performance.

\noindent\textbf{($\mathcal{Q}3$) \emph{Can multi-objective learning improve all fairness measures including those not used in model training?}}

To answer $\mathcal{Q}3$, we will plot and analyse the convergence curves of HV values of the 9-objective optimisation algorithm $F_{Repre}$ on the test data of the 15 datasets.

We use $F_{Repre}$ to directly optimise Metric Set I, and then calculate the average HV values of every 10 generations in the Metric Set I, the Metric Set II and the Metric Set III, respectively, on the test data over 30 trials. The way of determining pseudo Pareto front described in Section \ref{sec:performan_Q1_Q2} is used except that the considered objectives are the Metric Set I, the Metric Set II and the Metric Set III, respectively.

\begin{figure}[htbp]
	\centering  
	\includegraphics[width=0.6\linewidth]{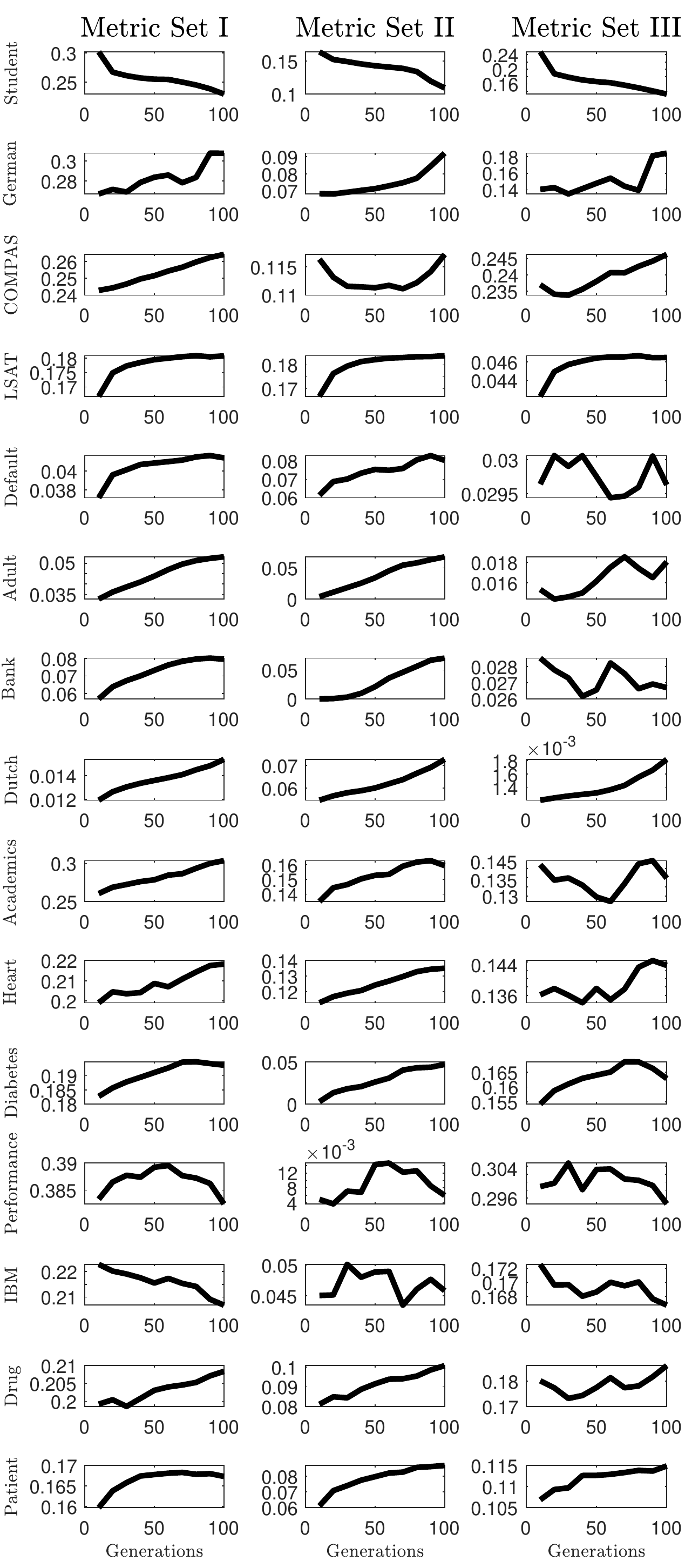}
		
    \caption{\label{fig:HV_represent}HV values of models in Metric Sets I-III, respectively, averaged over 30 trials on 15 datasets by applying our algorithm to directly optimise the Metric Set I. \emph{x-axis}: generation number; \emph{y-axis}: HV value.}
\end{figure}

Fig. \ref{fig:HV_represent} plots the convergence curves of HV values according to the three metric sets respectively. As the Metric Set I is optimised by $F_{Repre}$, the HV values of $F_{Repre}$ become better (increase) along with the evolution process except for \emph{Student}, \emph{Performance} and \emph{IBM}. A possible reason for the decrease of \emph{Student} in the Metric Sets I-III is overfitting, since \emph{Student} is a small dataset. According to the Metric Set II, the HV values improve (increase) after 100 generations on all datasets except for \emph{Student}, \emph{Performance}, and \emph{IBM}. This is worth noting because Fair9-16 have never been used anywhere during model training. Yet models trained according to Fair1--Fair8 can still perform well according to Fair9-16. It is clear that $F_{Repre}$ using multi-objective learning can improve all fairness measures including those not used in model training. 

However, some interesting observations are made by examining the column under Metric Set III in Fig. \ref{fig:HV_represent}. On the \emph{Default}, \emph{Bank}, \emph{Academics} and \emph{Diabetes} datasets, HV values are both increasing according to Metric Set I and Metric Set II, but decreasing according to Metric Set III, which is a union of Metric Set I and Metric Set II. It's one of our future work to investigate this phenomenon further.

\noindent\textbf{($\mathcal{Q}4$) \emph{Can multi-objective learning generate an ensemble model consisting of base models to balance accuracy and multiple fairness measures better?}}

To answer $\mathcal{Q}4$, three perspectives are considered: evaluation of base models, evaluation based on G-mean, and comparison according to the accuracy and multiple fairness measures.

\begin{figure}[htbp]
	\centering  
	\includegraphics[width=1\linewidth]{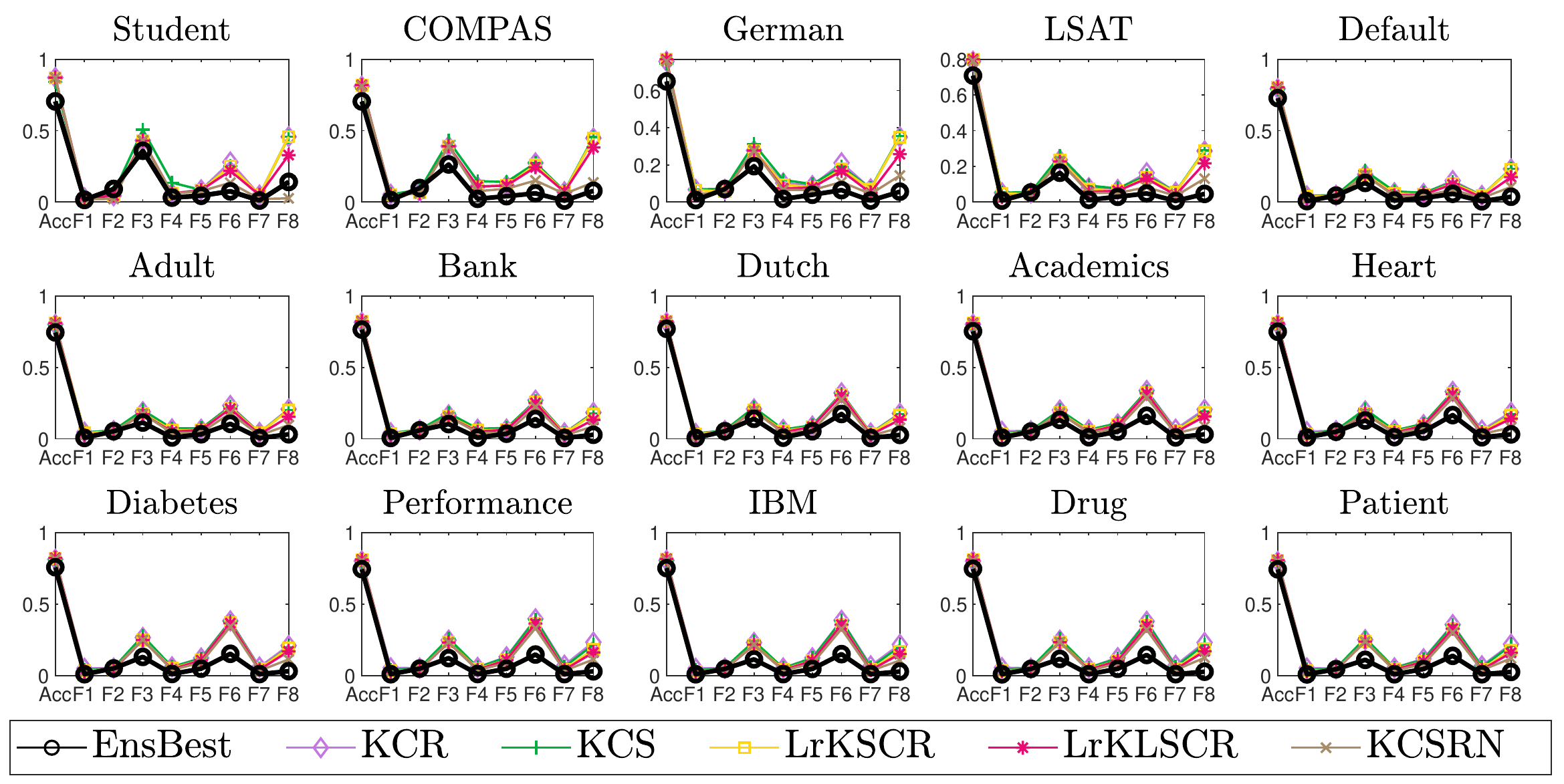}
    \caption{\label{fig:best_base} 
    The best performance of the whole base model set in terms of each measure of Accuracy and Fair1--Fair8 (abbreviated as ``F1--F8'') over 30 trials. 
    }
\end{figure}

The quality of base models in terms of accuracy and Fair1--Fair8 on the test set are analysed from two perspectives: the visualisation of base models and HV values of base models. We record the best performance of the whole base model set in terms of accuracy and Fair1--Fair8 for every trial. Then, each averaged measure value on 30 trials is plotted in Fig. \ref{fig:best_base} for EnsBest, KCR, KCS, LrKSCR, LrKLSCR and KCSRN. Since the base models of EnsBest have the lowest number of base models than other Ens*, we only plot the performance of the base model sets of EnsBest to clearly demonstrate the advantages of using multi-objective learning. It's observed that EnsBest is better than KCR, KCS, LrKSCR, LrKLSCR and KCSRN in terms of fairness. 
Regarding EnsBest that trains base models considering accuracy and the fairness measures through multi-objective learning, base models with better performance or even the optimal performance in terms of Fair1--Fair8 can be found since most values of fairness measures are close to 0, as shown in Fig. \ref{fig:best_base}.

Furthermore, HV is used to evaluate the overall performance of the base model sets obtained by different algorithms in terms of Accuracy and Fair1--Fair8, where the value 1-Accuracy is used as the accuracy objective value to make all nine objectives be minimised in the calculation process of HV. Specifically, for each dataset, the pseudo Pareto front based on Accuracy and Fair1--Fair8 objectives is obtained from all the compared algorithms on 30 trials. Then, HV values of the base model sets are calculated using pseudo Pareto front for each trial. The averaged HV values on 30 trials for 15 datasets are shown in Table \ref{tab:base_HV}, where ``$+/\approx/-$'' indicates the averaged HV values of corresponding algorithms are statistically better/similar/worse than that of \emph{EnsBest} according to the Wilcoxon rank sum test with a 0.05 significance level. For KCR, KCS, LrKSCR, LrKLSCR and KCSRN, we also highlight the values that are statistically better than those of EnsBest using underline ``\_''. As Table \ref{tab:base_HV} shows, the base model set of EnsBest can achieve better tradeoffs than those of KCR, KCS, LrKSCR, LrKLSCR and KCSRN on 10 out of 15 datasets including \emph{Student}, \emph{German}, \emph{COMPAS}, \emph{LSAT}, \emph{Default}, \emph{Dutch}, \emph{Academic}, \emph{Diabetes}, \emph{IBM} and \emph{Drug}. This means that the base models obtained by multi-objective learning can have better quality than other approaches in terms of fairness measures, which can contribute to generating an ensemble with better performance of Accuracy and Fair1--Fair8. Table \ref{tab:base_HV} also shows that EnsBest is a baseline for our ensemble because EnsDiv, EnsKnee and EnsAll are all better than EnsBest in all cases based on HV values. They are all better than KCR, KCS, LrKSCR, LrKLSCR and KCSRN.

As to the overall performance of the ensemble models considering Accuracy and Fair1--Fair8, Table \ref{tab:G_mean} gives the G-mean values on the 15 datasets averaged over 30 trials. The last row of Table \ref{tab:G_mean} also shows the overall rankings of the nine algorithms on the 15 datasets. ``$+/\approx/-$'' indicates that the average G-mean values of the corresponding algorithm (specified by column header) is statistically better/similar/worse than the one of \emph{EnsBest} according to the Wilcoxon rank sum test with a 0.05 significance level. It's observed that EnsBest has the best averaged ranking 3.47 among all the compared algorithms. EnsBest outperforms KCR, KCS, LrKSCR, LrKLSCR and KCSRN on 11 out of 15 datasets except \emph{Adult}, \emph{Bank}, \emph{Heart} and \emph{Diabetes}. A closer examination of the datasets and the fairness measures reveal that the poor performance of EnsBest on \emph{Adult}, \emph{Bank} is partially caused by imbalanced data distribution~\cite{6170916,onan2019consensus}, in the datasets.

\begin{figure*}[htbp]
	\centering  
	\includegraphics[width=0.8\linewidth]{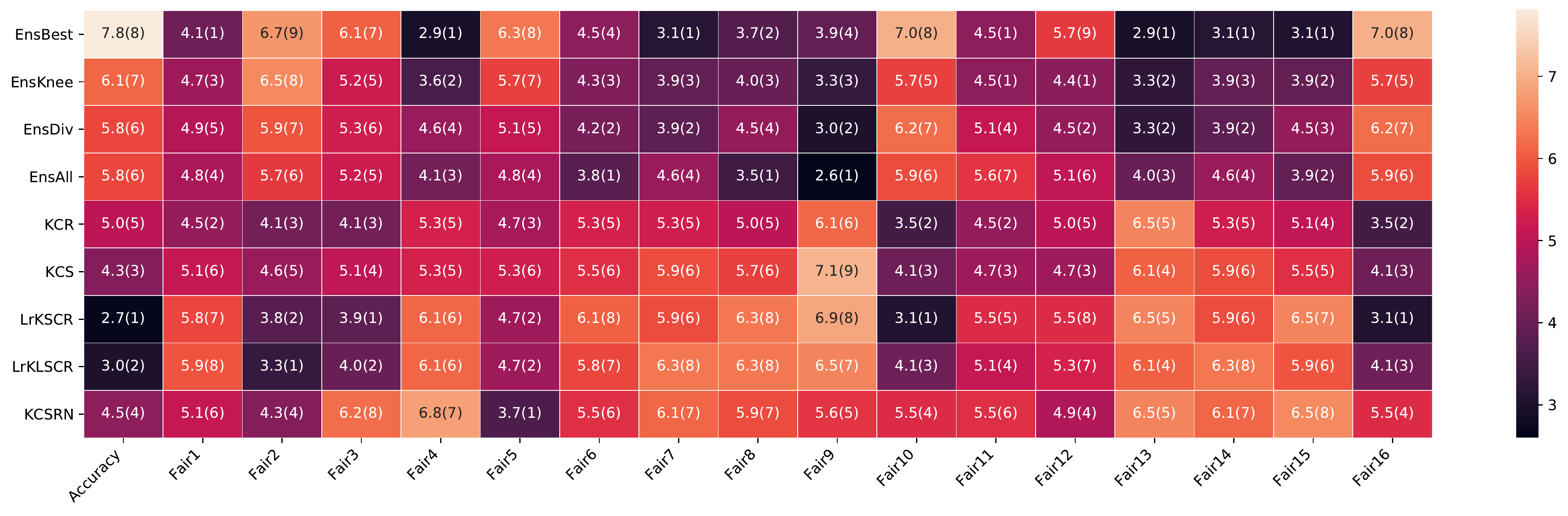}
    \caption{
    The averaged rankings of the nine algorithms in terms of each measure in Accuracy and Fair1--Fair16 on the 15 datasets, where the smaller value means the better performance. The numbers in brackets indicate the ranks of the algorithms based on the average ranking values. 
    }
	\label{fig:heatmap}
\end{figure*}

We take a closer analysis of the performance of the 9 algorithms in terms of each measure in Accuracy and Fair1--Fair16. Fig. \ref{fig:heatmap} ranks the 9 algorithms according to the averaged values of each measure in terms of Accuracy and Fair1--Fair16 on 15 datasets, where the smaller ranking value means the better performance. According to Fig. \ref{fig:heatmap}, several observations can be made. First, no method is best across all metrics, which is expected because of inherent conflicts among metrics. If a method excels at one metric, it is very likely to achieve sub-optimal values for other conflicting metrics. Second, if we consider the overall performance among all 17 metrics, our EnsBest is the best, which achieves the best ranking according to the most number of metrics and also achieves good rankings on other metrics. In fact, our four ensemble methods, Ens*, all achieve better overall performance than others. Third, if we examine individual objectives, including the accuracy and 16 fairness metrics, separately, Fig. \ref{fig:heatmap} shows that our methods Ens* achieve the best performances according to fairness metrics 1, 4, 6-9 and 11-15. Ens* do not perform as well as others on the accuracy and fairness metrics 2-3, 5, 10 and 16. In short, our methods performed the best on 11 out of 17 metrics, keeping in mind that our methods also have the best overall performance according to G-means.

Note that these 8 fairness metrics, Fair1--Fair8, were used to answer $\mathcal{Q}3$, not specifically used to address the balance between group and individual fairness. By considering group fairness metrics only (Fair1--Fair8), we will compromise our performance on individual fairness metrics, as shown by other papers~\cite{pessach2022review, mehrabi2021survey, speicher2018unified}. There is an inherent conflict between group and individual fairness, which is also evident from our previous work~\cite{QingquanFair2021} and from our experimental results related to $\mathcal{Q}1$ and $\mathcal{Q}2$. It is important to maintain the balance among different metrics. If we are to consider all possible fairness metrics, we should select representative metrics from different categories, e.g., group and individual fairness, as objectives in our multi-objective ensemble learning framework.

\begin{table*}[htbp]
  \centering
  \caption{HV values of base models averaged over 30 trials. ``+/$\approx$/-'' indicates that the average HV values of the corresponding algorithm (specified by column header) are statistically better/similar/worse than the one of EnsBest according to the Wilcoxon rank sum test with a 0.05 significance level. Larger HV values imply better performance. For KCR, KCS, LrKSCR, LrKLSCR and KCSRN, the values that are statistically better than those of EnsBest are highlighted using underline ``\_''}
  \begin{adjustbox} {max width=\linewidth}
    \begin{tabular}{lcccc|ccccc}
    \toprule
          & EnsBest & EnsDiv& EnsKnee & EnsAll & KCR   & KCS   & LrKSCR & LrKLSCR & KCSRN \\
    \midrule
    \emph{Student} & 0.07087(4.136e-02) & 0.11891(5.349e-02)+ & 0.14114(5.073e-02)+ & 0.14102(5.067e-02)+ & 0.02821(2.733e-02)- & 0.02967(2.162e-02)- & 0.03332(2.502e-02)- & 0.03310(2.534e-02)- & 0.04723(2.411e-02)- \\
    \emph{German} & 0.13198(7.326e-02) & 0.20712(5.169e-02)+ & 0.18956(5.134e-02)+ & 0.25920(3.588e-02)+ & 0.02775(1.133e-02)- & 0.01990(7.655e-03)- & 0.02831(1.129e-02)- & 0.02847(1.126e-02)- & 0.02780(1.133e-02)- \\
    \emph{COMPAS} & 0.08805(1.419e-02) & 0.15806(6.997e-03)+ & 0.15354(8.135e-03)+ & 0.19568(5.157e-03)+ & 0.02145(2.230e-03)- & 0.02264(2.240e-03)- & 0.02353(2.232e-03)- & 0.03099(1.920e-03)- & 0.03444(1.695e-03)- \\
    \emph{LSAT} & 0.44313(5.875e-02) & 0.55692(1.992e-02)+ & 0.56563(1.951e-02)+ & 0.63065(1.175e-02)+ & 0.27294(2.308e-02)- & 0.27841(2.321e-02)- & 0.29003(2.035e-02)- & 0.30081(2.180e-02)- & 0.29652(2.340e-02)- \\
    \emph{Default} & 0.04468(1.043e-02) & 0.07612(9.460e-03)+ & 0.08152(7.598e-03)+ & 0.09964(8.413e-03)+ & 0.02610(1.246e-03)- & 0.03329(6.854e-04)- & 0.03763(1.061e-03)- & 0.03784(1.028e-03)- & 0.03744(1.069e-03)- \\
    \emph{Adult} & 0.02098(5.632e-03) & 0.04101(2.569e-03)+ & 0.04963(3.097e-03)+ & 0.05858(2.977e-03)+ & 0.02256(1.001e-03)$\approx$ & \underline{0.02587(2.708e-04)+} & \underline{0.03198(7.392e-04)+} & \underline{0.03225(7.494e-04)+} & \underline{0.03024(7.163e-04)+} \\
    \emph{Bank} & 0.08969(8.740e-02) & 0.13873(8.026e-02)+ & 0.15482(8.279e-02)+ & 0.18006(8.082e-02)+ & 0.08771(1.018e-02)- & 0.06071(7.692e-03)- & \underline{0.09498(1.030e-02)+} & \underline{0.09521(1.025e-02)+} & \underline{0.09048(1.007e-02)+} \\
    \emph{Dutch} & 0.01914(3.560e-03) & 0.03493(2.063e-03)+ & 0.03436(1.423e-03)+ & 0.04665(8.133e-04)+ & 0.00628(1.275e-04)- & 0.00605(1.244e-04)- & 0.00671(1.109e-04)- & 0.00685(1.100e-04)- & 0.01756(5.472e-05)- \\
    \emph{Academics} & 0.17108(6.375e-02) & 0.33458(3.345e-02)+ & 0.34767(3.128e-02)+ & 0.40557(2.466e-02)+ & 0.02253(4.664e-03)- & 0.04858(1.050e-02)- & 0.05165(1.070e-02)- & 0.05372(1.232e-02)- & 0.05950(1.062e-02)- \\
    \emph{Heart} & 0.11956(4.245e-02) & 0.31770(4.978e-02)+ & 0.31612(4.716e-02)+ & 0.38580(4.107e-02)+ & \underline{0.13913(3.128e-02)+} & 0.11311(1.424e-02)$\approx$ & \underline{0.14850(2.784e-02)+} & \underline{0.14929(2.695e-02)+} & \underline{0.14852(2.747e-02)+} \\
    \emph{Diabetes} & 0.07269(2.898e-02) & 0.23022(2.232e-02)+ & 0.19034(2.782e-02)+ & 0.30311(1.346e-02)+ & 0.00481(1.696e-03)- & 0.00459(1.669e-03)- & 0.00486(1.693e-03)- & 0.00486(1.698e-03)- & 0.00486(1.695e-03)- \\
    \emph{Performance} & 0.12993(4.199e-02) & 0.23985(2.734e-02)+ & 0.27498(2.553e-02)+ & 0.28569(2.333e-02)+ & 0.01098(9.154e-03)- & 0.01219(2.264e-03)- & \underline{0.15422(1.372e-03)+} & \underline{0.15532(1.145e-03)+} & 0.06451(5.413e-03)- \\
    \emph{IBM} & 0.45228(5.264e-02) & 0.58949(2.691e-02)+ & 0.60951(2.499e-02)+ & 0.64891(1.464e-02)+ & 0.39724(1.445e-02)- & 0.39231(1.462e-02)- & 0.40309(1.377e-02)- & 0.40474(1.353e-02)- & 0.41625(1.331e-02)- \\
    \emph{Drug} & 0.07909(2.621e-02) & 0.20507(1.061e-02)+ & 0.18995(9.031e-03)+ & 0.25454(7.465e-03)+ & 0.02448(4.366e-03)- & 0.02515(4.170e-03)- & 0.02685(4.124e-03)- & 0.02697(4.161e-03)- & 0.03246(2.811e-03)- \\
    \emph{Patient} & 0.24200(3.961e-02) & 0.41055(1.743e-02)+ & 0.41297(1.644e-02)+ & 0.47878(1.220e-02)+ & 0.24858(2.166e-02)$\approx$ & 0.20988(1.998e-02)- & 0.26058(1.887e-02)$\approx$ & \underline{0.26900(1.616e-02)+} & 0.25652(2.137e-02)$\approx$ \\
    \bottomrule
    \end{tabular}%
  \label{tab:base_HV}%
  \end{adjustbox}
\end{table*}%

\begin{table*}[htbp]
  \centering
  \caption{G-mean values of final solutions averaged over 30 trials in terms of accuracy and Fair1--Fair8. Smaller G-mean values imply better performance. The best and second best averaged G-mean values are highlighted in grey and light grey backgrounds, respectively. ``+/$\approx$/-'' indicates that the average G-mean values of corresponding algorithm (specified by column header) is statistically better/similar/worse than the one of EnsBest according to the Wilcoxon rank sum test with a 0.05 significance level. }
  \begin{adjustbox} {max width=\linewidth}
    \begin{tabular}{lcccc|ccccc}
    \toprule
          & EnsBest & EnsKnee &  EnsDiv& EnsAll & KCR   & KCS   & LrKSCR & LrKLSCR & KCSRN \\
    \midrule
    \emph{Student} & 0.06743(7.827e-02) & \cellcolor[rgb]{ .751,  .751,  .751}0.04870(7.008e-02)$\approx$ & \cellcolor[rgb]{ .880,  .880,  .880}0.05078(7.902e-02)$\approx$ & \cellcolor[rgb]{ .880,  .880,  .880}0.05078(7.902e-02)$\approx$ & 0.16111(5.758e-02)- & 0.18530(5.572e-02)- & 0.20432(3.829e-02)- & 0.19982(7.503e-02)- & 0.16271(7.268e-02)- \\
    \emph{German} & 0.17634(1.078e-01) & \cellcolor[rgb]{ .880,  .880,  .880}0.09392(1.101e-01)+ & 0.18414(1.024e-01)$\approx$ & \cellcolor[rgb]{ .751,  .751,  .751}0.02445(6.327e-02)+ & 0.20332(3.859e-02)$\approx$ & 0.20705(4.872e-02)$\approx$ & 0.21524(6.282e-02)$\approx$ & 0.25669(5.590e-02)- & 0.20502(2.435e-02)$\approx$ \\
    \emph{COMPAS} & \cellcolor[rgb]{ .751,  .751,  .751}0.07208(2.787e-02) & 0.11051(1.879e-02)- & 0.11138(1.398e-02)- & 0.11384(1.636e-02)- & 0.11561(7.120e-03)- & 0.11822(5.424e-03)- & 0.11532(8.332e-03)- & \cellcolor[rgb]{ .880,  .880,  .880}0.09891(8.493e-03)- & 0.13322(2.651e-03)- \\
    \emph{LSAT} & \cellcolor[rgb]{ .751,  .751,  .751}0.00919(1.388e-02) & 0.03568(1.571e-02)- & 0.03804(1.537e-02)- & 0.03182(1.463e-02)- & \cellcolor[rgb]{ .880,  .880,  .880}0.02638(7.756e-03)- & 0.02898(9.322e-03)- & 0.03030(6.060e-03)- & 0.02639(7.372e-03)- & 0.04452(4.566e-03)- \\
    \emph{Default} & \cellcolor[rgb]{ .751,  .751,  .751}0.02455(9.449e-03) & \cellcolor[rgb]{ .880,  .880,  .880}0.02622(4.923e-03)$\approx$ & 0.02860(5.085e-03)- & 0.03448(6.870e-03)- & 0.03164(6.005e-03)- & 0.03611(3.482e-03)- & 0.03805(4.746e-03)- & 0.03986(5.023e-03)- & 0.04071(2.922e-03)- \\
    \emph{Adult} & 0.19218(1.899e-02) & 0.18492(1.793e-02)$\approx$ & 0.18986(1.610e-02)$\approx$ & 0.17877(1.344e-02)+ & 0.12679(8.833e-03)+ & \cellcolor[rgb]{ .880,  .880,  .880}0.11673(1.243e-02)+ & \cellcolor[rgb]{ .751,  .751,  .751}0.11157(1.122e-02)+ & 0.11838(9.092e-03)+ & 0.12805(7.267e-03)+ \\
    \emph{Bank} & 0.09690(3.534e-02) & 0.13031(2.447e-02)- & 0.13636(2.233e-02)- & 0.12787(2.223e-02)- & 0.07475(1.784e-02)+ & 0.07546(1.979e-02)+ & \cellcolor[rgb]{ .751,  .751,  .751}0.07053(1.636e-02)+ & \cellcolor[rgb]{ .880,  .880,  .880}0.07053(1.268e-02)+ & 0.08848(1.603e-02)+ \\
    \emph{Dutch} & 0.09510(3.156e-02) & \cellcolor[rgb]{ .880,  .880,  .880}0.08531(1.972e-02)$\approx$ & 0.08532(1.962e-02)$\approx$ & \cellcolor[rgb]{ .751,  .751,  .751}0.07362(1.999e-02)+ & 0.10588(1.436e-03)$\approx$ & 0.11782(1.252e-03)- & 0.10988(2.196e-03)$\approx$ & 0.11011(2.165e-03)$\approx$ & 0.09494(9.509e-04)$\approx$ \\
    \emph{Academics} & 0.18193(9.131e-02) & \cellcolor[rgb]{ .880,  .880,  .880}0.11329(8.290e-02)+ & 0.12501(8.720e-02)+ & \cellcolor[rgb]{ .751,  .751,  .751}0.09556(8.499e-02)+ & 0.18489(1.348e-01)$\approx$ & 0.19449(1.505e-01)$\approx$ & 0.18436(1.555e-01)$\approx$ & 0.23502(1.123e-01)- & 0.16211(7.398e-02)$\approx$ \\
    \emph{Heart} & 0.12058(7.786e-02) & 0.14403(7.821e-02)$\approx$ & 0.15138(5.969e-02)$\approx$ & 0.18015(4.897e-02)- & 0.10215(5.231e-02)$\approx$ & 0.10041(5.251e-02)+ & \cellcolor[rgb]{ .880,  .880,  .880}0.09086(4.532e-02)+ & \cellcolor[rgb]{ .751,  .751,  .751}0.08610(4.586e-02)+ & 0.13920(1.705e-02)$\approx$ \\
    \emph{Diabetes} & 0.18120(1.080e-01) & 0.24509(3.742e-02)$\approx$ & 0.24451(5.977e-02)- & 0.23195(3.181e-02)$\approx$ & \cellcolor[rgb]{ .751,  .751,  .751}0.09112(6.978e-02)+ & 0.14887(3.445e-02)+ & 0.13009(6.273e-02)+ & \cellcolor[rgb]{ .880,  .880,  .880}0.12013(8.931e-02)+ & 0.13973(7.901e-02)+ \\
    \emph{Performance} & \cellcolor[rgb]{ .751,  .751,  .751}0.09323(4.717e-02) & 0.10742(4.310e-02)$\approx$ & 0.10530(4.653e-02)$\approx$ & \cellcolor[rgb]{ .880,  .880,  .880}0.09844(3.477e-02)$\approx$ & 0.15062(2.444e-02)- & 0.14143(3.318e-02)- & 0.13697(2.988e-02)- & 0.13950(2.039e-02)- & 0.14143(4.597e-02)- \\
    \emph{IBM} & \cellcolor[rgb]{ .751,  .751,  .751}0.03211(3.915e-02) & 0.04853(2.658e-02)- & 0.05127(2.433e-02)- & \cellcolor[rgb]{ .880,  .880,  .880}0.04789(2.024e-02)- & 0.06768(1.980e-02)- & 0.08141(2.961e-02)- & 0.08602(2.768e-02)- & 0.08993(2.604e-02)- & 0.08677(2.554e-02)- \\
    \emph{Drug} & \cellcolor[rgb]{ .751,  .751,  .751}0.09964(6.357e-02) & \cellcolor[rgb]{ .880,  .880,  .880}0.15445(4.872e-02)- & 0.16374(3.813e-02)- & 0.16663(3.885e-02)- & 0.19229(1.770e-02)- & 0.18907(1.645e-02)- & 0.21032(8.059e-03)- & 0.21886(6.814e-03)- & 0.19071(7.638e-03)- \\
    \emph{Patient} & 0.05519(3.048e-02) & 0.06116(1.070e-02)$\approx$ & 0.05580(1.035e-02)$\approx$ & 0.06820(1.228e-02)- & \cellcolor[rgb]{ .880,  .880,  .880}0.05132(5.040e-03)$\approx$ & \cellcolor[rgb]{ .751,  .751,  .751}0.04980(8.573e-03)$\approx$ & 0.05916(7.632e-03)$\approx$ & 0.06344(5.111e-03)$\approx$ & 0.06276(5.871e-03)$\approx$ \\
    \midrule
    ``+/$\approx$/-'' &   -  & 2/8/5  & 1/7/7  & 4/3/8  & 3/5/7  & 4/3/8  &  4/4/7 & 4/2/9  & 3/5/7 \\
    Averaged Ranking &  3.47  & 4.33  & 4.93  & 4.53  & 4.73  & 5.53  &  5.33 & 5.80  & 6.33 \\
    \bottomrule
    \end{tabular}%
  \label{tab:G_mean}%
  \end{adjustbox}
\end{table*}%

\section{Conclusion}\label{sec:conclusion}

To deal with the conflict among accuracy and different fairness measures, this paper applies a novel multi-objective evolutionary learning framework to mitigate unfairness. Two studies of our proposed framework, the tri- and 9-objective optimisation algorithms, are carried out focusing on the four research questions raised in Section \ref{sec:intro}. In particular, we have demonstrated through extensive experimental studies that our multi-objective learning framework is able to learn fair models with high accuracy and outperforms the state-of-the-art. We have shown that we are able to find a more diverse set of fair models than the state-of-the-art. Such a diverse set has enabled us to develop an ensemble of learning models with good performance, outperforming existing ensemble approaches. We have analysed our experimental results from different perspectives in order to understand the results in-depth. We have also shown that our multi-objective learning framework is able to consider a broad range of fairness measures, even those not used in model training. This is an indication that the fair models we have found are robust against different fairness metrics, rather than ``overfiting'' to a specific fairness measure.

In the future, we plan to improve upon several aspects of our work. First, we would like to carry out a deeper analysis of various fairness measures in order to reduce the number of objectives used in multi-objective learning. Second, we plan to investigate the impact of training data distribution on our multi-objective learning framework. Third, new ensemble formulation methods will be investigated. Fourth, since our framework is a population-based approach and each individual in the population is an ML model, if the model size becomes very large, the computation time will increase. Our framework is a good option when multiple objectives are considered or when loss functions are non-differentiable or non-convex. Our framework is better at providing a model set to balance multiple objectives. Our future work will study how to improve the efficiency of the proposed framework on very large models through parallel processing \cite{akay2017parallel,yang2019parallel,yang2021parallel}.

\bibliographystyle{IEEEtran}

\bibliography{fairness}

\begin{IEEEbiography}[{\includegraphics[width=1in,height=1.25in,clip,keepaspectratio]{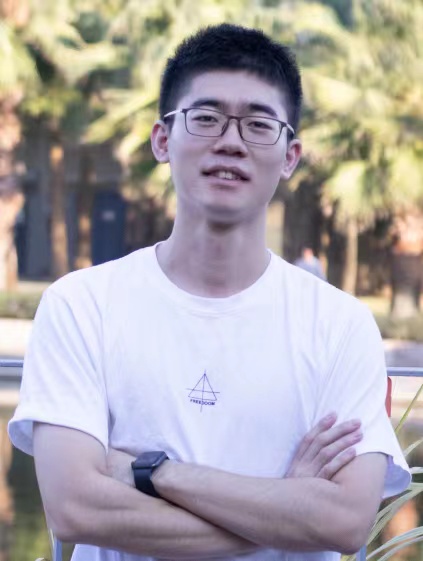}}]{Qingquan Zhang} \textbf{(Student Member, IEEE)} received his M.S. degree in 2022 from the Southern University of Science and Technology (SUSTech), Shenzhen, China and his B.S. degree in 2019 from the Xidian University, Xi'an, China. His current research interests include multi-objective optimisation and fair machine learning.
\end{IEEEbiography}
\vspace{-4em}

\begin{IEEEbiography}[{\includegraphics[width=1in,height=1.25in,clip,keepaspectratio]{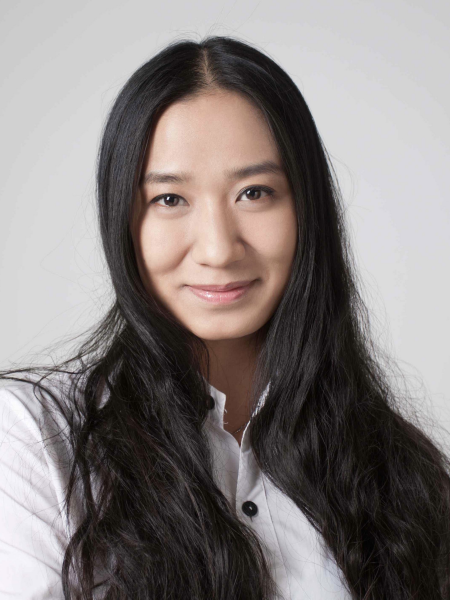}}]{Jialin Liu} \textbf{(Senior Member, IEEE)} received her Ph.D. in 2016 from Universit\'e Paris-Saclay, MSc in 2013 from the \'{E}cole Polytechnique \& Universit\'{e} Paris-Sud, France, and BSc in 2010 from the Huazhong University of Science and Technology, China. Currently, she is a Tenure-Track Assistant Professor at the Department of Computer Science and Engineering, Southern University of Science and Technology (SUSTech), China. Her recent research interests include trustworthy autonomous systems.
\end{IEEEbiography}
\vspace{-4em}

\begin{IEEEbiography}[{\includegraphics[width=1in,height=1.25in,clip,keepaspectratio]{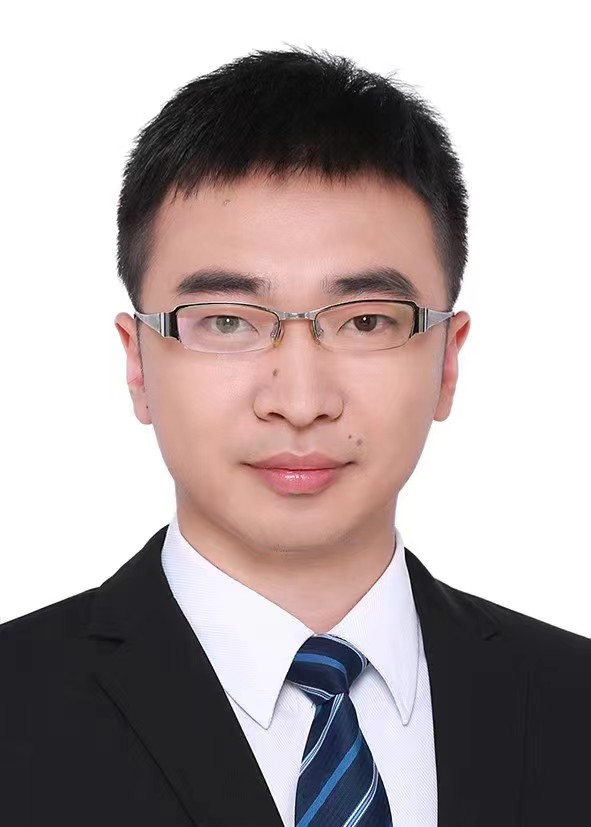}}]{Zeqi Zhang} \textbf{(Member, IEEE)} received his B.S and Ph.D. degrees in electronic engineering from Tsinghua University, Beijing, China, in 2012 and 2018, respectively. He is currently a Senior Researcher with the Trustworthiness Theory Research Center, Huawei Technologies Co., Ltd., Beijing, China. His research interests include trustworthy AI, AI ethics \& governance, and AI standards \& certification.
\end{IEEEbiography}
\vspace{-4em}

\begin{IEEEbiography}[{\includegraphics[width=1in,height=1.25in,clip,keepaspectratio]{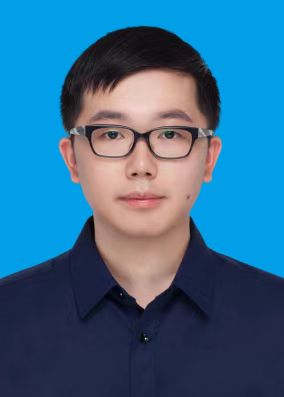}}]{Junyi Wen} received his Ph.D. in 2020 from the Academy of Mathematics and Systems Science, the Chinese Academy of Sciences, China, and his B.S. in 2015 from the Jilin University, China. He is currently a senior engineer in Huawei Technologies Co., Ltd., Shenzhen, China. His current research interests include systematology, trustworthiness theory, AI ethics governance, and symbolic computation.
\end{IEEEbiography}
\vspace{-4em}

\begin{IEEEbiography}[{\includegraphics[width=1in,height=1.25in,clip,keepaspectratio]{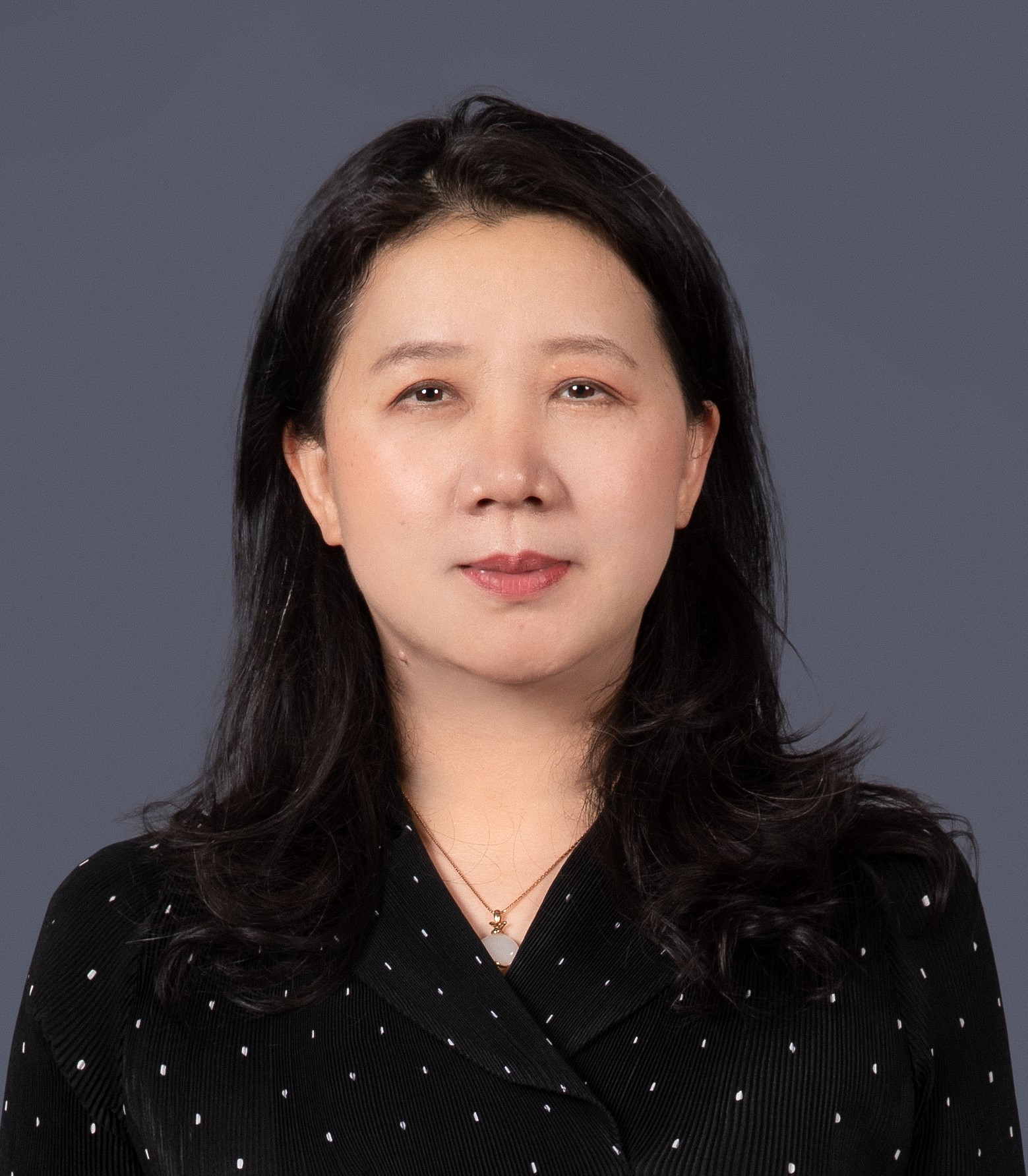}}]{Bifei Mao} received her M.S. and B.S. from the Beihang University, China, in 1994 and 1991, respectively. She is currently a senior researcher of the Trustworthiness Theory Research Center in Huawei Technologies Co., Ltd., Shenzhen, China. Her current research interests include trustworthiness theory, technical ethics and AI ethics governance.
\end{IEEEbiography}
\vspace{-4em}

\begin{IEEEbiography}[{\includegraphics[width=1in,height=1.25in,clip,keepaspectratio]{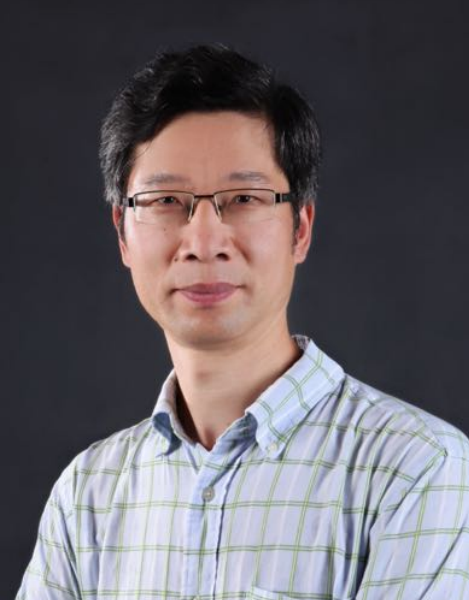}}]{Xin Yao}
\textbf{(Fellow, IEEE)} obtained his Ph.D. in 1990 from the University of Science and Technology of China (USTC), MSc in 1985 from North China Institute of Computing Technologies, and BSc in 1982 from USTC. He is currently a Chair Professor of Computer Science at the Southern University of Science and Technology (SUSTech), China, and a part-time Professor of Computer Science at the University of Birmingham, UK. His recent research interests include trustworthy autonomous systems.
\end{IEEEbiography}

\end{document}